%% file: main.tex
\documentclass{article}
\PassOptionsToPackage{numbers,sort&compress}{natbib}

\usepackage[eandd, preprint, nonatbib]{neurips_2026}
\usepackage[numbers,compress]{natbib}
\makeatletter
\renewcommand{\@noticestring}{
  \centering

}
\makeatother

\input{extra_pkgs}

\usepackage{mathtools}
\usepackage{multirow}
\usepackage{makecell}
\usepackage{subcaption}
\usepackage{wrapfig}
\usepackage{float}
\usepackage{colortbl}
\usepackage{threeparttable}
\usepackage[inline]{enumitem}

\newcommand{\bench}{\textsc{MemLens}}
\newcommand{\cmark}{\ding{51}}
\newcommand{\xmark}{\ding{55}}

\title{\bench{}: Benchmarking Multimodal Long-Term Memory in Large Vision-Language Models}

\author{%
  Xiyu Ren$^{1}$ \quad
  Zhaowei Wang$^{1}$ \quad
  Yiming Du$^{2}$ \quad
  Zhongwei Xie$^{1}$ \\
  \textbf{
  Chi Liu$^{1}$ \quad
  Xinlin Yang$^{1}$ \quad
  Haoyue Feng$^{1}$ \quad
  Wenjun Pan$^{1}$ \quad
  Tianshi Zheng$^{1}$} \\
  \textbf{Baixuan Xu$^{1}$ \quad
  Zhengnan Li$^{3}$ \quad
  Yangqiu Song$^{1}$ \quad
  Ginny Wong$^{4}$ \quad
  Simon See$^{4}$} \\[6pt]
  $^{1}$CSE Deparment, HKUST \quad $^{2}$CUHK \\
  $^{3}$OmniMemory (Shenzhen) Intelligent Technology Co., Ltd. \\
  $^{4}$ NVIDIA AI Technology Center (NVAITC), NVIDIA, Santa Clara, USA \\[4pt]
  \texttt{\{xrenaf, zwanggy, yqsong\}@cse.ust.hk} \\
  \texttt{ydu@se.cuhk.edu.hk} \quad \texttt{lzhengnan389@gmail.com}
}

\begin{document}

\maketitle

\begin{abstract}
Memory is essential for large vision-language models (LVLMs) to handle long, multimodal interactions, with two method directions providing this capability: long-context LVLMs and memory-augmented agents. However, no existing benchmark conducts a systematic comparison of the two on questions that genuinely require multimodal evidence. To close this gap, we introduce \bench{}, a comprehensive benchmark for memory in multimodal multi-session conversations, comprising 789 questions across five memory abilities (information extraction, multi-session reasoning, temporal reasoning, knowledge update, and answer refusal) at four standard context lengths (32K--256K tokens) under a cross-modal token-counting scheme. An image-ablation study confirms that solving \bench{} requires visual evidence: removing evidence images drops two frontier LVLMs below 2\% accuracy on the 80.4\% of questions whose evidence includes images. Evaluating 27 LVLMs and 7 memory-augmented agents, we find that long-context LVLMs achieve high short-context accuracy through direct visual grounding but degrade as conversations grow, whereas memory agents are length-stable but lose visual fidelity under storage-time compression. Multi-session reasoning caps most systems below 30\%, and neither approach alone solves the task. These results motivate hybrid architectures that combine long-context attention with structured multimodal retrieval. Our code is available at \url{https://github.com/xrenaf/MEMLENS}.
\end{abstract}

\section{Introduction}
\label{sec:intro}

Memory is essential for enabling large vision-language models (LVLMs)~\citep{seed2_0,singh2025openaigpt5card} to maintain consistency across extended multimodal interactions and to continuously incorporate new information over time~\cite{openai2023gpt4,anthropic2024claude3,team2024gemini}. When LVLMs are deployed as agents, the inputs they handle accumulate incrementally rather than arriving all at once. As interaction histories expand, the agent must recall past content and reason over the growing context to remain consistent with prior turns. At the same time, it must integrate new facts as they arrive and revise outdated information to keep its knowledge current. Two directions have emerged to provide this capability, namely long-context LVLMs and memory-augmented agents. Long-context LVLMs enlarge the native context window so that the complete dialogue history, including interleaved images, can be processed directly by the model~\cite{team2024gemini,seed1_8,anthropic2025claudesonnet45card}. Memory-augmented agents, building on retrieval-augmented generation, instead compress, index, and selectively retrieve past content from an external store~\cite{packer2024memgptllmsoperatingsystems,mem0,jin2024long}.

Despite progress along both directions, no current benchmark conducts a length-controlled comparison of long-context LVLMs and memory-augmented agents on questions that genuinely require visual evidence. Long-context multimodal benchmarks measure context-length scaling on long documents and retrieval tasks~\cite{mmlongbench,wang2024needlemultimodalhaystack,ma2024mmlongbenchdocbenchmarkinglongcontextdocument}, but cover mainly LVLMs and do not place them alongside memory-augmented agents for direct comparison. Text-only conversational memory benchmarks such as LongMemEval~\cite{wu2025longmemevalbenchmarkingchatassistants} and MemoryAgentBench~\cite{memoryagentbench} overlook the visual modality entirely, treating memory as a single-modality problem. Multimodal conversational benchmarks such as LoCoMo~\cite{maharana2024evaluatinglongtermconversationalmemory} and Mem-Gallery~\cite{memgallery} retain both visual and text modalities, yet most of their questions admit a text-only shortcut, rendering the visual modality redundant. As Table~\ref{tab:benchmark_comparison_full} summarizes, no existing benchmark requires visual evidence to answer its questions while supporting a length-controlled comparison of long-context LVLMs and memory-augmented agents.

\definecolor{softred}{RGB}{200,60,60}
\definecolor{softredbg}{RGB}{253,232,232}
\setlength{\tabcolsep}{4pt}
\begin{table*}[t]
\centering
\caption{%
\textbf{Left:} Comparison of \bench{} with existing long-context and conversational memory benchmarks.
\emph{Type} — Dial (Dialogue) or Doc (Document);
\emph{Max $L$} — maximum context length reported (``---'' if not applicable);
\emph{$L$ Control} — standardized token length control;
\emph{Comp. Eval.} — paper benchmarks both LVLMs and memory-augmented agents on the same data;
\emph{Memory Tasks} — dot pattern showing coverage of the five evaluation tasks
($\bullet$ = supported, $\circ$ = not supported) in fixed order:
\textbf{IE} (Information Extraction), \textbf{MSR} (Multi-Session Reasoning),
\textbf{TR} (Temporal Reasoning), \textbf{KU} (Knowledge Update), \textbf{AR} (Answer Refusal).
\textbf{Right:} Distribution of the 789 evaluation questions in \bench{} across the
five major task types (inner ring) and nine reporting subtypes (outer ring).%
}
\label{tab:benchmark_comparison_full}
\begin{minipage}[c]{0.68\textwidth}
\centering
\resizebox{\linewidth}{!}{%
\small
\begin{tabular}{@{}l cc >{\columncolor{softredbg}}c ccc c@{}}
\toprule
\textbf{Benchmark} & \textbf{Type} & \makecell{\textbf{Max}\\\textbf{$L$}} & \textcolor{softred}{\makecell{\textbf{Multi-}\\\textbf{modal}}} & \makecell{\textbf{Multi-}\\\textbf{Sess.}} & \makecell{\textbf{$L$}\\\textbf{Control}} & \makecell{\textbf{Comp.}\\\textbf{Eval.}} & \makecell{\textbf{Memory}\\\textbf{Tasks}} \\
\midrule
MemoryBank~\cite{memorybank}                                              & Dial & ---  & \xmark & \cmark & \xmark & \xmark & $\bullet\circ\circ\circ\circ$ \\
LoCoMo~\cite{maharana2024evaluatinglongtermconversationalmemory}          & Dial & 10K  & \cmark & \cmark & \xmark & \xmark & $\bullet\bullet\bullet\circ\circ$ \\
MM-NIAH~\cite{wang2024needlemultimodalhaystack}                           & Doc & 128K & \cmark & \xmark & \cmark & \xmark & $\bullet\circ\circ\circ\circ$ \\
MMLongBench-Doc~\cite{ma2024mmlongbenchdocbenchmarkinglongcontextdocument}& Doc & 128K & \cmark & \xmark & \xmark & \xmark & $\bullet\circ\circ\circ\circ$ \\
MRAG-Bench~\cite{mragbench}                                               & Doc & ---  & \cmark & \xmark & \xmark & \xmark & $\bullet\circ\circ\circ\circ$ \\
LongMemEval~\cite{wu2025longmemevalbenchmarkingchatassistants}            & Dial & 1.5M & \xmark & \cmark & \cmark & \xmark & $\bullet\bullet\bullet\bullet\bullet$ \\
MMLongBench~\cite{mmlongbench}                                            & Doc & 128K & \cmark & \xmark & \cmark & \xmark & $\bullet\circ\circ\circ\circ$ \\
Multimodal NIAH~\cite{wang2025multimodalneedlehaystackbenchmarking}       & Doc & 128K & \cmark & \xmark & \cmark & \xmark & $\bullet\circ\circ\circ\circ$ \\
MemAgentBench~\cite{memoryagentbench}                                     & Dial & 115K & \xmark & \cmark & \xmark & \cmark & $\bullet\circ\circ\circ\circ$ \\
\midrule
\textbf{\bench{} (Ours)} & \textbf{Dial} & \textbf{256K} & \textcolor{softred}{\textbf{\cmark}} & \cmark & \cmark & \cmark & $\bullet\bullet\bullet\bullet\bullet$ \\
\bottomrule
\end{tabular}%
}
\end{minipage}%
\hfill
\begin{minipage}[c]{0.30\textwidth}
\centering
\includegraphics[width=\linewidth]{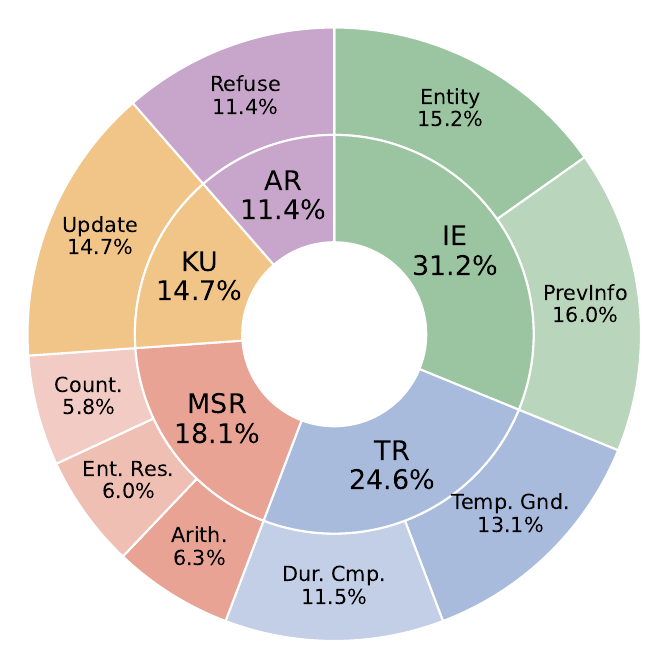}
\end{minipage}
\end{table*}

To bridge the gap, we introduce \bench{}, a comprehensive benchmark for assessing memory in multimodal multi-session conversations. \bench{} consists of 789 questions covering five core memory abilities: information extraction, multi-session reasoning, temporal reasoning, knowledge update, and answer refusal. We design questions that demand cross-modal reasoning over both visual and textual evidence, requiring the retrieval of multimodal information hidden within one or more conversations between a user and an assistant. To verify that solving \bench{} requires visual evidence, we conduct an image-ablation study on the 80.4\% of questions whose evidence includes images: when these images are removed, the accuracy of two frontier LVLMs collapses below 2\% (\S\ref{subsec:cross_modality}). Following the needle-in-a-haystack construction paradigm~\cite{kamradt2023needle}, we design a four-stage pipeline that builds a coherent multi-session chat history for each question, distributing the evidence across one or more user-assistant sessions alongside topically related distractor turns. Because distractor turns can be added independently of the evidence, the chat length is freely extensible. We release four standard context lengths (32K, 64K, 128K, and 256K tokens) under a cross-modal token-counting scheme~\cite{mmlongbench} that aligns text and vision tokens. \bench{} thus enables a length-controlled comparison of long-context LVLMs and memory-augmented agents on multimodal questions spanning five comprehensive memory abilities.

Using \bench{}, we evaluate 27 LVLMs and 7 memory-augmented agents across all four context lengths (32K, 64K, 128K, and 256K). Our evaluation yields three key findings. First, the five memory abilities are largely independent: strong information extraction does not predict multi-session reasoning, and multi-session reasoning caps most evaluated systems below 30\%. Second, the two approaches exhibit complementary failure modes: long-context LVLMs deliver high short-context accuracy through direct visual grounding, but this advantage shrinks as conversations grow; memory agents, in contrast, are length-stable but lose visual fidelity under storage-time compression, and memory-oriented post-training of agent backbones can additionally weaken their abstention behavior. Third, neither approach comes close to solving the task of long-term memory. These results point to a clear next step: architectures that combine long-context attention with structured multimodal retrieval, rather than scaling either component in isolation.

\section{Related Work}
\label{sec:related}

\paragraph{Memory-Augmented LLM Agents.}
Recent surveys systematize memory representations and operations across LLM-based agents~\cite{du2025rethinkingmemoryllmbased}. Text-only memory agents span structured-fact stores~\cite{memorybank,scm}, OS-inspired paging~\cite{packer2024memgptllmsoperatingsystems}, tree-summarized retrieval~\cite{raptor}, neurobiological graphs~\cite{gutierrez2025hipporagneurobiologicallyinspiredlongterm}, relational embeddings~\cite{mem0}, agentic self-organizing notes~\cite{amem}, intent-driven memory selection~\cite{du2025memguideintentdrivenmemoryselection}, RL-selected time-aware memory~\cite{du2025memoryt1reinforcementlearningtemporal}, layered memory tiers~\cite{li2025memosoperatingmemoryaugmentedgeneration}, and sliding-window RL agents~\cite{yu2025memagentreshapinglongcontextllm}. Multimodal extensions add ColPali-style document retrieval~\cite{cho2024m3docragmultimodalretrievalneed,faysse2025colpaliefficientdocumentretrieval}, multimodal embeddings~\cite{vlm2vec}, dual-layer semantic memory~\cite{m2a}, entity-centric audio-visual memory~\cite{long2025seeinglisteningrememberingreasoning}, sparse video memory~\cite{song2024moviechat}, LoRA-tuned session retrieval~\cite{jang2025enablingchatbotseyesears}, and intent-guided multimodal response retrieval over multi-session conversations~\cite{Wang_Du_Liang_Bai_Yang_Wang_Wong_Xu_2025}. Recent work probes the intersection of long-context LLMs and retrieval-augmented generation~\cite{jin2024long,jiang2024longrag,asai2023self}, suggesting retrieval and long attention are complementary rather than competing.

\paragraph{Long-Context and Conversational Memory Benchmarks.}
Most current long-context benchmarks established core protocols for retrieval and length scaling~\cite{bai2024longbench,bai2024longbenchv2,hsiehruler,yen2024helmet,an2024eval,zhang2024bench,li2024needlebench}, but neglect the visual modality. Multimodal extensions, such as MMLongBench~\citep{mmlongbench}, scale context length over documents~\cite{mmlongbench,ma2024mmlongbenchdocbenchmarkinglongcontextdocument}, needle-style retrieval~\cite{wang2024needlemultimodalhaystack,wang2025multimodalneedlehaystackbenchmarking}, long-form video~\cite{zhou2025mlvubenchmarkingmultitasklong,wu2024longvideobenchbenchmarklongcontextinterleaved}, super-long documents~\cite{deng2024longdocurl,chiam}, and multi-image inputs~\cite{wang2024divscene,song2024milebench,wang2024longllava,ye2024mplug}. Yet their inputs are documents or videos rather than multi-session conversations. Memory under conversational interaction therefore remains unexercised. Conversational memory benchmarks restore the multi-session structure but rarely require visual perception: LongMemEval~\cite{wu2025longmemevalbenchmarkingchatassistants} (500 questions over five memory abilities, up to 1.5M tokens), PerLTQA~\cite{du2024perltqapersonallongtermmemory}, and MemoryAgentBench~\cite{memoryagentbench} (retrieval, test-time learning, long-range understanding, selective forgetting) discard images entirely, while LoCoMo~\cite{maharana2024evaluatinglongtermconversationalmemory} and Mem-Gallery~\cite{memgallery} embed images in persona-grounded dialogue but allow most questions to be answered from text alone, so neither stress-tests multimodal memory.

\section{The \bench{} Benchmark}
\label{sec:benchmark}
We propose \bench{}, a multimodal long-term conversational memory benchmark that comprises 789 questions instantiated at four standardized input lengths (32K/64K/128K/256K tokens). In contrast to prior benchmarks restricted to text-only conversations and questions, \bench{} provides multimodal conversation sessions in which text and images are interleaved, together with questions that require cross-modal reasoning over evidence images and the surrounding textual context.
\S\ref{subsec:formulation} defines the five memory abilities, \S\ref{subsec:construction} describes the construction pipeline (Figure~\ref{fig:pipeline}), \S\ref{subsec:quality_control} presents the quality control, and  \S\ref{subsec:cross_modality} describes the cross-modality validation.

\subsection{Memory Abilities}
\label{subsec:formulation}
To comprehensively reflect the retrieval, reasoning, and update that a conversational assistant must perform over a long multimodal history, \bench{} formulates five core memory abilities. We break each ability into subtypes that target specific reasoning operations.
\begin{itemize}
    \item \textit{Information Extraction~(IE)} tests the recall of a specific fact from a single evidence session, with two subtypes. \emph{Entity} questions form a two-hop chain: the model first identifies an abstracted entity in the evidence image, then retrieves the associated information from the surrounding text. \emph{PrevInfo} (previous information) questions instead abstract the session reference, asking the model to recall a visual detail from an image shared in an earlier session. The visual identification hop draws on image-understanding skills such as fine-grained recognition, object counting, spatial reasoning, and numerical reasoning.
    \item \textit{Multi-Session Reasoning~(MSR)} evaluates aggregation across three to eight sessions: \emph{counting} tallies unique items identified only by their evidence images, \emph{arithmetic} sums values stated in text or embedded in visual artifacts, and \emph{entity resolution} determines co-reference by comparing images across sessions.
    \item \textit{Temporal Reasoning~(TR)} assesses joint reasoning over temporal references, including both natural-language expressions and session timestamps, together with visual content: \emph{duration comparison} compares two intervals derived from textual or visual cues, and \emph{temporal grounding} either orders events chronologically or extracts the specific date of an event. Beyond entity abstraction, TR also replaces temporal expressions with visual artifacts such as clock faces and calendar pages.
    \item \textit{Knowledge Update~(KU)} tests the ability to track an evolving user attribute across a chain of four successive updates~\cite{xu2024knowledge} (e.g., ``I used to like apples'' $\to$ ``now I prefer kiwi''), requiring the model to reason from the final state of the chain rather than earlier superseded ones.
    \item \textit{Answer Refusal~(AR)} removes all supporting evidence from an otherwise answerable question, so it can no longer be answered from the remaining context; a correct model must decline rather than hallucinate~\cite{zhang2024rtuning}. AR serves as a calibration check for hallucination detection, not a core memory retrieval task.
\end{itemize}
A single cross-modal principle unifies the four answerable types: the evidence image carries information that the text deliberately withholds, so correct answers require joint visual--textual reasoning (\S\ref{subsec:construction}); \S\ref{subsec:cross_modality} validates this dependency empirically. The formal problem specification appears in Appendix~\ref{app:problem_formulation}. Appendix~\ref{app:taxonomy_rationale} grounds this taxonomy in established memory research, and Appendix~\ref{app:subtype_detail} provides the complete subtype breakdown with representative examples.

\subsection{Data Curation}
\label{subsec:construction}

\begin{figure}[t]
  \centering
  \includegraphics[width=\linewidth]{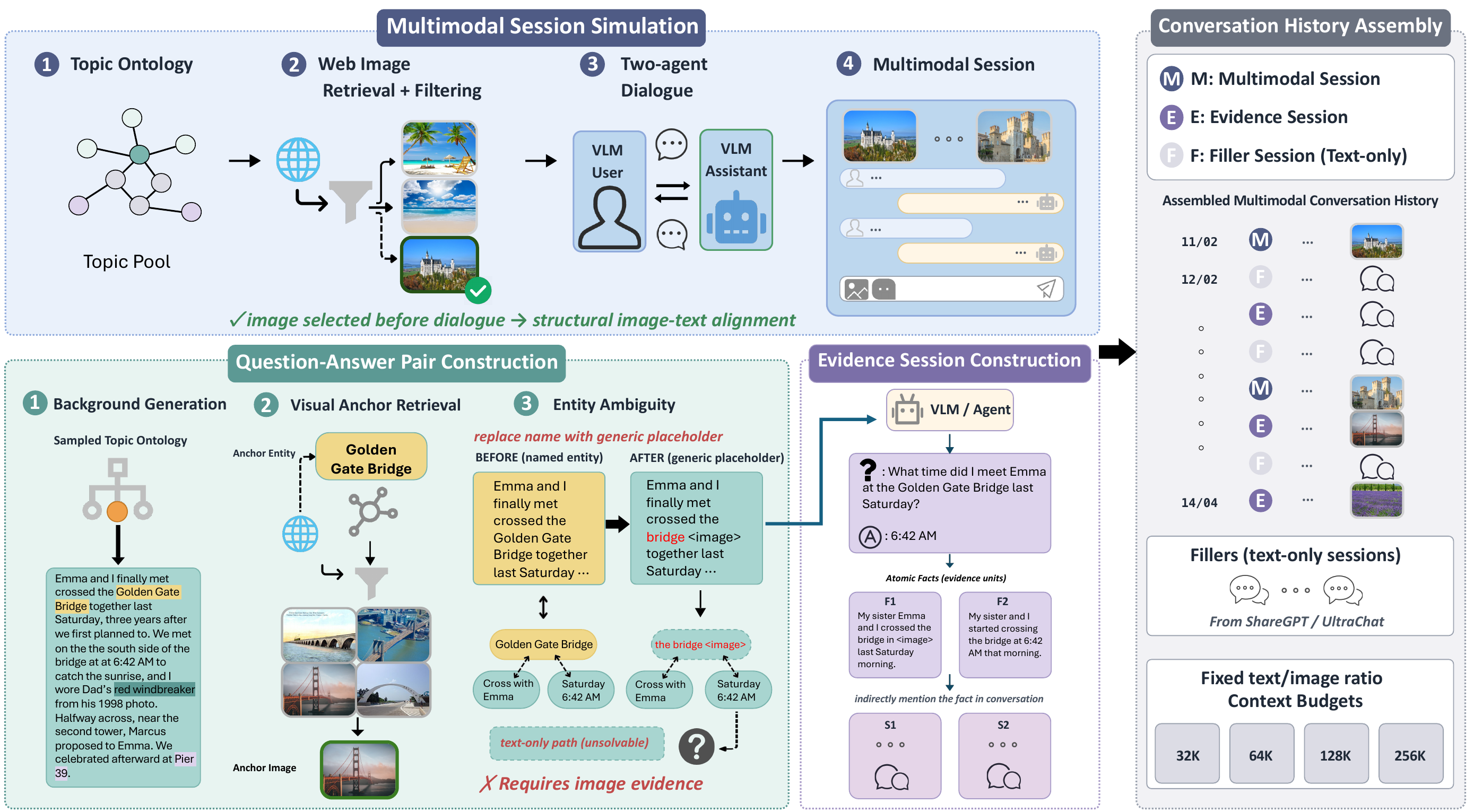}
  \caption{\bench{} construction pipeline.
    }
  \label{fig:pipeline}
\end{figure}

The construction proceeds in four components (Figure~\ref{fig:pipeline}):
\begin{enumerate*}[label=(\roman*)]
\item first, multimodal session simulation generates topic-grounded multimodal dialogue sessions that form the conversation background;
\item next, question construction produces evaluation questions whose answers require visual content through entity abstraction;
\item evidence session construction wraps each evidence fact into a complete session that matches its topical and stylistic profile; and
\item conversation history assembly interleaves evidence, haystack, and text-only filler sessions in timestamp order and compiles each question into four standardized input lengths.
\end{enumerate*}

\paragraph{Multimodal session construction.}
Each session begins with topic sampling from a hierarchical ontology. For each topic, we generate an image query to retrieve a batch of candidate images from the web\footnote{\url{https://github.com/hellock/icrawler}; sourcing, licensing, privacy, and release detailed in Appendix~\ref{app:image_release}.}. We then filter these candidates for visual relevance. From this filtered set, we select the final image. Given this image, we generate sessions through a dual-model simulation with GPT-5.1~\citep{singh2025openaigpt5card} as the user and Gemini-3-Pro~\citep{googledeepmind2026gemini3procard} as the assistant, producing multi-turn dialogues that incorporate the selected images into interleaved image-text sequences~\cite{laurenccon2023obelics}. Because image selection precedes dialogue generation, image--text alignment is a structural property of the data rather than a post-hoc filter. The image retrieval pipeline (\S\ref{app:image_filtering}), full topic ontology (\S\ref{app:topic_ontology}), and image diversity statistics (\S\ref{app:image_diversity}) are detailed in the Appendix.

\paragraph{Question-Answer Pair.}
Question generation follows a four-step pipeline. (i)~A topic is sampled from the hierarchical ontology, and a LVLM generates a background paragraph containing salient named-entities, for instance, a paragraph about San Francisco landmarks that mentions the Golden Gate Bridge. (ii)~One entity is selected as the visual anchor for the question. A text query derived from the entity is issued to the same web-crawling and multi-model scoring pipeline used for haystack images (\S\ref{app:image_filtering}), yielding a high-relevance image~\cite{hu2023open}; the anchor may range from a famous landmark to a specific product model. (iii)~Cross-modal dependency is enforced through \emph{entity abstraction}~\cite{wang2024abspyramid,wang2024absinstruct,viquae}: the entity in the background paragraph is replaced with a higher level concept drawn from a dictionary spanning 55 entity categories (e.g., ``Golden Gate Bridge'' $\to$ ``the bridge shown in \texttt{<image>}''). After replacement, the paragraph no longer names the entity, and only the evidence image can resolve the reference. (iv)~The abstracted paragraph, evidence image, and original entity name are provided to Gemini-3-Pro~\citep{googledeepmind2026gemini3procard}, which generates a (question, answer) pair together with atomic evidence facts. The generation prompt constrains the question to require both the image and the surrounding text, closing the text-only shortcut documented in prior work~\cite{maharana2024evaluatinglongtermconversationalmemory}. The full pipeline and per-subtype generation routes appear in Appendix~\ref{app:abstraction}.

\paragraph{Evidence session.}
Directly inserting evidence statements into the conversation history creates abrupt semantic shifts that make the evidence trivially locatable by similarity-based retrieval. Prior work has shown that increasing contextual similarity between evidence and distractors raises retrieval difficulty, a design principle also adopted by LongMemEval~\cite{wu2025longmemevalbenchmarkingchatassistants}. Accordingly, each evidence fact is wrapped in a complete evidence session generated using the same pipeline as haystack sessions but grounded on the evidence fact rather than a sampled topic. Consequently, the evidence facts match the topical and stylistic profile of the surrounding haystack, so that evidence cannot be located by surface-level shortcuts. To further increase the difficulty, we also prompt the user model to mention the facts indirectly without stressing them: To embed the fact “I started a new job last month,” the user turn might open by asking about updating the tax withholding and mention the job change incidentally later. When an evidence fact carries an evidence image, the image is placed adjacent to the corresponding textual mention within the session, preserving unambiguous image-text co-reference after entity abstraction.

\begin{wraptable}{r}{0.33\textwidth}
  \centering
  \vspace{-12pt}
  \caption{\small Dataset statistics.}
  \label{tab:dataset_stats}
  \small
  \begin{tabular}{@{}lr@{}}
  \toprule
  \textbf{Statistic} & \textbf{Value} \\
  \midrule
  Questions           & 789 \\
  Types / Subtypes    & 5 / 9 \\
  Evidence sessions   & 2{,}145 \\
  Avg.\ turns/session & $\sim$10 \\
  Avg.\ images/session & $\sim$1.5 \\
  Tokens/image        & $\sim$2{,}000 \\
  \midrule
                      & \footnotesize\textit{32K $\to$ 256K} \\
  Sessions/instance   & 14 $\to$ 93 \\
  Images/instance     & 20 $\to$ 138 \\
  \bottomrule
  \end{tabular}
  \vspace{-8pt}
\end{wraptable}

\paragraph{Conversation history assembly.}
For each question, evidence sessions are inserted into a timestamp-ordered history of haystack sessions, with positions chosen uniformly at random (for KU, we preserve the evidence-session order; Appendix~\ref{app:history_assembly}). Haystack sessions are curated to be contextually related but uninformative for the question, and never include answer-relevant details. We vary the number of haystack sessions to produce four standardized context lengths (32K/64K/128K/256K tokens) using the cross-modal counting scheme of MMLongBench~\cite{mmlongbench} (Table~\ref{tab:dataset_stats}). To avoid revealing evidence positions via image clustering, we keep a fixed text-per-image ratio across the history, padding with text-only filler sessions from ShareGPT and UltraChat~\cite{ding2023enhancing}. A post-hoc classifier achieves only marginally above-chance accuracy when separating evidence from haystack text (Appendix~\ref{app:indistinguishability}), and the generator scales beyond 256K.

\subsection{Quality Control}
\label{subsec:quality_control}

\paragraph{Automated filtering.}
Each candidate question is screened by two automated checks. A rule-based pre-filter removes images from the question and evidence and drops cases where the remaining text already determines the answer. An LLM judge (GPT-5.1~\citep{singh2025openaigpt5card}) then sees only the question text (without evidence or conversation history) and removes items solvable from parametric knowledge alone~\cite{mallen2023not}. Remaining questions therefore require using the evidence image, which is a prerequisite for the multimodal analyses below.

\paragraph{Human review.}
Three rounds of human review operate as a cumulative quality gate on the filtered questions. Round~1 verifies that the evidence image carries answer-critical information; for AR questions, which are constructed by removing the evidence facts, this round instead confirms that the removed facts was answer-critical. Round~2 checks that evidence facts are naturally embedded and recoverable from their sessions, and that each session reads as a plausible user--assistant exchange with natural dialogue flow; sessions that fail the naturalness criterion are subsequently refined by humans. Round~3 reviews the haystack sessions to assess both the quality of the images and the naturalness of the dialogues. Together, the automated filters and three-round review reduce the initial pool of 20k candidates to the final 789 questions. The filler sessions drawn from ShareGPT and UltraChat (\S\ref{subsec:construction}) further ground the surrounding context in authentic user--AI conversational patterns. Annotation guidelines and inter-annotator agreement appear in Appendix~\ref{app:annotation}.

\subsection{Cross-modality Validation}
\label{subsec:cross_modality}
The construction pipeline targets image-necessary questions: across \bench{}, 65.7\% of questions are image-essential (the answer is unrecoverable without the evidence image), 14.7\% are image-supportive (the image confirms or disambiguates a textual fact), and 19.6\% are text-sufficient (all AR questions plus a subset of MSR items that test cross-session reasoning over purely textual evidence). We verify this cross-modal dependency with two empirical checks (Table~\ref{tab:mm_purity}). An answerability test supplies each image-essential and image-supportive question ($n = 634$) with its full evidence (textual facts and evidence images) and confirms that the questions are answerable: GPT-5.4 reaches 93.13\% overall and Gemini-3.1-Pro 89.42\%. A multimodal ablation then removes all evidence images: overall accuracy collapses below 2\% for both models. Two frontier proprietary LVLMs converge on near-identical collapses, showing our questions are highly multimodal.

\begin{table}[t]
\centering
\caption{Cross-modality ablation on the image-essential and image-supportive questions. With evidence: question $+$ evidence facts $+$ evidence images, no haystack ($n = 634$). W/o evidence: question $+$ evidence facts, no evidence images.}
\label{tab:mm_purity}
\small
\begin{tabular}{@{}llcccccc@{}}
\toprule
\textbf{Model} & \textbf{Input} & \textbf{Overall} & \textbf{IE} & \textbf{MSR} & \textbf{TR} & \textbf{KU} & \textbf{$\Delta$} \\
\midrule
\multirow{2}{*}{GPT-5.4}
 & With evidence image & 93.13 & 94.31 & 100.00 & 96.91 & 75.86 & --- \\
 & W/o evidence image  &  1.74 &  0.41 &  0.00 &   5.15 &  0.00 & $-$91.39 \\
\addlinespace
\multirow{2}{*}{Gemini-3.1-Pro}
 & With evidence image & 89.42 & 89.02 & 90.21 & 96.19 & 82.24 & --- \\
 & W/o evidence image  &  1.89 &  0.00 &  0.00 &   6.19 &  0.00 & $-$87.53 \\
\bottomrule
\end{tabular}
\end{table}

\section{Evaluation and Analysis}
\label{sec:eval_analysis}

\subsection{Experimental Setup}
\label{subsec:exp_setup}
We evaluate 27 LVLMs and seven memory-augmented agents on \bench{}. The LVLMs span closed-source systems including GPT-5.4~\citep{singh2025openaigpt5card}, Claude Sonnet 4.5~\citep{anthropic2025claudesonnet45card}, and Gemini-3.1-Pro~\citep{googledeepmind2026gemini31pro}, alongside major open-source families such as Kimi-K2.5~\citep{kimiteam2026kimik25visualagentic}, Qwen3.5~\citep{qwenteam2026qwen35nativemultimodalagents}, GLM-4.6V~\citep{zai2025glm46vcard}, and Gemma3~\citep{gemmateam2025gemma3technicalreport}; the complete list is provided in Appendix~\ref{app:eval_setup}. The memory-augmented agents comprise three multimodal pipelines (M3-Agent~\cite{long2025seeinglisteningrememberingreasoning}, M2A~\cite{m2a}, and M3C~\cite{jang2025enablingchatbotseyesears}) and four text-only pipelines (Mem0~\cite{mem0}, MemOS~\cite{li2025memosoperatingmemoryaugmentedgeneration}, MemAgent-7B~\cite{yu2025memagentreshapinglongcontextllm}, and Memory-T1~\cite{du2025memoryt1reinforcementlearningtemporal}), with backbone and adapter details reported in Appendix~\ref{app:supplementary_experiments}. Since the four text-only memory agents do not accept visual inputs, we follow standard practice and replace the image inputs with captions generated by BLIP-2~\cite{li2023blip2}; the remaining input adaptations are documented in Appendix~\ref{app:eval_setup}. For comparability, the LVLMs are evaluated at three context lengths (32K, 64K, and 128K), as many of them do not natively support 256K, whereas the agents are evaluated across all four context lengths (32K, 64K, 128K, and 256K). We report LLM-as-Judge accuracy~\cite{zheng2023judging} using Qwen3-VL-235B-A22B-Instruct as the judge, cross-validated by GPT-5.4-mini re-judging ($\kappa = 0.93$) and a three-annotator human consensus (Appendix~\ref{app:judge_validation}).

\subsection{Main Results}
\label{subsec:main_results}

\begin{figure*}[!t]
\centering
\includegraphics[width=\linewidth]{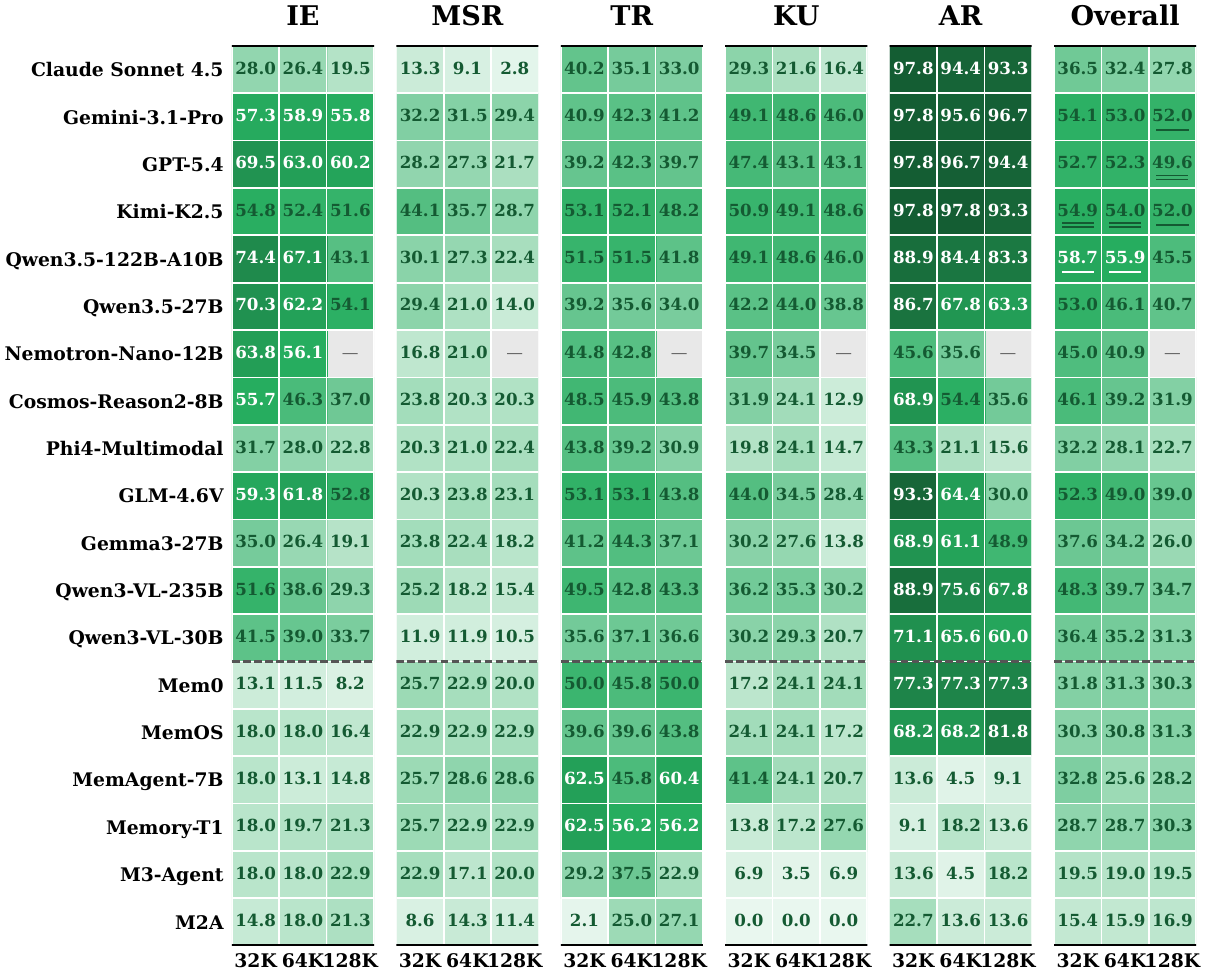}
\caption{Per-type accuracy (\%) by context length for representative 13 LVLMs and 6 memory-augmented agents. Each panel shows one question type; cells use a green colormap. Missing cells indicate models that exceed their usable context budget at 128K. LVLMs are evaluated on the full 789-question benchmark, agents on the 195-question canonical subset. Full rosters appear in Tables~\ref{tab:per_type_full_vlm} and~\ref{tab:per_type_full_agent}.}
\label{fig:per_type_heatmap}
\end{figure*}

\paragraph{Overall performance.}
Figure~\ref{fig:per_type_heatmap} reports per-type accuracy across context lengths for a representative 13-LVLM cohort and six memory agents; the full 27-LVLM and seven-agent rosters are in Appendix~\ref{app:extended}. At 32K, the top eight LVLMs fall within a 6.34\% band, so short-context accuracy no longer separates frontier systems. The picture inverts at 128K: several open-weight leaders lose more than 13\%, while Gemini-3.1-Pro retains 51.99\% accuracy and degrades least overall (a 2.11\% drop). AR shows the steepest context-driven decline in the open-weight LVLM family, suggesting that hallucination control is the ability most exposed to evidence dilution at long contexts. Memory agents occupy a narrower range, with the top four text-only systems clustering within 5\% of each other, while M3-Agent, M3C, and M2A fall substantially lower. Memory agents trail LVLMs across nearly all types, with the largest gaps on visually grounded retrieval (IE, KU) and on answer refusal (AR).

\paragraph{Type-specific difficulty.}
Per-type accuracy ceilings range from 97.78\% on AR down to 44.06\% on MSR (Kimi-K2.5) at 32K. AR, which serves as a calibration check for hallucination detection, is the easiest type as expected, although its apparent ease erodes substantially at long contexts for open-weight LVLMs (e.g., GLM-4.6V AR drops from 93.33\% at 32K to 30.00\% at 128K). TR follows (60.82\% ceiling), because timestamps and dates in session metadata provide explicit retrieval anchors. IE (74.39\% ceiling) and KU (50.86\%) form an intermediate band: IE requires visual grounding, with Entity questions demanding two-hop reasoning through the abstracted image reference, while KU requires tracking a four-fact update chain in which missing a single image anchor flips the predicted state. MSR is the hardest, as cross-session aggregation over three to eight sessions defeats every evaluated system: only Kimi-K2.5 (44.06\%) and Gemini-3.1-Pro (32.17\%) clear 30\% by margin, exposing this as the shared capability ceiling of \bench{}. Within types, IE-Entity is consistently harder than single-hop IE-PrevInfo, and MSR-Arithmetic is the most difficult subtype overall (Appendix~\ref{app:subtype_detail}).

\begin{wrapfigure}[13]{r}{0.3\linewidth}
\vspace{-8pt}
\centering
\includegraphics[width=\linewidth]{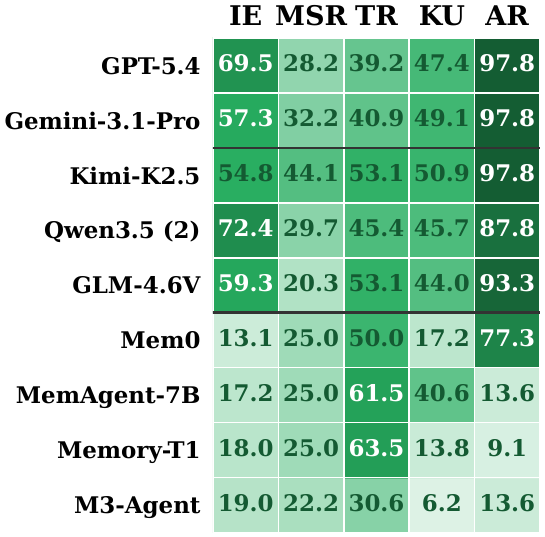}
\caption{\small Memory-ability specialization across representative LVLMs and memory agents.}
\label{fig:specialization}
\vspace{-6pt}
\end{wrapfigure}

\paragraph{No model dominates all memory abilities.}
No single model family dominates across all types (Figure~\ref{fig:specialization}). GLM-4.6V~\citep{zai2025glm46vcard} leads TR but collapses on KU, while Qwen3.5~\citep{qwenteam2026qwen35nativemultimodalagents} inverts the pattern. Kimi-K2.5~\citep{kimiteam2026kimik25visualagentic} is relatively strongest on MSR at 32K, though this advantage fades at longer contexts. Gemini-3.1-Pro is the only model simultaneously competitive on IE, KU, and MSR at 128K. Memory agents exhibit an inverted profile: Memory-T1 reaches high TR accuracy through BM25 date matching but falls well below direct LVLMs on IE, substituting keyword retrieval for the visual grounding that IE demands.

\subsection{Analysis}
\label{subsec:analysis}

\begin{figure}[!t]
\centering
\begin{subfigure}[b]{0.48\linewidth}
\centering
\includegraphics[width=\linewidth]{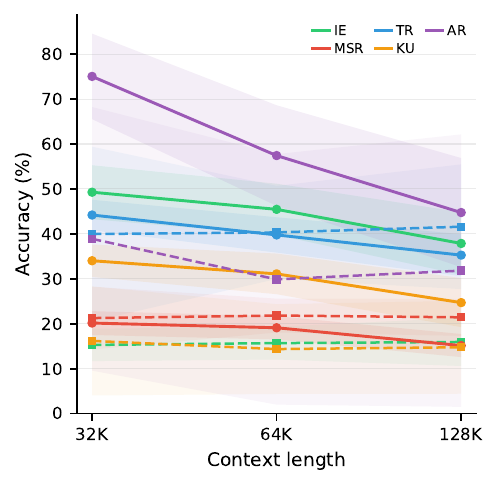}
\caption{Per-type accuracy vs.\ input length. Solid: LVLM average (27 models, $n{=}789$); dashed: agent average (7 systems, $n{=}195$). Bands: 95\% CI.}
\label{fig:context_degradation}
\end{subfigure}\hfill
\begin{subfigure}[b]{0.48\linewidth}
\centering
\includegraphics[width=\linewidth]{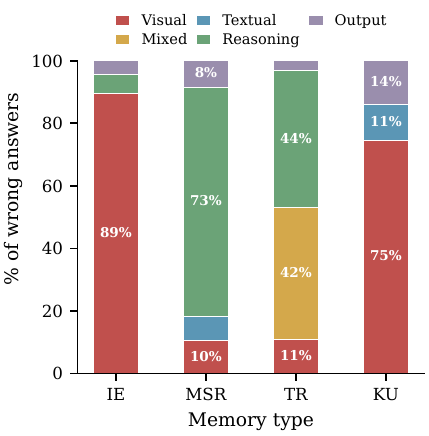}
\caption{Wrong-answer error decomposition at 128K context across the four answerable memory types, split by failure modality (visual, textual, mixed, reasoning, output). Category definitions appear in Table~\ref{tab:modality_mapping}.}
\label{fig:visual_error}
\end{subfigure}
\end{figure}

\paragraph{Current memory pipelines lose faithfulness to original visual evidence.}
The gap between agents and LVLMs is largest on the visually grounded types (IE and KU). Despite their different input formats (\S\ref{subsec:exp_setup}), both text-only and multimodal pipelines compress evidence visual information into a fixed memory representation at storage time, leaving the original image pixels inaccessible at query time. Retrieval over these compressed encodings is also less reliable than retrieval over the raw conversation text. The loss is sharpest on IE and KU because captions and embeddings retain only the gist of an image and discard fine-grained visual cues, such as counts, attributes, and spatial relations, that these two types specifically probe. The same gap appears when blip-captions replace raw images for text-only agents, indicating that the bottleneck lies in lossy cross-modality storage rather than in the answering model. Closing this gap requires memory architectures that preserve image-level evidence rather than caption-based compression.

\paragraph{Post-training on memory agent backbones weakens abstention.}
Memory agents fall far below their direct-inference counterparts on AR. We evaluate two agents that keep the backbone frozen, Mem0 (77.27\%) and MemOS (68.18\%), and five that further finetune the backbone via RL or LoRA for memory management (M2A, M3-Agent, M3C, MemAgent-7B, and Memory-T1). While the frozen-backbone agents preserve much of the abstention behavior of the underlying LLM, the others collapse to 9--22\% AR. M2A reaches only 22.73\% on the same Qwen3-VL-8B backbone that score 81.82\% under direct inference, and the backbone ablation in Appendix~\ref{app:agent_underperformance} confirms that stronger backbones alone do not close this gap. This suggests that the reward design of current RL/SFT fine-tuning on memory agent backbones optimizes mainly answer correctness and retrieval success, providing no signal that refusing an unanswerable question is correct, so abstention degrades after training. Future agent designs should therefore optimize memory access, answer accuracy, and evidence-sensitive abstention jointly~\citep{zhai2026abstainr1}, rather than treating memory management as independent of hallucination control.

\paragraph{LVLMs and memory agents show complementary scaling trade-offs.}
Figure~\ref{fig:context_degradation} plots accuracy across five abilities from 32K to 128K for both LVLMs and memory agents. The two approaches respond to context scaling in structurally different ways. LVLMs degrade substantially on retrieval-heavy types, with IE and KU losing $\sim$20\% and $\sim$12\% respectively as evidence images become harder to locate among growing filler content. MSR shows apparent flatness that reflects a floor effect near 30\% rather than genuine robustness. AR shows the steepest LVLM decline ($\sim$75\% at 32K to $\sim$45\% at 128K), suggesting that growing filler content erodes abstention and pushes LVLMs to hallucinate on unanswerable questions rather than refuse them. Six of seven memory agents, by contrast, stay within $\pm 7\%$ from 32K to 256K because their retrieve-then-reason pipeline is length-invariant by construction. The two failure modes are orthogonal: LVLMs lose to context length, while memory agents lose to lossy multimodal compression at storage time. Because each architecture covers only one axis, scaling along that axis leaves the other failure mode untreated, motivating hybrid designs that span both axes.

\begin{wrapfigure}{r}{0.3\linewidth}
\vspace{-8pt}
\centering
\includegraphics[width=\linewidth]{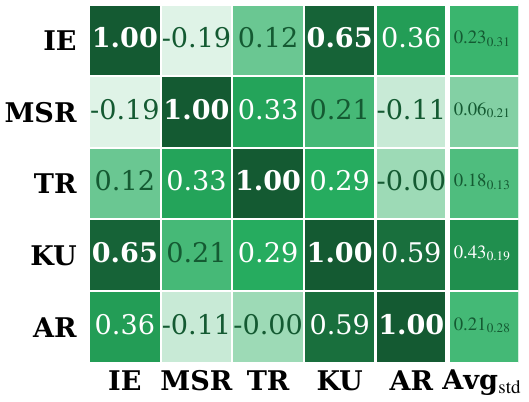}
\caption{\small Spearman rank correlation ($\rho$) at 32K across all 34 evaluated LVLMs and memory agents.}
\label{fig:type_correlation}
\vspace{-6pt}
\end{wrapfigure}

\paragraph{Memory ability correlations reveal distinct sources of difficulty.}
We analyze pairwise Spearman correlations among the five question types at 32K (Figure~\ref{fig:type_correlation}). Correlations vary substantially across type pairs, indicating that the types do not measure a single capability. IE and KU form the most consistent retrieval-oriented pair: they have the strongest correlation at 32K and remain significantly correlated across context lengths, reflecting their shared need to first locate the relevant evidence image. KU also correlates with AR at 32K, suggesting that some KU questions still depend on accurate evidence selection. In contrast, MSR shows weak correlations with IE and AR at all lengths, implying that its main challenge is aggregating information across multiple evidence pieces rather than retrieving a single image. These findings reveal two complementary difficulty axes: an evidence-retrieval axis (IE, KU) and an aggregation axis (MSR). They also illustrate how a single aggregate score can obscure differences among long-context abilities, motivating per-type evaluation alongside overall performance~\cite{hsiehruler,yen2024helmet}. Full correlation matrices at 64K and 128K appear in Appendix~\ref{app:extended_analysis}.

\paragraph{Error analysis.}
Figure~\ref{fig:visual_error} decomposes wrong answers at 128K across the four answerable types, with the category definitions provided in Appendix~\ref{app:wrong_answer_taxonomy}. On IE and KU, nearly 90\% of errors fall in the Visual category, showing that the model fails to locate or read the evidence image; once the image is reached, the answer is usually extracted correctly. TR errors split between Mixed and Reasoning, reflecting both image-supportive grounding and small closed-set selection. MSR is the only type whose errors are dominated by the Reasoning category (73\%). We further conduct an oracle-retrieval diagnostic in Appendix~\ref{app:msr_ceiling}, which restores accuracy on MSR to 90--100\% when only the required sessions are supplied. This suggests a reasoning error at long context is largely induced by upstream retrieval failure, since the model computes correctly over the wrong subset of evidence. LVLMs and memory agents fail differently. LVLMs lose visual evidence as filler content accumulates in the context window, whereas memory agents lose it through lossy cross-modal compression at storage time. As memory grows, near-misses at 32K shift toward total misses at 128K (Appendix~\ref{app:wrong_answer_taxonomy})~\cite{levy2024same}, suggesting that growth weakens evidence retrieval rather than reasoning. Future memory designs should therefore focus more on cross-modal evidence-retrieval fidelity instead of solely reasoning improvements.

\section{Conclusion}
\label{sec:conclusion}

We introduced \bench{}, the first benchmark for multimodal conversational memory to evaluate LVLMs and memory-augmented agents under a unified, length-controlled protocol (32K–256K tokens; five memory abilities). The task remains challenging: even the strongest LVLM reaches only 58.68\% at 32K. LVLMs degrade sharply once input contexts grow, while memory agents stay length-stable but discard fine-grained visual evidence at storage time, and post-training on memory agent backbones can erode the agents' abstention behavior. The five abilities show low cross-type correlation, confirming that per-type evaluation is necessary, and nearly 90\% of information-extraction errors stem from visual perception rather than comprehension, indicating that scaling first harms evidence retrieval, not reasoning. Visual-evidence retention and retrieval, rather than raw scaling of either context or memory, therefore emerges as the principal bottleneck to address in the future.


\begin{ack}
The authors of this paper were supported by the National Key Research and Development Program of China (2025YFE0200500), the ITSP Platform Research Project (ITS/189/23FP) from ITC of Hong Kong, SAR, China, and the AoE (AoE/E-601/24-N), the RIF (R6021-20) and the GRF (16205322) from RGC of Hong Kong, SAR, China. We also thank the NVIDIA AI Technology Center (NVAITC) and OmniMemory (Shenzhen) Intelligent Technology Co., Ltd. for their support and additional funding.
\end{ack}

\section*{Ethics Statement}

\bench{} is designed exclusively for evaluating multimodal long-term conversational memory in large vision-language models (LVLMs) and memory-augmented agents under a unified, length-controlled protocol. The benchmark is not intended as a training dataset for developing memory-equipped systems but rather as a controlled diagnostic for fair comparison across architectures. We release every artifact under frozen version tags so that any reported leaderboard cell remains traceable to the exact item set on which it was scored.

The 4{,}695 source images underlying \bench{} are retrieved from public web image search via iCrawler against a non-person-centric topic ontology; candidates carrying watermarks, stock-photo logos, or copyright overlays are excluded at retrieval time, and the construction prompts never request the model to identify, name, or describe any depicted person. Author-produced artifacts (questions, evidence facts, prompts, and human-annotation records) are released under CC-BY-4.0 and the evaluation harness under MIT, while third-party images retain their original source-site licenses; per-image provenance metadata (source URL, retrieval timestamp, perceptual hash) accompanies the release, and a takedown contact in the project repository allows seven-day removal of any flagged image.

Human review is conducted by project members rather than crowd-workers, so no third-party data collection or human-subjects protocol is involved. Three sequential rounds (Appendix~\ref{app:annotation}) audit cross-modal necessity at the question level, naturalness and recoverability at the session level, and a stratified sample of haystack sessions before any item reaches the released set.

We caution that \bench{} is not intended as a training dataset: exposure of the 789 evaluation items as supervised data would compromise their diagnostic value, and the datasheet explicitly discourages such use. While the benchmark is grounded in multi-session multimodal conversation, the five memory abilities it isolates (information extraction, multi-session reasoning, temporal reasoning, knowledge update, and answer refusal) describe general functional requirements of any long-horizon conversational assistant. We therefore expect \bench{} to serve as a useful indicator of memory robustness in adjacent settings, such as document-grounded or voice-grounded assistants, where retrieval, integration, and calibrated abstention play comparable roles.

\section*{Reproducibility Statement}

All closed-source LVLM evaluations and synthetic data construction were performed via the public APIs of OpenAI, Anthropic, and Google, with a total experimental cost of approximately 4{,}500 USD covering both pipeline construction and benchmarking. Open-weight LVLMs are served locally with vLLM v0.17--0.18 on 8$\times$A100-80GB nodes with tensor parallelism for context windows at and above 128K. Per-model decoding budgets, judge protocol, retrieval depth for memory agents, and adapter details for the four text-only pipelines (Mem0, MemOS, MemAgent-7B, Memory-T1) are documented in Appendix~\ref{app:eval_setup}. The four-stage construction pipeline (multimodal session simulation, question construction, evidence session construction, conversation history assembly) is described in Section~\ref{sec:benchmark}, with per-subtype generation routes in Appendix~\ref{app:abstraction} and full prompt templates in Appendix~\ref{app:prompts}. Quality control combines automated filtering with a three-round human review (Section~\ref{subsec:quality_control}, Appendix~\ref{app:annotation}) and an LLM-as-Judge protocol cross-validated by an independent judge family and a three-annotator human consensus; agreement statistics and per-question-type breakdowns appear in Appendix~\ref{app:judge_validation}. The 789-question benchmark, 4{,}695 source images with provenance metadata, evaluation harness, and prompt templates are publicly released under frozen version tags at \url{https://huggingface.co/datasets/xiyuRenBill/MEMLENS} (dataset) and \url{https://github.com/xrenaf/MEMLENS} (code) so that every reported leaderboard cell can be reproduced end-to-end.


\bibliographystyle{unsrtnat}
\bibliography{ref}

\newpage
\appendix

\section{Use of Large Language Models}
\label{app:llm_use}

During the preparation of this paper, we made controlled use of LLMs (specifically ChatGPT and Claude) as auxiliary writing tools. The LLMs were employed solely for stylistic refinement---improving the fluency, grammar, and readability of paragraphs that were originally drafted by the authors. Importantly, the scientific content, methodology, experimental design, and main narrative of the paper were fully conceived, written, and validated by the authors without reliance on LLMs. Therefore, LLMs served purely in a supportive role for polishing author-written text, and their contribution does not rise to the level of co-authorship.
\vspace{-3mm}

\section{Evaluation Setup}
\label{app:eval_setup}

\subsection{Models}
\label{subsec:models}

We evaluate 27 LVLMs and seven memory-augmented agent systems. Because compute is limited and not all LVLMs support a 256K-token context, we evaluate the LVLMs at 32K, 64K, and 128K for a fair comparison, and the memory-augmented agents at all four lengths.

\paragraph{Closed-source API models.}
GPT-5.4~\citep{singh2025openaigpt5card} (OpenAI), Claude Sonnet 4.5~\citep{anthropic2025claudesonnet45card} (Anthropic), and Gemini-3.1-Pro~\citep{googledeepmind2026gemini31pro} (Google).

\paragraph{Open-source LVLMs.}
Kimi-K2.5~\citep{kimiteam2026kimik25visualagentic} (Moonshot), 
Qwen3-VL family~\citep{bai2025qwen3vltechnicalreport} (235B-A22B and 30B-A3B MoE; 8B, 4B, 2B dense) in both Instruct and Thinking modes, GLM-4.6V~\citep{zai2025glm46vcard}, GLM-4.5V~\citep{vteam2026glm45vglm41vthinkingversatilemultimodal}, Gemma3~\citep{gemmateam2025gemma3technicalreport} (27B, 12B, 4B), Phi4-Multimodal~\citep{microsoft2025phi4minitechnicalreportcompact}, Cosmos-Reason2-8B~\citep{nvidia2025cosmosreason2card}, Nemotron-Nano-12B~\citep{nvidia2025nvidianemotronnanov2}, and the Qwen3.5 family~\citep{qwenteam2026qwen35nativemultimodalagents} (122B-A10B MoE; 27B, 9B, 4B, 2B dense).

\paragraph{Memory-augmented agents.}
We evaluate seven memory-augmented agents, split into three multimodal agents and four text-only agents. The multimodal agents are M3-Agent~\cite{long2025seeinglisteningrememberingreasoning} (ColPali~\cite{faysse2025colpaliefficientdocumentretrieval} retrieval + RL-trained Qwen2-VL-7B), M2A~\cite{m2a} (dual-layer SQLite + SigLIP2 + Qwen3-VL-8B), and M3C~\cite{jang2025enablingchatbotseyesears} (LoRA-adapted Qwen2-VL-2B session retrieval). The text-only agents are Mem0~\cite{mem0} (FAISS vector store + Qwen3-8B), MemOS~\cite{li2025memosoperatingmemoryaugmentedgeneration} (layered memory architecture + Qwen3-8B), MemAgent-7B~\cite{yu2025memagentreshapinglongcontextllm} (recurrent sliding-window + RL-trained Qwen2.5-7B), and Memory-T1~\cite{du2025memoryt1reinforcementlearningtemporal} (BM25 text retrieval + RL-trained Qwen2.5-3B).

For agents that include RL-finetuned or LoRA-adapted models (Memory-T1, MemAgent-7B, M2A, M3C, M3-Agent), we use their released checkpoints to reflect each system's published capability. For Mem0 and MemOS, which are framework-based systems without custom-trained models, we use Qwen3-8B~\citep{yang2025qwen3technicalreport} as the backbone for both, providing a matched-scale comparison against the open-weight Qwen3 family in our roster. To disentangle backbone quality from architecture, we additionally evaluate both frameworks with alternative backbones, including the original gpt-4.1-mini for Mem0 and Qwen2.5-7B~\citep{qwen2024qwen25} for MemOS (Table~\ref{tab:backbone_ablation}).

\paragraph{Agent evaluation protocol.}
Because agent pipelines are substantially slower than direct LVLM inference (M2A takes roughly $60\times$ longer per question), we evaluate all agents on a stratified 195-question subset ($\sim$1/4 of the benchmark; derivation in Appendix~\ref{app:canonical195}). LVLMs are evaluated at 32K, 64K, and 128K; agents are evaluated at all four context lengths including 256K. The four text-only agents receive BLIP-2~\cite{li2023blip2} generated image captions in place of actual images as text input. Among the three multimodal agents, M3-Agent is a video-based model that does not natively support interleaved image-text conversations; we render each session as a composite image and feed sessions as a sequence of images. M2A and M3C process the multimodal input directly. Table~\ref{tab:new_model_list} lists the full model specifications.

\begin{table}[h]
    \caption{
        Model specifications for all 27 evaluated LVLMs. Length means the training length (default) or claimed context window (denoted by $^\dagger$). All models are instruction-tuned. ``Image Proc.'' stands for Image Processing, which is mainly Dynamic Resolution ViT~\citep{wang2024qwen2} or Dynamic Tiling~\citep{wu2024deepseek}. Most models extend context length via RoPE~\citep{su2024roformer} with position interpolation techniques~\citep{chen2023extending,ding2024longrope,pengyarn}.
    }
\label{tab:new_model_list}
\centering
\small
\begin{tabular}{lrcr}
\toprule
\textbf{Name} & \textbf{Length} & \textbf{Image Proc.} & \textbf{\# Params} \\
\midrule
\multicolumn{4}{l}{\textit{Proprietary}} \\
\midrule
Claude Sonnet 4.5 & 200,000$^\dagger$ & ? & ? \\
Gemini-3.1-Pro & 1,048,576$^\dagger$ & ? & ? \\
GPT-5.4 & 1,000,000$^\dagger$ & ? & ? \\
\midrule
\multicolumn{4}{l}{\textit{Qwen3.5}} \\
\midrule
Qwen3.5-122B-A10B & 262,144 & Dynamic-Resolution ViT & 122B \\
Qwen3.5-27B & 262,144 & Dynamic-Resolution ViT & 27B \\
Qwen3.5-9B & 262,144 & Dynamic-Resolution ViT & 9B \\
Qwen3.5-4B & 262,144 & Dynamic-Resolution ViT & 4B \\
Qwen3.5-2B & 262,144 & Dynamic-Resolution ViT & 2B \\
\midrule
\multicolumn{4}{l}{\textit{Other Open-Source}} \\
\midrule
Nemotron-Nano-12B & 131,072 & Dynamic Tiling & 12B \\
Cosmos-Reason2-8B & 262,144 & Dynamic-Resolution ViT & 8B \\
Phi4-Multimodal & 131,072 & Dynamic Tiling & 5.6B \\
Kimi-K2.5 & 262,144$^\dagger$ & Dynamic-Resolution ViT & 1T \\
\midrule
\multicolumn{4}{l}{\textit{GLM}} \\
\midrule
GLM-4.6V & 131,072 & Dynamic Resolution ViT & 106B \\
GLM-4.5V & 65,536 & Dynamic Resolution ViT & 106B \\
\midrule
\multicolumn{4}{l}{\textit{Gemma3}} \\
\midrule
Gemma3-27B & 131,072$^\dagger$ & Dynamic Tiling & 27B \\
Gemma3-12B & 131,072$^\dagger$ & Dynamic Tiling & 12B \\
Gemma3-4B & 131,072$^\dagger$ & Dynamic Tiling & 4B \\
\midrule
\multicolumn{4}{l}{\textit{Qwen3-VL}} \\
\midrule
Qwen3-VL-235B (T) & 262,144 & Dynamic-Resolution ViT & 235B \\
Qwen3-VL-235B (I) & 262,144 & Dynamic-Resolution ViT & 235B \\
Qwen3-VL-30B (T) & 262,144 & Dynamic-Resolution ViT & 30B \\
Qwen3-VL-30B (I) & 262,144 & Dynamic-Resolution ViT & 30B \\
Qwen3-VL-8B (T) & 262,144 & Dynamic-Resolution ViT & 8B \\
Qwen3-VL-8B (I) & 262,144 & Dynamic-Resolution ViT & 8B \\
Qwen3-VL-4B (T) & 262,144 & Dynamic-Resolution ViT & 4B \\
Qwen3-VL-4B (I) & 262,144 & Dynamic-Resolution ViT & 4B \\
Qwen3-VL-2B (T) & 262,144 & Dynamic-Resolution ViT & 2B \\
Qwen3-VL-2B (I) & 262,144 & Dynamic-Resolution ViT & 2B \\
\bottomrule
\end{tabular}
\end{table}

\paragraph{Agent input-format adapters and protocol asymmetry.}
Memory agents and direct LVLMs do not consume the same input. Each agent ingests the conversation through an adapter that depends on its architecture: the four text-only agents see BLIP-2~\cite{li2023blip2} captions in place of every evidence image, M3-Agent sees one composite image per session because its video-style backbone does not natively accept interleaved image-text sequences, and only M2A and M3C process the original interleaved messages directly. At answer time the asymmetry persists: the text-only agents have no path back to pixel evidence, M3-Agent re-attends a session-level composite, while M2A and M3C retrieve embedding-based memory entries (no raw pixels at query time). Direct LVLMs, in contrast, attend over the original conversation pixel-for-pixel within the model's context window. Table~\ref{tab:agent_adapters} lists each adapter explicitly. We do not normalize this asymmetry because the adapter is part of the system being evaluated---released checkpoints assume the input format their authors trained on, and any uniform substitute would either degrade architectures that depend on caption-only memory (Mem0, MemOS, MemAgent-7B, Memory-T1) or block agents whose backbones cannot accept interleaved input (M3-Agent). Reported deficits relative to direct LVLMs therefore conflate adapter-induced visual information loss with retrieval and reading quality. The matched-backbone contrast in Appendix~\ref{app:agent_underperformance} (M2A vs.\ direct Qwen3-VL-8B-Instruct on identical weights) and the backbone ablations for Mem0 and MemOS (Table~\ref{tab:backbone_ablation}) isolate the architectural component, while the BLIP-2 captioning step bounds the visual-information ceiling for text-only agents from above.

\begin{table}[h]
\caption{Per-agent input adapters. ``Write-time visual'' is the form in which evidence images enter the memory store; ``answer-time visual'' is the form available to the backbone when the question is asked. Direct LVLMs (omitted from the table) attend over the original interleaved conversation pixel-for-pixel within the model's context window.}
\label{tab:agent_adapters}
\centering
\small
\begin{tabular}{llll}
\toprule
\textbf{Agent} & \textbf{Backbone} & \textbf{Write-time visual} & \textbf{Answer-time visual} \\
\midrule
\multicolumn{4}{l}{\textit{Multimodal-backbone agents}} \\
\midrule
M3-Agent      & Video LVLM (Qwen2-VL-7B)    & Composite per-session image       & Retrieved session composite(s) \\
M2A           & Native LVLM (Qwen3-VL-8B)   & Original images                   & Stored embeddings \\
M3C           & Native LVLM (Qwen2-VL-2B)   & Original images                   & Stored embeddings \\
\midrule
\multicolumn{4}{l}{\textit{Text-only-backbone agents}} \\
\midrule
Mem0          & Text LLM (Qwen3-8B)        & BLIP-2 captions only              & Captions only \\
MemOS         & Text LLM (Qwen3-8B)        & BLIP-2 captions only              & Captions only \\
MemAgent-7B   & Text LLM (Qwen2.5-7B)      & BLIP-2 captions only              & Captions only \\
Memory-T1     & Text LLM (Qwen2.5-3B)      & BLIP-2 captions only              & Captions only \\
\bottomrule
\end{tabular}
\end{table}

\subsection{Metrics}
\label{subsec:metrics}

\paragraph{LLM-as-Judge (J).}
LLM-as-Judge accuracy~\cite{zheng2023judging} is our primary metric; a single canonical judge, Qwen3-VL-235B-A22B-Instruct with thinking disabled, scores every accuracy number reported in this paper.
String-match metrics such as substring exact match fail across \bench{} because answers span heterogeneous formats---binary choice, counts, currency and date values, ranked orderings, short fill-ins, and explicit refusals---and LVLMs frequently wrap the correct answer in a multi-sentence rationale or a thinking trace.
The judge reads the question, reference answer, and raw model output and emits a binary correct/incorrect verdict under task-specific criteria (Appendix~\ref{app:eval_setup}).
To prevent degenerate outputs from contaminating scores, the pipeline tail-truncates outputs beyond 6{,}000 characters, auto-zeros responses exceeding 500 parsed words, and instructs the judge to reject circular reasoning.
Judge reliability is quantified in Appendix~\ref{app:judge_validation}: the Qwen3-VL-235B judge agrees with GPT-5.4-mini on 96.40\% of a stratified 800-item re-judge sample (Cohen's $\kappa = 0.93$; Spearman $\rho = 0.97$ at the model level) and with a three-annotator human consensus on 93.60\% of 484 items ($\kappa = 0.86$), and the residual gap in either direction does not reorder the leaderboard.
A format-dependent leniency on very short (1--3 word) answers is documented and corrected in Appendix~\ref{app:judge_validation}.

\paragraph{Substring Exact Match (SE).}
The SE metric checks whether the normalized reference answer appears as a substring of the model output.
This rule is generous to verbose models but inflates scores when the output contains reasoning traces that happen to mention the reference string.

\paragraph{Coverage and Per-Answer Accuracy.}
Beyond overall accuracy, we decompose performance into Coverage (Cov; fraction of the 699 answerable questions the model attempts) and Per-Answer Accuracy (PA; accuracy on attempted answers only).
The two components recover overall accuracy via $J \approx (\text{Cov} \times \text{PA} \times 699 + \text{AR}_{\text{correct}})\,/\,789$, where 699 is the answerable subset and 789 is the full benchmark; the decomposition exposes the coverage--accuracy trade-off discussed in Appendix~\ref{app:coverage_analysis}.

\subsection{Infrastructure}
\label{subsec:infra}

Local models are served via vLLM (v0.17--0.18) with FlashAttention-2~\cite{dao2023flashattention2} on 8$\times$A100-80GB nodes with tensor parallelism for 128K inputs.
API models use provider endpoints with concurrent requests (4--8 threads).
Generation length is set to 2,048 tokens for direct models and 16,384 for thinking models to accommodate reasoning traces.

\section{Dataset Construction Details}
\label{app:dataset_construction}

\subsection{Problem Formulation}
\label{app:problem_formulation}

An evaluation instance in \bench{} is the 4-tuple $(S, q, I, a)$, where $S = [(t_1, M_1), \dots, (t_N, M_N)]$ is a sequence of $N$ time-stamped multi-turn sessions ($t_1 < \dots < t_N$), each interleaving text and images; $\mathcal{V}(S)$ denotes the set of all images appearing in $S$, and $I \subseteq \mathcal{V}(S)$ is the subset that carries answer-critical visual information not recoverable from the surrounding text; $q$ is a query targeting one of five memory abilities; and $a$ is the gold answer, or the literal string \texttt{NOT\_MENTIONED} for answer-refusal items. A correct system must (i)~localize the relevant evidence sessions within a long, distractor-heavy history and (ii)~ground its answer in cross-modal reasoning over $I$.

\subsection{Topic Ontology}
\label{app:topic_ontology}

Each of the four answerable memory abilities (IE, MSR, TR, KU) is backed by a dedicated ontology of $\sim$100 topics organized into three complementary tracks: \emph{identification} (recognizing real-world entities such as products, landmarks, and dishes), \emph{experience} (everyday activities and lifestyle moments), and \emph{document} (text-rich artifacts such as receipts, menus, and schedules). Each topic is further expanded into $\sim$30 fine-grained subtopics, yielding more than 12{,}000 subtopics across the four answerable types. Table~\ref{tab:topic_ontology} groups the de-duplicated topic titles into 19 sub-attributes, each illustrated by representative leaves.

\begin{longtblr}[
  caption = {\bench{} topic ontology. Three tracks decompose into 19 sub-attributes; each row lists a small sample of the de-duplicated topic titles drawn from the underlying $\sim$400-title pool.},
  label = {tab:topic_ontology},
]{
  colspec = {@{}lX@{}},
  width = \linewidth,
  rowhead = 1,
  cells = {font=\footnotesize},
  rowsep = 1.5pt,
}
\toprule
\textbf{Sub-attribute} & \textbf{Representative topic titles} \\
\midrule
\SetCell[c=2]{l} {\textit{Track 1: Identification --- recognizing real-world entities ($\sim$40\%)}} & \\
\midrule
1.1 Retail \& commerce & supermarket shelves; bakery counters; pharmacy aisles; hardware/tool shops; music and sports stores; thrift stores; street food carts; fish and flower markets; secondhand electronics stalls; warehouse club bulk aisles \\
1.2 Home objects \& belongings & kitchen appliances; knife and utensil drawers; pantry; spice jars; bookshelves; CD/vinyl collections; gaming desk; TV streaming setup; closet; jewelry/watch layout; makeup vanity; perfume shelf; skincare bottles; pet accessories; board games \\
1.3 Vehicles \& mobility & personal cars; scooters; motorcycles; bicycle helmets and locks; bike-sharing docks; parked bicycles; license plates; parking lot vehicles; public transit vehicles; e-scooter models on sidewalks \\
1.4 Urban environment \& landmarks & street signs; shopfront names; building numbers; mailboxes and doorbells; landmarks and tourist attractions; street murals and sculptures; residential houses; recycling bins and trash trucks; playgrounds; basketball courts; public swimming pools \\
1.5 Workspace \& institutions & office kitchen; supply closet; lobby directories; vending machines; public library shelves; gym equipment; pharmacy windows; temple/church exteriors \\
\midrule
\SetCell[c=2]{l} {\textit{Track 2: Experience --- everyday activities and lifestyle moments ($\sim$40\%)}} & \\
\midrule
2.1 Dining \& food activities & casual restaurants; fast-food counters; food courts; cafes; street food walks; late-night noodle shops; home cooking; weekend brunch; batch cooking; baking experiments; cooking classes; asian grocery hunts; farmer market pickups \\
2.2 Fitness \& outdoor recreation & indoor climbing; badminton; trampoline park; futsal; table tennis; jogging routes; soccer at local field; pickup basketball; public swimming; sunrise yoga; weekend cycling; camping; hiking; beach outings; ski trips \\
2.3 Social \& community gatherings & house parties; potluck dinners; game and movie nights; karaoke rooms; casual bar meetups; community festivals; charity runs; language exchanges; quiz nights; board game cafés; dance socials; volunteer shifts; sports stadium visits \\
2.4 Cultural \& entertainment outings & art museum visits; science museum visits; gallery visits; historic house tours; concerts and small gigs; movie theater; outdoor summer cinema; theme park rides; indoor arcades; cosplay meetups; comic store browsing; photography walks \\
2.5 Routines \& errands & morning bathroom and kitchen routines; commuting by bus or train; weekend laundry folding; apartment cleaning; grocery top-up trips; salon and barber visits; pet grooming; ride-hailing pickups; fuel stops; DIY home repair; content drafting on phone \\
2.6 Travel \& mobility outings & weekend road trips; vacation rental stays; hotel check-ins; airport and station transit; riverside walks; coastal promenades; neighborhood evening jogs; weekend tram rides; day-trip walks \\
2.7 Home life \& indoor hobbies & morning balcony coffee; evening living-room TV; evening reading corner; family breakfast; hobby painting sessions; sewing and knitting; pottery class; home garden balcony; video recording for reels \\
\midrule
\SetCell[c=2]{l} {\textit{Track 3: Document --- text-rich artifacts ($\sim$20\%)}} & \\
\midrule
3.1 Receipts \& bills & grocery self-checkout receipts; restaurant takeout receipts; monthly utility bills; taxi and ride receipts; parking garage receipts; car service invoices; veterinarian receipts \\
3.2 Tickets, passes \& itineraries & event tickets and seating plans; museum day passes; transit monthly passes; intercity bus tickets; hotel booking confirmations; train reservations; travel itineraries; gym class punch cards; university club ID cards \\
3.3 Schedules \& calendars & school and work weekly schedules; printed school lunch menus; cinema film schedules; fitness class timetables; chore rotation charts; carpool rotation sheets; weekly meal plans; fitness challenge calendars \\
3.4 Contracts \& application forms & rental leases; job offer letters; gym membership contracts; international visa packets; school field trip permission slips; gym class registration forms; university club membership forms; workplace sign-in sheets \\
3.5 Personal records \& logs & personal budget sheets and bank statements; fitness tracker logs; meal tracking notebooks; handwritten to-do lists; travel packing checklists; recipe cards; phone contact lists; household grocery expense notebooks \\
3.6 Subscriptions \& digital records & streaming subscription billing emails; digital movie rental history; online shopping order screenshots; streaming watch history lists; social media profile/settings screens; social media content calendars \\
3.7 Health \& service records & medical prescriptions and pharmacy labels; clinic appointment cards; home appliance warranty cards; smartphone repair invoices; household appliance repair job cards; product warranty manuals \\
\bottomrule
\end{longtblr}

\subsection{Subtype Detail}
\label{app:subtype_detail}

\begin{figure}[h]
\centering
\includegraphics[width=0.85\linewidth]{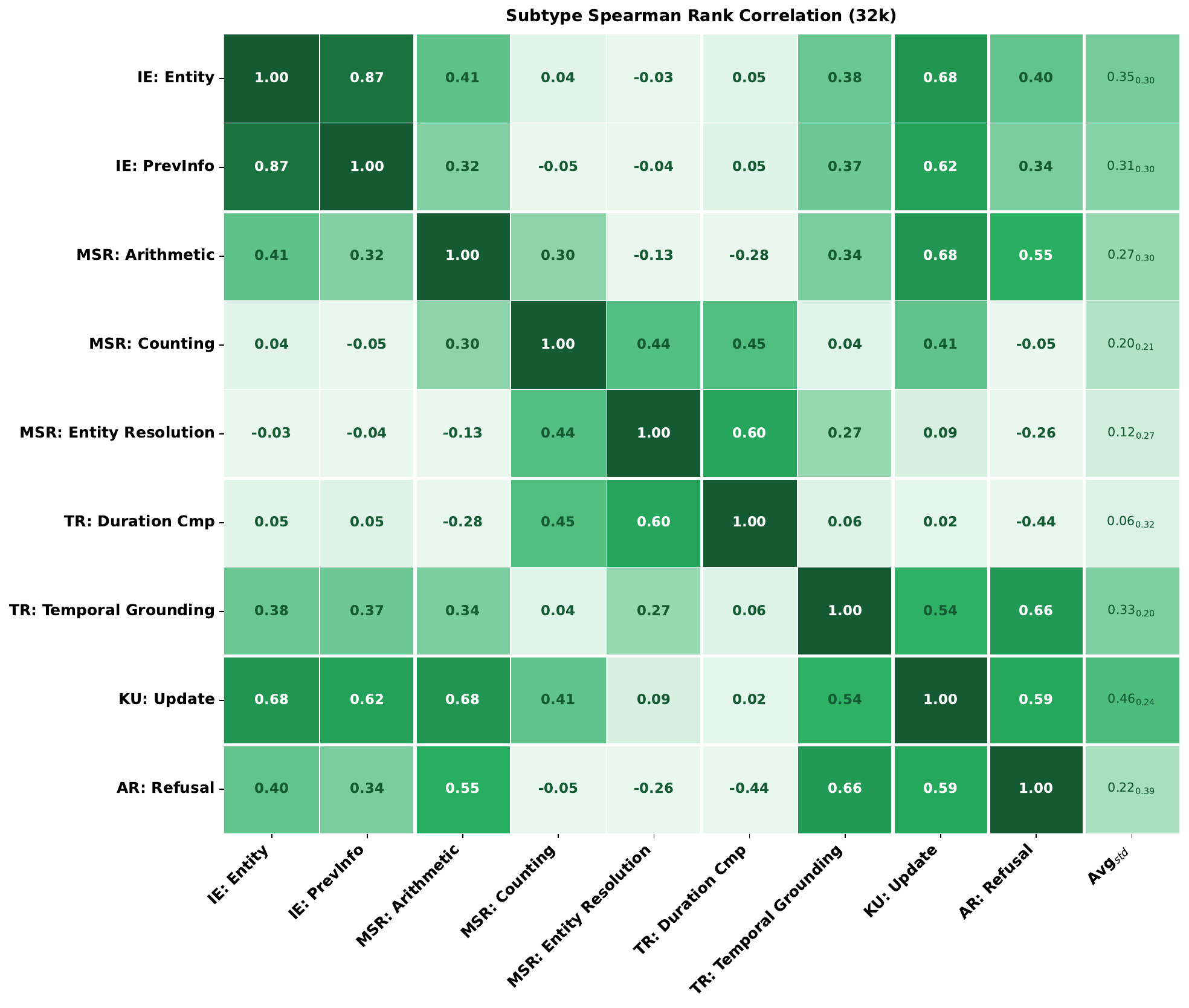}
\caption{Cross-subtype Spearman rank correlation across the evaluated models ($n=789$, 9 reporting subtypes). IE Entity and IE PrevInfo form the only near-ceiling pair ($\rho = 0.87$ at 32K, $0.94$ at 128K), reflecting their shared retrieval skill. MSR-internal correlation is weak at 32K (mean pairwise $\rho = 0.20$) and rises to $0.38$ at 128K as a shared-failure artifact of MSR collapsing toward the floor; TR-internal correlation stays near zero at both contexts ($\rho = 0.06$). The heterogeneity supports evaluating all five major types separately rather than reporting a single aggregate (\S\ref{subsec:analysis}).}
\label{fig:subtype_correlation}
\end{figure}

Table~\ref{tab:subtype_detail} provides the complete subtype taxonomy of \bench{}, listing the 8 answerable subtypes plus Answer Refusal with per-subtype question counts, the visual skill or reasoning operation each subtype isolates, and a representative example. Across the benchmark, 65.7\% of questions are image-essential (the evidence image is required to recover the answer), 14.7\% are image-supportive (the image confirms or disambiguates a textual fact that a strong text-only model could otherwise guess), and 19.6\% are text-sufficient (all AR questions plus a subset of MSR items retained by design). The image-essential share is highest in IE and MSR, substantial in KU, and lower in TR, where a portion of items renders the temporal cue as explicit textual dates or session-boundary timestamps. The cross-subtype correlation structure (Figure~\ref{fig:subtype_correlation}) confirms that the nine subtypes do not consistently correlate with each other, supporting per-type evaluation rather than a single aggregate.

\begin{table}[h]
\centering
\caption{Complete subtype taxonomy of \bench{} ($n=789$). Each subtype isolates a distinct visual skill or reasoning operation. Example questions are shortened for space; full versions appear in the released dataset.}
\label{tab:subtype_detail}
\small
\begin{tabular}{@{}llcp{6.2cm}@{}}
\toprule
\textbf{Subtype} & \textbf{Type} & \textbf{$n$} & \textbf{Skill Tested / Representative Example} \\
\midrule
Entity & IE & 120 & Identify a visually grounded entity via two-hop reasoning (disambiguation, alignment, counting, spatial, arithmetic). \emph{``What phrase does that message spell out?''} \\
\addlinespace
PrevInfo & IE & 126 & Recall a visual detail from an earlier session's image (screenshot, app interface, photo). \emph{``What color is the engine bay painted?''} \\
\addlinespace
Arithmetic & MSR & 50 & Sum or compute over prices/quantities scattered across 3--8 sessions. \emph{``How much total have I spent on weights?''} \\
\addlinespace
Counting & MSR & 46 & Count entities matching a criterion across sessions. \emph{``How many cats do I have?''} \\
\addlinespace
Entity & MSR & 47 & Resolve whether two cross-session references denote the same entity, either by counting distinct entities or by Y/N identity matching. \emph{``Is the new bird the same species as Rio?''} \\
\addlinespace
Duration Cmp & TR & 91 & Compare two durations derived from session timestamps and visual cues (clocks, calendars). \emph{``Which duration is longer?''} \\
\addlinespace
Temporal Grounding & TR & 103 & Locate when an event occurred---either by sorting events chronologically or extracting a specific date---using session timestamps, textual dates, and clock/calendar images. \emph{``Sort these facts in chronological order.''~/ ``When did I complete X? (YYYY/MM/DD)''} \\
\addlinespace
Update & KU & 116 & Track a 4-fact preference chain and report the current state, distinguishing it from outdated values. \emph{``What do I prefer now?''} \\
\addlinespace
Refusal & AR & 90 & Decline to answer when the evidence image has been deliberately removed from the context. \emph{``How many cables are visible?'' (image absent)} \\
\midrule
\multicolumn{2}{@{}l}{\textbf{Total}} & \textbf{789} & \\
\bottomrule
\end{tabular}
\end{table}

Table~\ref{tab:question_examples} presents representative examples from each question type, illustrating the cross-modal reasoning chain required to reach the correct answer.

\begin{table}[h]
\centering
\caption{Representative evaluation questions from \bench{}, one per question type. Each question requires joint visual and textual reasoning: the evidence image carries the discriminative information that the surrounding text deliberately withholds. Fact sketches are abbreviated; full evidence sessions appear in the released dataset.}
\label{tab:question_examples}
\small
\begin{tabular}{@{}lp{5.0cm}lp{4.8cm}@{}}
\toprule
\textbf{Type} & \textbf{Question} & \textbf{Answer} & \textbf{Why Multimodal} \\
\midrule
IE & What shape is the Mount Fuji detail in the artwork my Tokyo friend was studying? & Triangular & Resolve ``the artwork'' from \texttt{<image>} of \emph{The Great Wave off Kanagawa}, then extract visual shape. \\
\addlinespace
MSR & How much total have I spent on Arkham Horror expansions recently? & \$124.94 & Sum \$59.99 + \$64.95 from two sessions; one price is visible only on the box \texttt{<image>}. \\
\addlinespace
TR & Which duration is longer: my time living in London vs.\ my stint as a barista? & A & London start date is on a boarding pass \texttt{<image>}; barista dates are textual. Compare two durations. \\
\addlinespace
KU & Back when I liked it mainly for quick, filling meals, what was my favorite? & Farro & Track a 4-image preference chain; each update is anchored by a different grain \texttt{<image>}. \\
\addlinespace
AR & How many vertical cables are visible on one side of the bridge? & \texttt{NOT\_MENTIONED} & The bridge photograph has been deliberately removed; a correct model must refuse. \\
\bottomrule
\end{tabular}
\end{table}

\subsection{Question Construction Pipeline}
\label{app:abstraction}

The question construction pipeline produces cross-modal evaluation questions from a topic ontology through four steps: background generation, entity selection with image retrieval, entity abstraction, and question generation. This pipeline is shared across all four answerable question types (IE, MSR, TR, KU), with type-specific adaptations described in the per-subtype routes at the end of this section.

\paragraph{Background generation.}
For each question, a topic is sampled from the hierarchical ontology (\S\ref{app:topic_ontology}), and Gemini-3-Pro generates a background paragraph of three to five sentences grounded in that topic. The paragraph is required to contain at least two named entities drawn from recognizable real-world referents, such as landmarks, commercial products, biological species, or cultural institutions, whose visual appearance is distinctive and web-searchable. For example, a paragraph about San Francisco landmarks might mention the Golden Gate Bridge, Alcatraz Island, and Fisherman's Wharf, providing both visual anchors and textual context from which questions can be derived. The background paragraph serves a dual purpose: it supplies the factual context that the question will target, and it introduces the named entity that will undergo abstraction in the next stage.

\paragraph{Entity selection and image retrieval.}
From the background paragraph, one named entity is selected as the visual anchor for the question. The selection requires that the entity be visually identifiable in a photograph, sufficiently specific that a web image search returns relevant results, and critical to at least one fact stated in the paragraph. A text query derived from the entity name is then issued to the web-crawling pipeline, which retrieves a batch of candidate photographs using iCrawler. Each candidate is scored by the multi-model filter described in \S\ref{app:image_filtering}, combining CLIP (ViT-L/14), SigLIP (ViT-SO400M), and a text--text cosine similarity channel between the query and a BLIP-2 caption of the candidate image. The highest-scoring candidate that passes negative-content filtering and global URL deduplication is selected as the evidence image. This retrieval pipeline is identical to the one used for haystack images (\S\ref{subsec:construction}), ensuring uniform visual quality across evidence and background content. The selected entity may range from a famous landmark to a specific product model or a biological species.

\paragraph{Entity abstraction.}
The surface form of the selected entity is then masked in the background paragraph and replaced with a natural anaphor that references the evidence image~\cite{viquae,mragbench,chen2023pretrainedvisionlanguagemodels}. The replacement proceeds in two stages. The entity is first classified into one of 55 semantic categories spanning places (museums, parks, restaurants, bridges, temples), organizations (companies, foundations, institutes), objects (books, paintings, vehicles, instruments), and generic fallbacks. A replacement phrase is then sampled from a type-aware dictionary of $\sim$170 entries, with each category providing three to four paraphrase variants to ensure lexical diversity across questions. For instance, ``Golden Gate Bridge'' may be replaced with ``the bridge shown in \texttt{<image>},'' while ``Portland Art Museum'' might become ``the gallery I visited, shown in \texttt{<image>}.'' When the entity type cannot be matched to any dictionary category, a generic fallback such as ``the place shown in \texttt{<image>}'' is used.

After replacement, the background paragraph no longer contains the entity name. The anaphor is deliberately under-specified: ``the bridge'' could refer to any of thousands of bridges worldwide, and only the accompanying evidence image allows the reader to identify the specific referent. This design enforces a cross-modal dependency in which the text provides the factual context (e.g., year of construction, architectural style) while the image provides the entity identity, following the visual information-seeking paradigm established in prior work~\cite{chen2023pretrainedvisionlanguagemodels,viquae}.

For KU questions, dictionary-based replacement alone is insufficient because the evolving attribute chain tracks concrete items within a single category (e.g., successive favorite fruits). In this case, an LLM rewrites each evidence fact with a short sensory or visual descriptor of at most five words that does not name the category. For instance, ``blood orange'' becomes ``this tangy round thing \texttt{<image>}'' and ``blueberries'' becomes ``these tiny purple spheres \texttt{<image>}.'' The descriptor is constrained to be plausible for multiple items within the same category, preserving the ambiguity needed for cross-modal dependency.

\paragraph{Question generation and quality verification.}
The abstracted background paragraph, the evidence image, and the original entity name are provided to Gemini-3-Pro, which generates a (question, answer) pair together with one or more atomic evidence facts. The generation prompt enforces two constraints: the question must be answerable only when both the image and the surrounding text are available, and the answer must be derivable from the evidence facts without requiring external knowledge beyond the provided context. For MSR questions, a three-layer text-hackability defense additionally verifies that no textual cue leaks the entity identity: anti-leakage prompt rules prevent the generation model from naming the entity, a rule-based pre-filter rejects facts containing the entity name or close synonyms, and an LLM text-only judge confirms that the answer cannot be derived from the textual evidence alone. All generated questions then pass through the automated filtering and human review pipeline described in \S\ref{subsec:quality_control}.

\paragraph{Per-subtype generation routes.}
The five question types follow different generation paths, each tailored to the visual skill under test:

\begin{itemize}
    \item \textbf{IE -- Entity two-hop} (5 subtypes, $n=120$). An LLM generates a two-hop chain~\cite{chang2022webqa}: the first hop resolves the entity from the evidence image, and the second hop retrieves a property of that entity from the surrounding text. The five subtypes vary the visual skill required for the first hop: disambiguation (distinguishing similar-looking entities), alignment (matching an image to a textual description), counting (enumerating items in the image), spatial reasoning (locating objects relative to each other), and arithmetic (computing from visually presented numbers).

    \item \textbf{IE -- PrevInfo} (3 subtypes, $n=126$). The question asks about a visual detail from an image shared in an earlier conversation session. The three subtypes correspond to the image source: a screenshot of a chat interface, an app or web interface, or a natural photograph. Entity abstraction is applied to the session reference rather than the entity itself.

    \item \textbf{KU -- Update} ($n=116$). A 4-fact atomic chain is generated: each fact updates a user attribute (e.g., favorite drink changes from tea $\to$ coffee $\to$ matcha $\to$ espresso martini). The question asks for the current state, requiring the model to locate all four updates and identify the latest. Entity abstraction masks the attribute's anchoring entity so the evidence image is needed to identify which preference chain is being queried.

    \item \textbf{MSR} (3 subtypes, $n=143$). Facts are distributed across 3--8 sessions. Arithmetic ($n=50$) requires summing prices or quantities; Counting ($n=46$) requires enumerating entities matching a criterion; Entity ($n=47$) requires determining whether two cross-session references denote the same entity, either by counting distinct entities or by Y/N identity matching. A 3-layer text-hackability defense (anti-leakage prompt rules, rule-based pre-filter, LLM text-only judge) ensures that the answer cannot be derived without the evidence images.

    \item \textbf{TR} (2 subtypes, $n=194$). Duration Comparison ($n=91$) derives two durations from session timestamps and visual cues (clocks, calendars) and asks which is longer. Temporal Grounding ($n=103$) bundles two operations: order ranking ($n=24$, sort events chronologically) and date extraction ($n=79$, answer \emph{``When did X happen?''} in \texttt{YYYY/MM/DD}). Three generation modes cross with these operations---Mode B renders the temporal cue itself as a visual artifact (clock, calendar, receipt), Mode C pairs an entity image with explicit textual dates, and Mode D pairs an entity image with session-level timestamps that serve as implicit temporal anchors---providing comprehensive coverage of temporal--visual integration.
\end{itemize}

\subsection{Image Filtering}
\label{app:image_filtering}

A unified cross-modal image filter is applied to every image in \bench{}---both needle images in evidence sessions and filler images in haystack sessions---to maintain consistent visual quality across the benchmark. The filter operates in two stages.

\paragraph{Stage 1: Multi-channel relevance scoring.}
Each candidate image is scored against its associated textual query on three independent channels: CLIP~\cite{radford2021learning} (ViT-L/14), SigLIP~\cite{zhai2023sigmoid} (ViT-SO400M), and a text--text cosine between the query and a BLIP-2~\cite{li2023blip2} caption of the image. The CLIP threshold set to 0.30. A candidate must exceed the CLIP threshold and at least one of the two secondary channels to pass.

\paragraph{Stage 2: Negative-content filtering.}
Candidates that pass relevance scoring undergo a negative-keyword filter that rejects images containing watermarks, stock-photo logos, copyright overlays, or resolution artifacts. For DocVQA-style images~\cite{mathew2021docvqa} (receipts, menus, forms), a separate multimodal judge (GPT-4V) inspects each image for watermark presence and rejects any image flagged as containing artifacts that could distract from the document content.

\paragraph{Deduplication.}
A persistent URL registry tracks every image URL used across all generation batches, enforcing global image uniqueness: no two questions in \bench{} share the same source image. Within each question, duplicate detection uses perceptual hashing (pHash) with a Hamming distance threshold of 6 to reject near-duplicate candidates from the same web-retrieval session.

\subsection{Image Diversity}
\label{app:image_diversity}

Unlike prior multimodal long-context benchmarks that categorize visual
content by overlapping semantic topics (mixing scenes, media types, and
tasks on a single axis), \bench{} partitions images by their dominant
visual format, with categories selected to cover distinct perceptual
regimes (Table~\ref{tab:image_categories}).

The first and largest category, natural photographs, covers real-world
scenes, objects, food, and lifestyle items retrieved through a
CLIP-filtered web pipeline. This category tests open-world object and
scene recognition, spatial understanding, and visual attribute extraction
in unstructured environments, and supplies the visual evidence for most
IE, MSR, and KU questions. The second category, text-rich documents,
includes receipts, menus, invoices, posters, price tags, and forms in
which dense text and spatial layout carry the answer; reasoning over these
images requires OCR together with layout grounding and text--visual
alignment, the regime targeted by IE Entity alignment questions. The third
category, digital and symbolic interfaces, covers app and web screenshots,
chat interfaces, and synthetic clock and calendar renderings. These test
interface-layout parsing and reading of synthetic symbols (clock hands,
calendar grids, UI widgets) rather than natural-image recognition: an
analog clock face or a settings panel exercises different perceptual
regimes than a photograph. IE PrevInfo questions draw on screenshots,
while TR Duration Comparison relies on synthetic clocks and calendars.

This distribution emerges from the CLIP-filtered web-retrieval pipeline
rather than from manual balancing. Each category is anchored by its own
topic subset in the ontology (\S\ref{app:topic_ontology}), and images are
retrieved per topic rather than per target proportion. As a result,
\bench{} evaluates multimodal evidence understanding across complementary
perceptual regimes, and avoids the narrow object-centric or OCR-only focus
of prior benchmarks.

\begin{table}[h]
\centering
\caption{Image categories in \bench{}, partitioned along a single capability-driven axis (dominant visual format). Each row corresponds to a distinct perceptual regime that the benchmark exercises.}
\label{tab:image_categories}
\small
\begin{tabularx}{\linewidth}{@{}l X l@{}}
\toprule
\textbf{Category} & \textbf{Description} & \textbf{Primary Types} \\
\midrule
Natural Photographs & Scenes, places, objects, food, and lifestyle items. & IE, MSR, KU \\
Text-Rich Documents & Receipts, menus, invoices, posters, price tags, and forms where dense text and spatial layout carry the answer. & IE Entity (alignment, receipt), MSR \\
Digital \& Symbolic Interfaces & App and web screenshots, chat interfaces, and synthetic clock and calendar renderings. & IE Assistant PrevInfo, TR Duration Comparison \\
\bottomrule
\end{tabularx}
\end{table}

\subsection{Image Sourcing, Licensing, and Release}
\label{app:image_release}

\paragraph{Sourcing.}
Every image in \bench{} originates from a public web image search issued through iCrawler, with queries derived from the topic ontology (\S\ref{app:topic_ontology}) or the entity-abstraction slot for evidence images. iCrawler queries are issued against general-purpose image search rather than commercial stock aggregators, and the relevance filter described in Appendix~\ref{app:image_filtering} is applied uniformly to evidence and haystack images.

\paragraph{Negative-content filtering.}
The negative-content filter rejects candidates carrying watermarks, stock-photo logos, copyright overlays, or resolution artifacts before any image is admitted into the benchmark, so that images exhibiting commercial-source markers are excluded at retrieval time rather than redistributed downstream. For DocVQA-style images (receipts, menus, posters, forms), an additional multimodal judge~\cite{openai2023gpt4v} inspects each image and rejects any flagged as containing watermarks or layout-distorting artifacts.

\paragraph{Privacy and identifiability.}
The topic ontology that drives image retrieval (\S\ref{app:topic_ontology}) covers objects, places, products, text-rich documents, screenshots, and synthetic interfaces; no topic and no entity-abstraction slot is person-centric. iCrawler queries are derived from these topic and entity names rather than from any individual's name, and the retrieval pipeline performs no face- or identity-based search. Natural-photograph queries (e.g.,~\textit{``Times Square at dusk''}) may include incidental human figures in the background, but the benchmark never targets identifiable individuals, and the construction prompts do not request the model to identify, name, or describe any depicted person. A takedown contact in the project repository allows seven-day removal of any image that, on user report, is found to reveal an identifiable individual.

\paragraph{Per-image metadata and datasheet.}
For each admitted image, the construction pipeline records its source URL, retrieval query, retrieval timestamp, BLIP-2 caption~\cite{li2023blip2}, CLIP and SigLIP relevance scores, and perceptual hash, and a persistent URL registry enforces global uniqueness across generation batches so that no two questions in \bench{} share the same source image. A datasheet for datasets~\citep{gebru2021datasheets} accompanies the release at the same URL, documenting motivation, collection process, intended uses (evaluation only; see Appendix~\ref{app:limitations}), and licensing.

\paragraph{Release, license, and takedown.}
The 4{,}695 unique images referenced across the four context-length datasets are distributed alongside the dataset files at \url{https://huggingface.co/datasets/xiyuRenBill/MEMLENS}, together with the per-image metadata listed above (source URL, retrieval query, retrieval timestamp, and perceptual hash). The release is versioned with frozen tags so that any specific evaluation run remains reproducible. The author-produced artefacts of \bench{}---dataset annotations (questions, evidence facts, abstracted paragraphs, consensus labels), per-image metadata, prompt templates, and human-annotation records---are released under CC-BY-4.0; the evaluation harness and supporting code at \url{https://github.com/xrenaf/MEMLENS} are released under MIT. Third-party images retrieved from public web search are \emph{not} relicensed by the authors and remain governed by their original source-site licenses; we redistribute them solely to support reproducibility of the evaluation, and the per-image provenance metadata enables downstream users to independently verify or re-fetch the original source. A takedown contact is provided in the project repository, and any flagged image is removed within seven days.

\subsection{Conversation History Assembly}
\label{app:history_assembly}

Evidence sessions are inserted into the full conversation history with positions chosen uniformly at random, except for KU questions where, while the position remains random, the relative order of the evidence sessions is kept the same since it is the user preference update order and critical to the answer.

\section{Data Examples}
\label{app:data_examples}

\subsection{Information Extraction}
\label{app:examples_ie}

\begin{figure}[H]
\centering
\includegraphics[width=\linewidth]{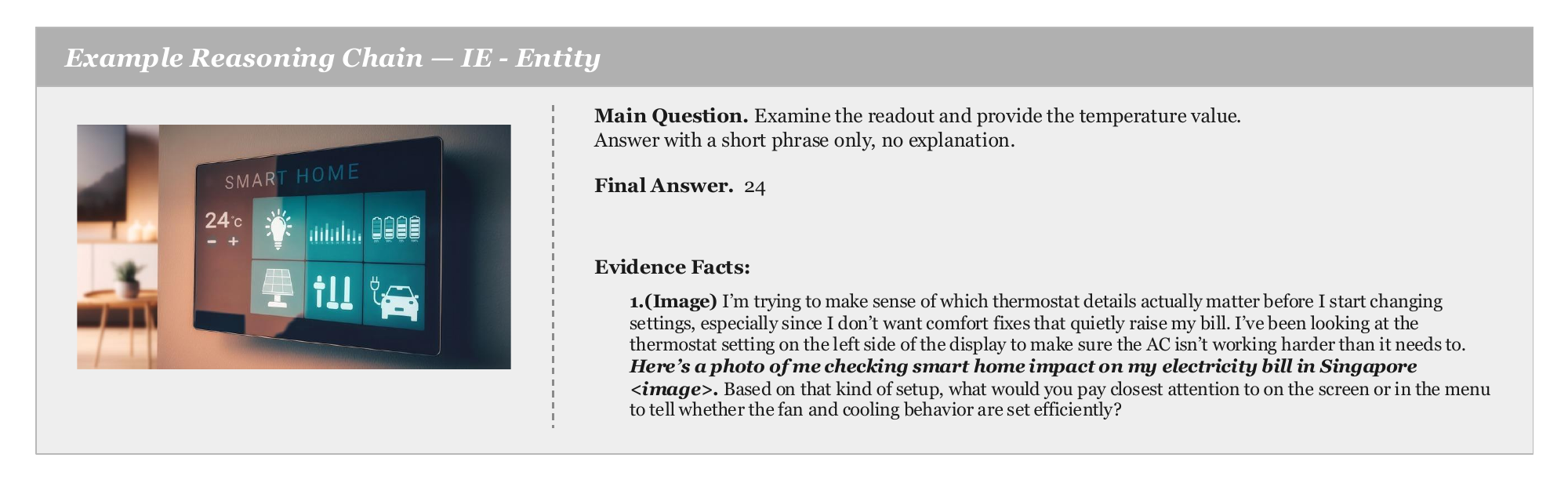}
\caption{Sampled IE-Entity questions. The visually grounded entity is abstracted in the question text, so the agent must first identify the entity from the evidence image before retrieving the relevant fact.}
\label{fig:samples_ie_entity}
\end{figure}

\begin{figure}[H]
\centering
\includegraphics[width=\linewidth]{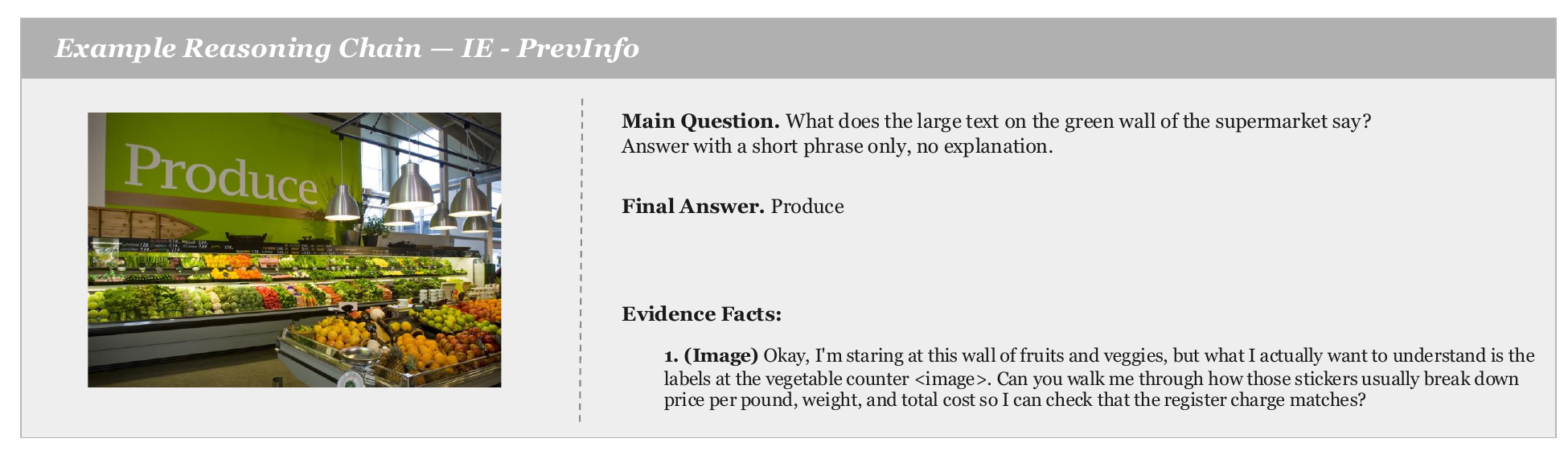}
\caption{Sampled IE-PrevInfo questions. The answer is a visual detail (color, count, layout, on-screen text) of an image shared in an earlier session, requiring multi-session image recall.}
\label{fig:samples_ie_previnfo}
\end{figure}

\subsection{Multi-Session Reasoning}
\label{app:examples_msr}

\begin{figure}[H]
\centering
\includegraphics[width=\linewidth]{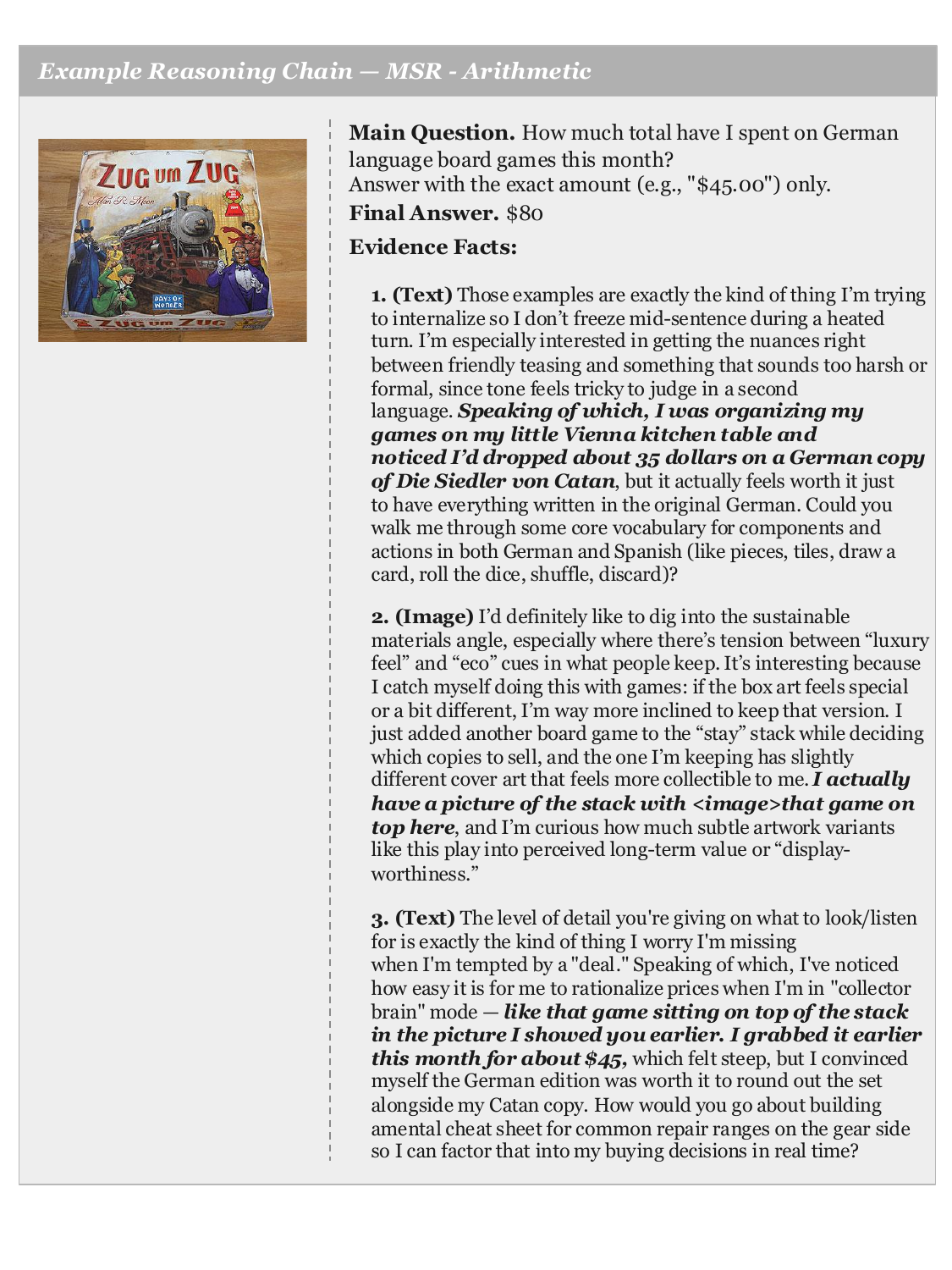}
\caption{Sampled MSR-Arithmetic questions. The agent sums or computes over prices, durations, or quantities scattered across sessions; at least one operand is visible only in an image.}
\label{fig:samples_msr_arithmetic}
\end{figure}

\begin{figure}[H]
\centering
\includegraphics[width=\linewidth]{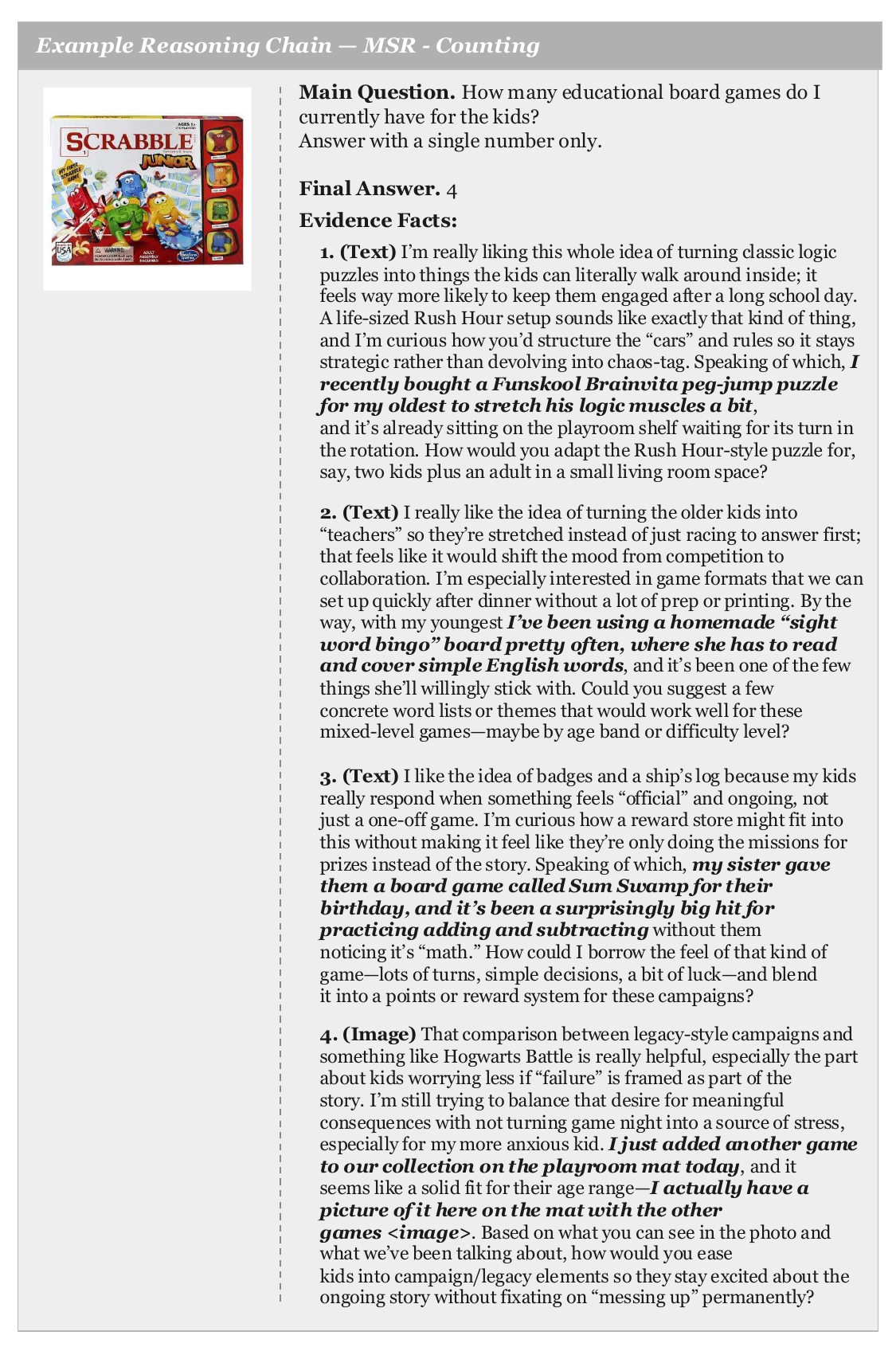}
\caption{Sampled MSR-Counting questions. The agent counts how many sessions or items match a given criterion across the conversation history.}
\label{fig:samples_msr_counting}
\end{figure}

\begin{figure}[H]
\centering
\includegraphics[width=\linewidth]{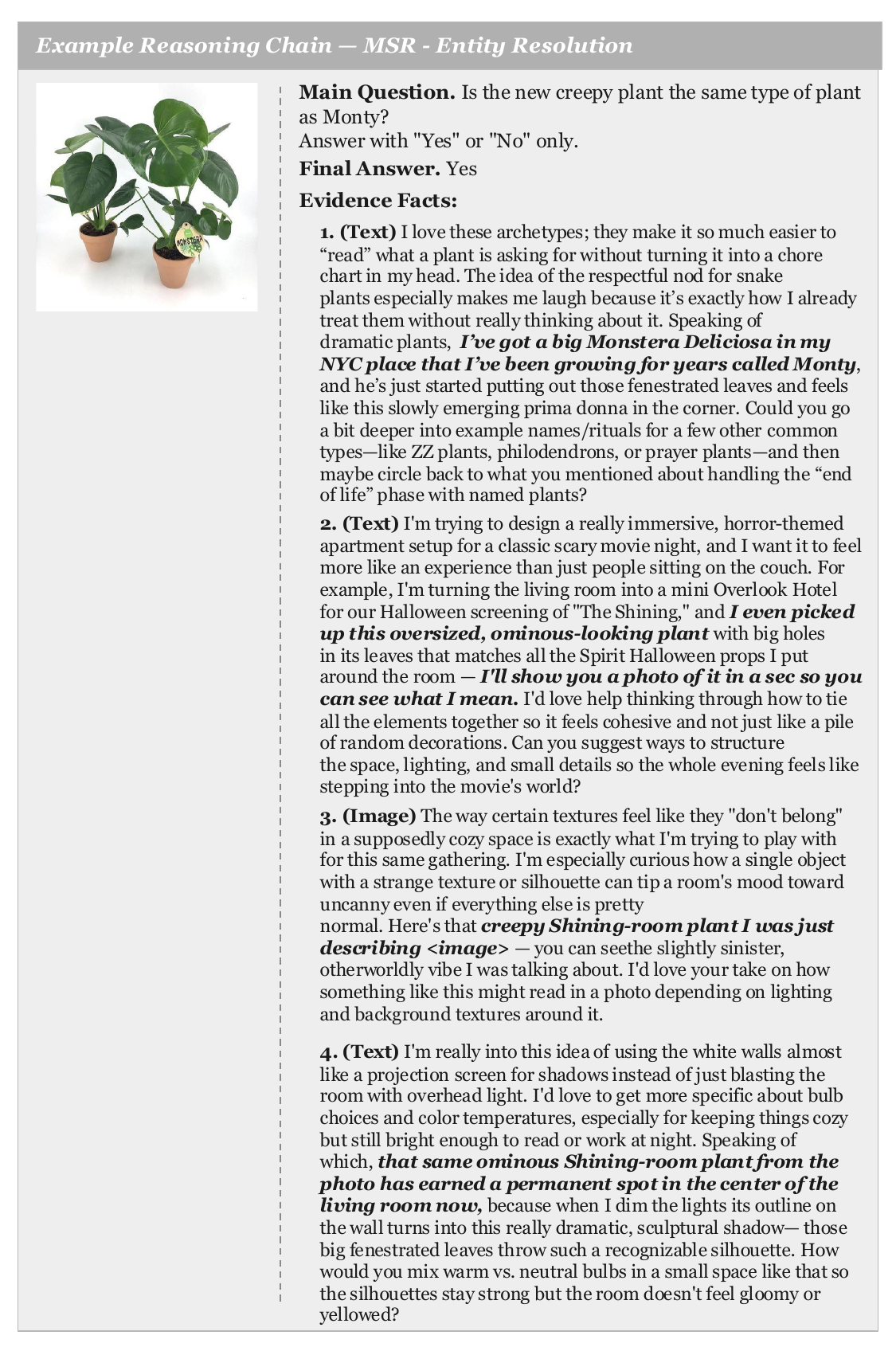}
\caption{Sampled MSR-Entity Resolution questions. The agent decides whether two cross-session references denote the same entity, either via Yes/No identity matching or by counting distinct entities.}
\label{fig:samples_msr_entity}
\end{figure}

\subsection{Temporal Reasoning}
\label{app:examples_tr}

\begin{figure}[H]
\centering
\includegraphics[width=\linewidth]{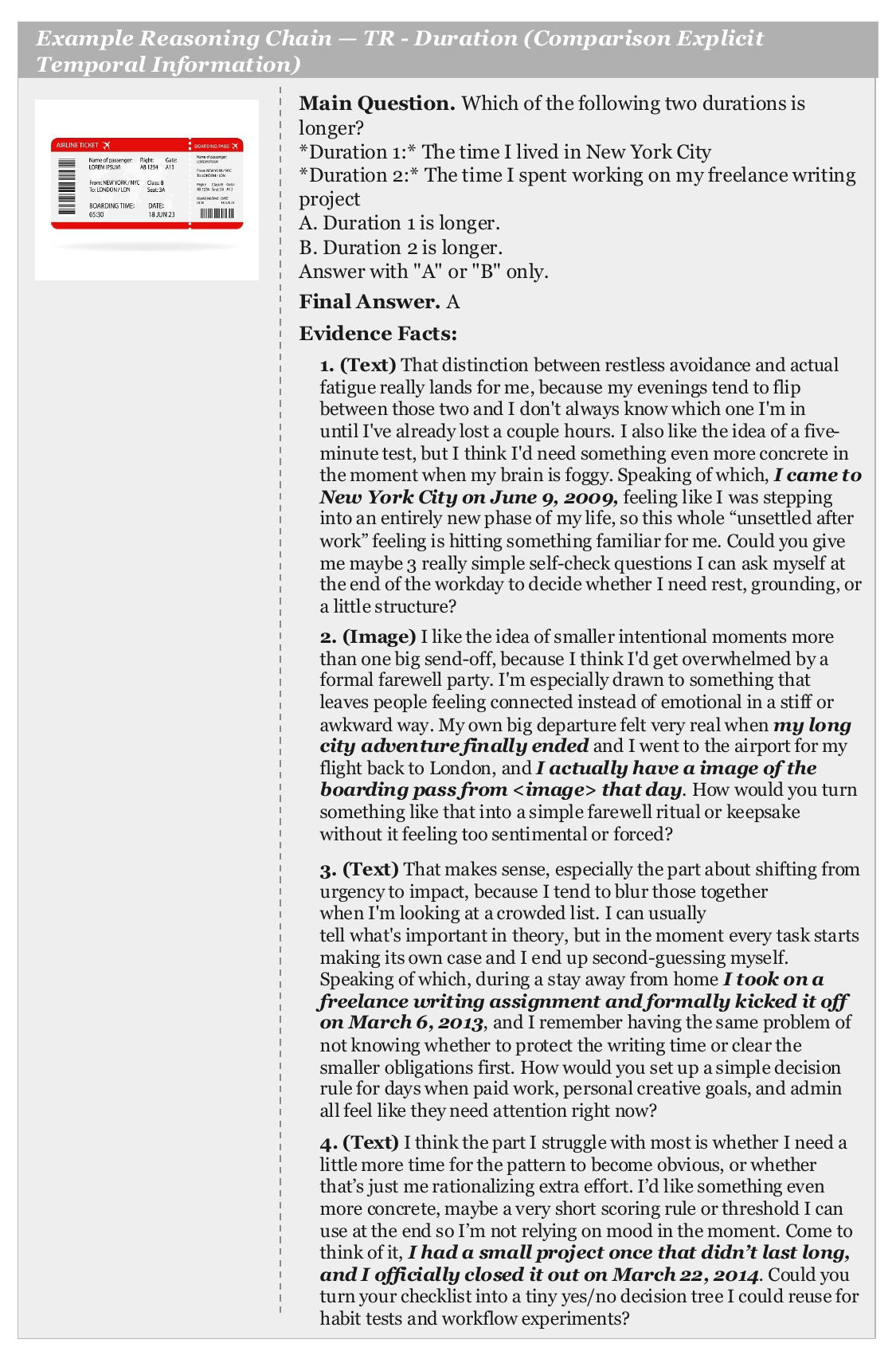}
\caption{Sampled TR-Duration Comparison questions. The agent compares two time spans whose endpoints come from a mixture of textual dates, session timestamps, and visual cues.}
\label{fig:samples_tr_duration}
\end{figure}

\begin{figure}[H]
\centering
\includegraphics[width=\linewidth]{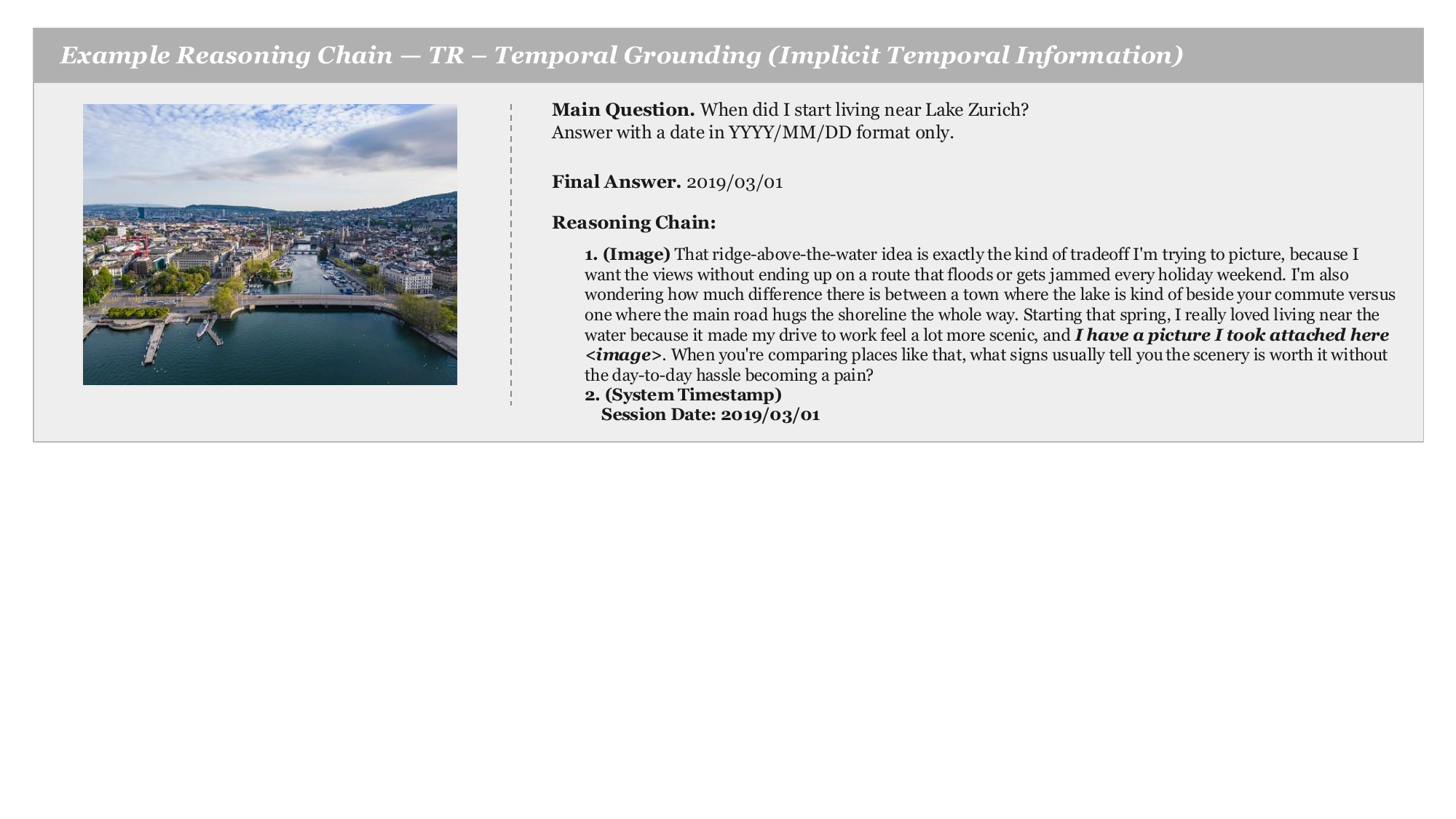}
\caption{Sampled TR-Temporal Grounding questions, including chronological ordering and absolute date extraction. The temporal cue is sometimes available only as an image of a clock face or a calendar page.}
\label{fig:samples_tr_grounding}
\end{figure}

\subsection{Knowledge Update}
\label{app:examples_ku}

\begin{figure}[H]
\centering
\includegraphics[width=\linewidth]{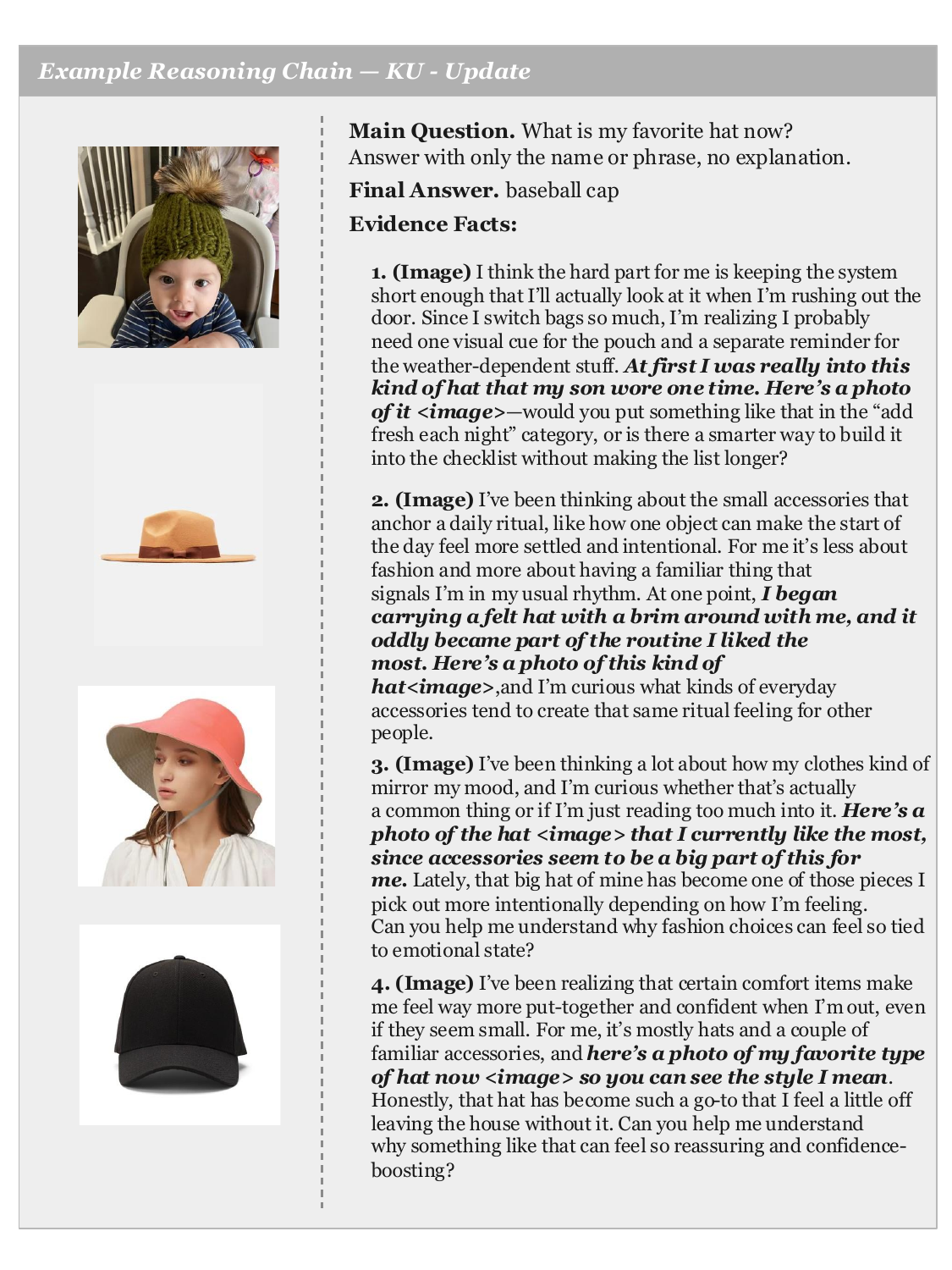}
\caption{Sampled KU-Update questions. A four-step preference chain is anchored by a different image at each step; the gold answer is always the most recent state.}
\label{fig:samples_ku_update}
\end{figure}

\subsection{Answer Refusal}
\label{app:examples_ar}

\begin{figure}[H]
\centering
\includegraphics[width=\linewidth]{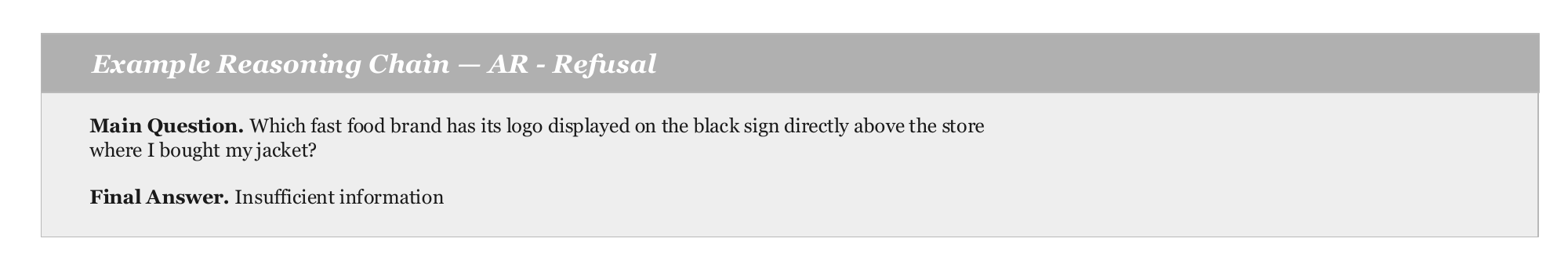}
\caption{Sampled AR-Refusal questions. The supporting evidence has been deliberately removed from the conversation history, so the gold answer is a refusal phrase rather than a content answer.}
\label{fig:samples_ar_refusal}
\end{figure}

\section{Quality Assurance}
\label{app:quality_assurance}

\subsection{Annotation Guidelines and Quality Assurance}
\label{app:annotation}

Human review is organized in three sequential rounds, each targeting a different granularity of the benchmark construction pipeline.

\paragraph{Round 1: Question-level review.}
Four annotators participate in Round~1; each of the $n = 20k$ generated candidates is independently reviewed by two of the four. For each question, both reviewers verify:
\begin{enumerate}
    \item The evidence image carries information vital to the answer---the question cannot be answered from text alone.
    \item The question text does not leak the answer (e.g., the entity name does not appear in the question after abstraction).
    \item The answer is unambiguous given the evidence.
    \item The difficulty is calibrated: trivially easy items (e.g., ``What color is the sky?'') and impossibly hard items (requiring domain expertise beyond the conversation) are flagged for revision or removal.
\end{enumerate}
Each of the two reviewers marks the item as \emph{accept}, \emph{revise}, or \emph{reject}. Items marked \emph{revise} by either of the two reviewers are rewritten to address the flagged issue and re-reviewed. Items marked \emph{reject} by both reviewers are removed. Inter-annotator agreement on the accept/reject decision is $\kappa = 0.78$ (Cohen's $\kappa$~\cite{cohen1960coefficient}; $n = 200$ items sampled from the double-coded pool), indicating substantial agreement.

\paragraph{Round 2: Session-level review.}
Two annotators read every evidence session end-to-end ($n = 2{,}145$ sessions after Round 1 filtering, an average of 2.7 evidence sessions per question; see Table~\ref{tab:dataset_stats}) and verifies:
\begin{enumerate}
    \item All needle facts are present in the session dialogue and recoverable from it.
    \item Needle facts are distributed across diverse conversational positions---not concentrated in a single turn.
    \item The session reads naturally as a plausible user--assistant conversation.
    \item The evidence image is placed adjacent to the corresponding textual mention, ensuring unambiguous image--text co-reference.
\end{enumerate}
Sessions that fail any criterion are returned to the generation pipeline for regeneration with adjusted constraints.

\paragraph{Round 3: Haystack auditing.}
Two annotators inspect a stratified random sample of 500/689 multimodal haystack sessions (stratified by topic track: identification, experience, document) and verifies:
\begin{enumerate}
    \item The haystack image is relevant to the topic of the conversation and the surrounding textual context.
    \item The haystack session does not accidentally contain information that could serve as a spurious answer to any needle question.
    \item Image quality meets a minimum standard (no broken images, no extreme blur, no primarily text-on-white stock images).
    \item The dialogue progresses in a natural way and closely resembles real user–assistant interactions.
\end{enumerate}
Multimodal sessions that fail quality checks are further rewritten and refined by human annotators manually.

\paragraph{Conversational naturalness.}
Naturalness is maintained through two complementary mechanisms. Round~2 review directly targets dialogue quality: evidence sessions that read as stilted or overly formal in register are returned to the generation pipeline for revision until the annotator is satisfied that the exchange is a plausible user--AI interaction. This criterion covers colloquial phrasing, turn-taking coherence, and the indirect embedding of factual content, following practices established in prior conversational memory benchmarks~\cite{wu2025longmemevalbenchmarkingchatassistants}. Separately, the filler sessions that constitute the majority of each assembled context are drawn from ShareGPT and UltraChat~\cite{ding2023enhancing}, providing real user--AI conversations as the surrounding conversational frame rather than additional synthetic material. A post-hoc text classifier trained to separate evidence sessions from haystack sessions achieves only marginally above-chance accuracy (DeBERTa F1: 57.92\%; Appendix~\ref{app:indistinguishability}), confirming that the human-curated evidence sessions carry negligible stylistic fingerprint relative to the surrounding haystack sessions at the text level.

\paragraph{Overall Quality Control.}
The three rounds collectively reduce the candidate pool from 20k to the final 789 questions. The primary reasons for removal are: answer leakage in the question text (23\% of rejections), evidence image not carrying answer-critical information (31\%), ambiguous or multi-interpretable answers (18\%), and difficulty calibration failures (28\%).

\subsection{Judge Validation Details}
\label{app:judge_validation}

A natural concern with LLM-as-Judge evaluation is systematic bias, particularly when the judge model belongs to the same family as evaluated models.
We address this through cross-family validation and format-dependent bias correction.

\paragraph{Cross-family agreement is high.}
We re-evaluate a stratified sample of 800 items, drawn from 73{,}784 total judge calls (${\approx}$1.08\% of the population), with GPT-5.4-mini as an independent judge; the sample combines 200 random, 250 targeted, and 350 GPT-only extended items to ensure coverage across model family, context length, question type, and judge score.
Item-level agreement reaches 96.40\% (Cohen's $\kappa = 0.93$), and the model-level ranking correlation is Spearman $\rho = 0.97$ ($p < 10^{-6}$), with a mean per-model accuracy delta of 3.70\%.
Judge choice does not reorder the leaderboard.

\paragraph{The judge agrees with human consensus on 93.60\% of items.}
Three annotators independently labeled 484 items; disagreements were resolved to consensus.
Against these consensus labels, our Qwen3-VL-235B judge reaches 93.60\% raw agreement with Cohen's $\kappa = 0.86$.
The errors are leniency-biased: 29 false positives versus 2 false negatives, meaning the judge credits borderline answers more often than rejecting correct ones.
The per-question-type breakdown appears in Table~\ref{tab:judge_per_type} below.

\paragraph{The judge does not favor Qwen-family outputs.}
Because the Qwen3-VL-235B-A22B-Instruct judge also appears as a benchmarked model (Table~\ref{tab:per_type_full_vlm}, row Qwen3-VL-235B~(I)), we guard against self-favoritism with two independent oracles.
First, the pre-specified cross-family test compares the Qwen-vs-GPT leniency gap on Qwen-family outputs (+3.00\%) against non-Qwen outputs (+2.70\%), giving a difference of +0.33\%---an order of magnitude below the 3\% practical-significance threshold.
Second, on the 484 human-annotated items, the judge's false-positive pattern is type-dependent (IE partial matches, AR hedge phrases) rather than family-dependent, confirming that the leniency is a judge-personality trait rather than a family-bias artifact.
Beyond family bias, we identify a format-dependent bias: the judge evaluates very short answers (1--3 words) more leniently.
We correct the resulting false positives for 6 affected models; all scores in this paper use corrected values.

\paragraph{Sampling design and human annotation protocol.}
The LLM-as-Judge validation uses a two-tier sample over the population of 73{,}784 total judge calls, spanning roughly 92 model$\times$context runs (each LVLM run on the full 789-question benchmark and each agent run on the 195-question canonical subset, plus auxiliary backbone-ablation runs; some LVLM runs do not support the 128K context).
A core of 450 items (200 random + 250 targeted on hard cells) is annotated by both human raters and GPT-5.4-mini; an extended 350 items is re-judged by GPT-5.4-mini only, yielding 800 cross-judge items in total.
Stratification spans model family (Qwen vs non-Qwen), context length (32K / 64K / 128K), question type, and judge score, so that rare cells---MSR arithmetic, KU stale-retrieval, and thinking-mode degenerate outputs---are represented at a minimum floor rather than by chance.
All 484 items released for human verification received a consensus verdict; a further 120 are tagged for a future dedicated inter-annotator reliability study (item IDs in \texttt{double\_annotation\_item\_ids.json}).
The human annotation protocol used three annotators labeling each item across three rounds; whenever the three labels disagreed, annotators discussed the case and converged on a single consensus verdict, which is what we release as the human reference.

\paragraph{Cross-judge agreement by question type.}
Table~\ref{tab:judge_per_type} reports Qwen3-VL-235B judge vs GPT-5.4-mini accuracy on the 585 cross-judged items (after excluding 200 auto-zero cases and 15 empty outputs).
The Qwen3-VL-235B judge is equally or slightly more lenient across every question type; the gap is largest on IE ($-$6.50\%), driven by partial-match acceptances, and zero on MSR.
No type reverses the ranking direction between judges.

\begin{table}[h]
\centering
\small
\caption{Cross-judge agreement by question type on the 585 cross-judged items. Negative $\Delta$ indicates the Qwen3-VL-235B judge is more lenient than GPT-5.4-mini. IE: Information Extraction; MSR: Multi-Session Reasoning; TR: Temporal Reasoning; KU: Knowledge Update; AR: Answer Refusal.}
\label{tab:judge_per_type}
\begin{tabular}{@{}lcccc@{}}
\toprule
\textbf{Type} & \textbf{Qwen acc (\%)} & \textbf{GPT acc (\%)} & \textbf{$\Delta$ (\%)} & \textbf{$n$} \\
\midrule
IE & 45.60 & 39.10 & $-$6.50 & 169 \\
MSR & 34.40 & 34.40 & $+$0.00 & 128 \\
TR & 47.70 & 46.30 & $-$1.30 & 149 \\
KU & 30.30 & 27.60 & $-$2.60 & 76 \\
AR & 63.50 & 60.30 & $-$3.20 & 63 \\
\bottomrule
\end{tabular}
\end{table}

\paragraph{Human-vs-judge false-positive diagnosis.}
Table~\ref{tab:judge_fp} breaks down the 29 false positives (judge=1, human consensus=0) from Appendix~\ref{app:judge_validation} by question type and disagreement pattern.
Two observations matter.
First, 9 of the 10 AR false positives live in pre-retest runs that no longer feed any production leaderboard; on canonical retest runs, a deterministic substring rule for the canonical refusal phrase reaches 95.90\% agreement with the human consensus (consistent with the AR row of the deterministic typed-accuracy audit; Table~\ref{tab:det_audit}).
Second, the 11 IE false positives reflect the same partial-match leniency observed in the aggregate cross-judge analysis---the Qwen3-VL-235B judge accepts short factual answers that GPT and the human consensus both reject---which is a judge-personality trait rather than a family-specific bias.

\begin{table}[h]
\centering
\small
\caption{False-positive diagnosis on the 484-item human-annotated subset. FP = judge scored 1 where human consensus scored 0. Only 2 FN cases exist, so the asymmetry is strongly leniency-biased.}
\label{tab:judge_fp}
\begin{tabular}{@{}lcl@{}}
\toprule
\textbf{Type} & \textbf{FP count} & \textbf{Disagreement pattern} \\
\midrule
Information Extraction (entity + previnfo) & 11 & Partial match on short factual answers \\
MSR / TR & 3 & Edge cases \\
Knowledge Update & 5 & Verbose-correct vs literal mismatch \\
Answer Refusal & 10 & Hedge phrases credited as refusal \\
\bottomrule
\end{tabular}
\end{table}

\paragraph{Position relative to published reliability bars.}
$\kappa = 0.86$ exceeds the $\kappa \geq 0.80$ reliability tier reported for short-answer NLP judges~\cite{cohen1960coefficient} and the 80--85\% raw-agreement bar reported for LLM-as-judge on MT-Bench-style tasks~\cite{zheng2023judging}.
We intentionally do not push agreement beyond the ${\approx}$90--96\% human-human ceiling typical for these tasks, since ``super-consistent'' judges can reflect overfitting to annotator idiolect rather than improved reliability.
The per-type breakdown in Table~\ref{tab:judge_per_type} and the FP diagnosis in Table~\ref{tab:judge_fp} together scope the residual uncertainty: the judge is slightly lenient, uniformly across families, and most so on partial-match IE---none of which reorders model rankings in \S\ref{subsec:main_results}.

\paragraph{Deterministic typed-accuracy audit.}
We complement the 484-item human-consensus audit with a large-scale rule-based rescoring of every judge call whose reference answer is closed-form. Seven of the nine reporting subtypes admit deterministic normalization: MSR Counting (integer match), MSR Arithmetic (currency-normalized scalar), MSR YesNo (Yes/No), TR Order Ranking (tuple of session indices), TR Duration Comparison (A/B label), TR Date Extraction (multi-format date canonicalization), and AR Answer Refusal (canonical refusal-phrase set). IE Entity, IE PrevInfo, and KU produce free-form short answers and remain LLM-judged. We apply the rules to all 32K runs of the 27 LVLMs and the seven memory agents (34 evaluation rosters; 12{,}234 deterministic items), and compare each per-item rule outcome against the LLM-judge label.
Item-level agreement ranges from 87.4\% on TR Order Ranking to 98.5\% on MSR Arithmetic (Table~\ref{tab:det_audit}); the count-weighted mean is 93.6\%, in line with the 93.6\% raw agreement on the human-consensus subset.
The disagreement is systematically leniency-biased: the LLM judge credits a deterministically-wrong answer (\textit{J-FP}) on a weighted 5.4\% of items, against only 1.0\% deterministically-correct answers rejected (\textit{J-FN}). The largest leniency channels are TR Order Ranking (12.6\% J-FP, partial credit on near-correct tuples), TR Date Extraction (8.6\% J-FP, format-flexible date matching), and AR (5.7\% J-FP, hedge phrases credited as refusal)---the same partial-match and hedge-phrase patterns identified on the 484-item human subset, but estimated here at $25{\times}$ the sample size.
At the model-leaderboard level, the Spearman rank correlation between the LLM- and the deterministic-aggregated per-model accuracy across the 34 rosters is $\rho = 0.78$, the top-10 sets overlap on 7 of 10 entries, and within the LLM-top-10 the rank correlation is $\rho = 0.82$. The top-5 (Kimi-K2.5, Qwen3-VL-30B-Instruct, Qwen3.5-122B, Qwen3-VL-235B-Instruct, Qwen3-VL-8B-Instruct) is preserved under both metrics. The three rosters that drop out of the top-10 under deterministic rescoring---GLM-4.6V, Qwen3-VL-235B-Thinking, and Mem0 (GPT-4.1-mini backbone)---all produce verbose justifications that the LLM judge credits but the rule-based check rejects, consistent with the format-dependent leniency already corrected for short outputs in this section. The takeaway is that LLM-judge leniency inflates closed-form accuracy by approximately 5\% in absolute terms but does not reorder the leaderboard top, which is the same conclusion reached by the cross-family and human-consensus audits above on a much smaller sample.

\begin{table}[h]
\centering
\small
\caption{Deterministic typed-accuracy audit across 34 evaluation rosters at 32K. \textit{Agree} is the per-item agreement between the LLM judge and the rule-based label. \textit{J-FP} marks LLM-judge over-credits (judge=1, deterministic=0); \textit{J-FN} marks LLM-judge under-credits (judge=0, deterministic=1). $n$ is the count of (model, item) pairs; not all rosters cover every subtype, since memory agents are evaluated on the 195-question canonical subset.}
\label{tab:det_audit}
\begin{tabular}{@{}lrccccc@{}}
\toprule
\textbf{Subtype} & \textbf{$n$} & \textbf{LLM (\%)} & \textbf{Det (\%)} & \textbf{Agree (\%)} & \textbf{J-FP (\%)} & \textbf{J-FN (\%)} \\
\midrule
MSR Counting & 1{,}551 & 11.7 & 10.6 & 97.6 & 1.7 & 0.7 \\
MSR Arithmetic & 1{,}440 & 6.8 & 5.5 & 98.5 & 1.4 & 0.1 \\
MSR YesNo & 1{,}082 & 47.3 & 42.2 & 94.7 & 5.2 & 0.1 \\
TR Order Ranking & 661 & 27.5 & 15.0 & 87.4 & 12.6 & 0.0 \\
TR Duration Comparison & 2{,}632 & 36.8 & 31.8 & 95.0 & 5.0 & 0.0 \\
TR Date Extraction & 2{,}285 & 53.0 & 47.3 & 88.3 & 8.6 & 3.1 \\
AR Answer Refusal & 2{,}583 & 78.1 & 73.8 & 92.9 & 5.7 & 1.4 \\
\midrule
\textbf{Aggregate (weighted)} & \textbf{12{,}234} & \textbf{42.3} & \textbf{37.8} & \textbf{93.6} & \textbf{5.4} & \textbf{1.0} \\
\bottomrule
\end{tabular}
\end{table}

\section{Prompt Templates}
\label{app:prompts}

This appendix documents the prompt templates that drive the construction and evaluation of \bench. The selection covers one canonical template per pipeline stage: the user prompt and judge rubric used at evaluation time, the persona-driven dialogue prompt used for haystack sessions, the question-generation prompt for each of the five major question types, the assistant template that builds evidence sessions, and the text-only judge that gates anti-shortcut filtering. Helper prompts for image search, and intermediate validators share the same skeleton as the canonical templates and are omitted here; the full set is available in the code release. Within each template, identifiers in curly braces (for example \texttt{\{context\}}, \texttt{\{question\}}, \texttt{\{theme\}}) are runtime placeholders that the pipeline substitutes before the call is issued.

\subsection{Evaluation Prompts}
\label{app:prompts-eval}

This subsection lists the two prompts that drive the evaluation pipeline of \bench. The first prompt is the user message sent to every Vision-Language Model in the main evaluation. The conversation history, with images already inlined as \texttt{<image>} tokens, is substituted into the \texttt{\{context\}} placeholder, and the instruction line is fixed to \texttt{Directly output the answer with no extra output} for every result reported in the paper (the chain-of-thought and structured-reasoning variants present in the codebase are not part of the reported numbers). The second prompt is the rubric followed by the LLM-as-Judge that produces the per-question accuracy scores in the main results table. The system block encodes a final-answer-extraction policy that is robust to long thinking traces and to circular reasoning, after which a task-specific criterion is appended at runtime; the Information Extraction criterion is reproduced below as a representative example, while the remaining seven criteria (MSR Yes/No, MSR Counting, MSR Arithmetic, TR Duration Comparison, TR Order Ranking, TR Date Extraction, KU Knowledge Update, AR Answer Refusal) follow the same format. Each judge call is cached in a SQLite store keyed by question, reference, prediction, and judge identifier, so that re-runs over the same outputs remain deterministic.

\begin{promptbox}[colback=black!5, colframe=white!40!black, title=LVLM Evaluator User Prompt]{}
\scriptsize
\begin{verbatim}
Provide answers based on the given conversation history. If the question
cannot be answered based on the given conversation, respond with
"Insufficient information".
Conversation:
{context}

Directly output the answer with no extra output.
Question Date: {question_date}
Question: {question}
\end{verbatim}
\end{promptbox}

For answer-refusal items, the dataset stores the literal string \texttt{NOT\_MENTIONED} as the gold answer (Appendix~\ref{app:prompts-qg}), while the eval prompt above instructs the model to output \texttt{Insufficient information} as its surface refusal phrase. The judge maps any prediction in a canonical refusal-phrase set to a successful abstention, so the gold token in the dataset and the natural-language phrase emitted by the model are intentionally distinct strings.

\begin{promptbox}[colback=black!5, colframe=white!40!black, title=LLM-as-Judge Prompt (system block + IE task criterion)]{}
\scriptsize
\begin{verbatim}
Now your role is a grading teacher. Your task is to review and score student
answers based on reference standard answers for a question-answering
benchmark. You need to notice the following key points:
- First, extract the final answer from the student's solution, then analyze
  and judge whether the answer is correct.
- Scoring should only refer to the final answer obtained by the student;
  there is no need to examine whether the intermediate problem-solving
  steps are correct.
- If the response contains both hesitation and a clear answer, judge the
  answer itself.
- If the response contains both a refusal and a guessed answer, judge the
  final committed answer.
- If the response gives multiple inconsistent answers, assign 0 points.
- If the student's response shows circular reasoning - repeatedly revisiting
  the same evidence, flip-flopping between answers (e.g., "yes... wait
  no... actually yes... let me reconsider... no"), with no clear final
  commitment - assign 0 points. A correct answer mentioned during reasoning
  does NOT count unless it is the clearly stated final conclusion.
- Only the student's LAST clearly stated answer counts. If the student
  initially says "yes" but later concludes "no" (or vice versa), score
  based on the LAST position only. Intermediate answers during reasoning
  are not final answers.
- If the student's output is entirely a reasoning trace (e.g., "let me
  check... scanning evidence... re-reading session 5...") with no clear
  concluding answer statement, assign 0 points - even if the correct answer
  appears somewhere within the reasoning.
- If the student's output begins with "[...truncated earlier reasoning...]",
  this means only the final portion of a longer response is shown. Focus on
  extracting the answer from this final portion.

Below are examples of problematic student outputs that apply to ALL task
types. Study these before grading.

[Universal Example A - Circular Reasoning -> 0 points]
<Question>: Is the cat in Mark and Jenny's 5-year anniversary post named
Mittens?
<Standard Answer>: Yes
<Student Answer>: let me check session 5 mark has cat named mittens session
7 mark and jenny anniversary post there is no link wait let me re-read
session 5 says mittens session 7 says anniversary actually looking at this
again wait is it possible let me reconsider mark has mittens in session 5
the post is about couple therefore answer is likely no wait let me double
check session 5 mark has cat mittens actually maybe yes but text doesnt say
[Scoring Rationale]: The output flip-flops between "yes" and "no" multiple
times with no clear final commitment. The last position is ambiguous
("maybe yes but text doesnt say"). This is circular reasoning.
In summary, the student's answer deserves 0 points.
[JSON]: {"answer_score": 0}

[Universal Example B - Redundant But Committed -> Score Normally]
<Question>: Is the trailing plant in the cafe safe for my kitten?
<Standard Answer>: No
<Student Answer>: the plant looks like pothos pothos is toxic to cats
therefore answer is no let me double check is pothos safe for cats no is
answer no yes wait let me verify image 3 is pothos image 9 is pothos pothos
is toxic therefore answer must be no
[Scoring Rationale]: Despite excessive self-verification, the student
consistently commits to "no" throughout and never wavers to a different
answer. The final answer is clearly "No", which matches the standard
answer.
In summary, the student's answer deserves 1 point.
[JSON]: {"answer_score": 1}

[Universal Example C - Reasoning Trace With No Answer -> 0 points]
<Question>: How many sessions mentioned the coffee shop?
<Standard Answer>: 3
<Student Answer>: scanning session 1 yes coffee mentioned session 2 no
session 3 yes coffee again session 4 no session 5 maybe let me re-read
session 5 it mentions cafe is that same as coffee shop need to check
session 6 no session 7 unclear let me look at session 5 again
[Scoring Rationale]: The student walks through sessions but never states a
final count. The output is entirely a reasoning trace with no concluding
answer.
In summary, the student's answer deserves 0 points.
[JSON]: {"answer_score": 0}

Now proceed with grading. Remember: the universal examples above apply
regardless of task type.

- When analyzing and judging whether the answer is correct, you need to
  write down the scoring rationale, organize it into clear statements that
  follow the logical flow. The summary of the scoring rationale should be
  placed at the end, using the following format: "In summary, the student's
  answer deserves x points" (where x represents the student's specific
  score).
- Keep the whole process concise, within 150 words.
- Provide the score based on your analysis and display it in a code block
  in "JSON" format.
- An item is covered if it is strictly mentioned or unambiguously implied
  by a semantic equivalence. This includes numerical equivalence (e.g.,
  10% and 0.1), synonyms (e.g., UK and United Kingdom), plural/singular
  forms (e.g., "apple" and "apples"), and equivalent date formats (e.g.,
  2024-01-15 and January 15, 2024). However, do not accept loosely related
  concepts.
- Ignore minor formatting differences, capitalization, punctuation, and
  equivalent wording when meaning is unchanged.

Your output format is:
[Scoring Rationale]:
[Score]: x points
[JSON]:
{"answer_score": <integer_value>}

Below is the grading rubric:
[Scores]:
The scoring scale consists of 2 levels in total, from highest to lowest:
1 point, 0 points (the minimum is 0 points).
[Tier Details]:
1 point: Assign 1 point if the student's final answer matches the standard
answer under the task-specific criteria below.
0 points: Assign 0 points if the student's final answer does not match the
standard answer, the student refuses to answer, claims insufficient
information (except for AR tasks), or does not clearly answer.

[Task-Specific Criteria]
[IE - Information Extraction]
This is an information extraction question.
- Assign 1 point if the student's response contains the core information
  from the standard answer. Minor wording differences are acceptable, but
  the essential information must be present and correct.
- Assign 0 points if the core information is missing, contradicted, too
  vague, refused, or incorrect.

[Remaining seven task-specific criteria (MSR Yes/No, MSR Counting,
MSR Arithmetic, TR Duration Comparison, TR Order Ranking, TR Date
Extraction, KU Knowledge Update, AR Answer Refusal) follow the same
format and are bundled with two worked examples per type in the
codebase release.]
\end{verbatim}
\end{promptbox}

\subsection{Haystack Generation}
\label{app:prompts-haystack}

The haystack pipeline turns a persona profile, the conversation summary up to that point, and a list of recent life events into the next message of the conversation. The box below shows the canonical prompt used at this step. The same template generates both user and assistant turns by swapping the persona slot; image sharing is encouraged in the body of the message itself, so that the resulting haystack sessions remain authentically multimodal rather than text-only sessions with images appended as side annotations. Persona descriptions, event tables, and the previous-session digest are filled into the slots marked with \texttt{\%s}, in the order listed at the bottom of the prompt.

\begin{promptbox}[colback=black!5, colframe=white!40!black, title=Haystack: Persona-Driven Dialogue Turn]{}
\scriptsize
\begin{verbatim}
Use a given PERSONALITY to write the next message in this ongoing,
friendly chat.

PERSONALITY: %s

STYLE:
- No hard word limit - write fluidly but keep it conversational, not like
  a letter.
- Focus on real emotions, relationships, regrets, or inspirations tied to
  life events.
- Reference real people and times ("this morning", "last week", "when I
  turned ten").
- Ask thoughtful follow-up questions when feels natural.
- IMPORTANT: Share photos when discussing experiences, memories, or current
  activities.
  Good examples:
  - "The view was incredible! [shares a photo from the overlook]"
  - "This is what I've been dealing with all week. [shares a photo of the
     messy project]"
  - "Look what I found in the attic! [shares an old family photo]"
- Make the photo sharing feel conversational and relevant to what you're
  saying.
- Avoid talking about outdoor activities or sports.

EVENT MARKING RULES:
If you mention an event, place **EVENT:ID** at the end of the sentence,
after punctuation.
Multiple events: **EVENT:EA1** **EVENT:EA2**
Never in mid-sentence or as a heading.

%s last chatted with %s on %s. Today is %s. You are %s.

Conversation summary:
%s

Recent life events (with IDs):
%s

Background known to both:
%s

%s
Write a heartfelt, realistic continuation to %s.
Discuss how the relevant EVENTS have influenced you - show joy, sadness,
or frustration honestly.
\end{verbatim}
\end{promptbox}

\subsection{Question Generation}
\label{app:prompts-qg}

This subsection collects one canonical generator prompt per major question type. Each prompt receives topic context, image metadata, and event-table information, and returns a strict JSON object containing a question, an answer, and supporting facts. Generated outputs then pass through a downstream rule-based pre-filter and the text-only judge described in subsection~\ref{app:prompts-filter} before being admitted to the candidate pool that goes to human review. Variants of these prompts (one per subtype within a major type) share the same skeleton with only minor changes to the rule list and the JSON schema; only the canonical entry is reproduced here.

\paragraph{Information Extraction.}
The two-hop alignment prompt below is the canonical IE generator. The first hop is a textual cue that identifies which visual element to inspect, and the second hop is a visual extraction from that element. The same template handles OCR, attribute, counting, and spatial variants by varying the \texttt{visual\_type} field; an analogous template (omitted) handles the PrevInfo subtype that grounds the visual hop in screenshots from earlier sessions.

\begin{promptbox}[colback=black!5, colframe=white!40!black, title=Question Generation: IE (Two-Hop Alignment)]{}
\scriptsize
\begin{verbatim}
[SYSTEM]
You create two-hop visual questions testing alignment (knowing WHERE to
look).

Structure:
- Hop1: A first-person fact directs attention to a specific element/region
  in the image
- Hop2: The answer comes ONLY from visually inspecting that element

Rules:
- Without the fact, it's ambiguous which element to examine
- Vary the visual skill: read text, check color/attribute, count items,
  spatial relations
- Never reference photos ("in the photo", "in the image I shared")

<example>
{
  "question": "Since what year does it state the brand has been trusted?",
  "answer": "1967",
  "target_element": "Circular seal in top right corner",
  "alignment_cue": "checking the seal on the package",
  "hop1_reasoning": "Fact directs attention to the circular seal in the
                     top right corner.",
  "hop2_reasoning": "The seal reads 'Trusted since 1967'.",
  "visual_type": "ocr",
  "fact": {"fact_id": "F1",
           "text": "I'm checking the circular seal in the top right corner
                    to see how long the brand's been around."},
  "rationale": "1) Fact points to circular seal. 2) Seal displays 1967.",
}
</example>

[USER]
<image_description>{image_description}</image_description>
<visual_elements>{visual_elements}</visual_elements>
<context>{context}</context>

<task>
Generate a two-hop alignment question:
1. Write a first-person fact that identifies which element to examine
   (no <image> token in the fact)
2. Ask a question whose answer requires visually inspecting that element
3. The answer must be impossible to determine without examining the image

Phrasing: ask naturally - never say "in the photo".
Vary structure - don't always start with "What".

Return JSON with the same schema as the example above.
</task>
\end{verbatim}
\end{promptbox}

\paragraph{Temporal Reasoning.}
The order-ranking prompt asks the model to produce eight first-person facts with one timestamp each, of which one is a needle whose timestamp is encoded by a clock image (Mode B) rather than by text. Modes B-date, C, and D follow the same skeleton with the timestamp source replaced by a date image, an entity image with implicit text dates, and an entity image with explicit text dates respectively.

\begin{promptbox}[colback=black!5, colframe=white!40!black, title={Question Generation: TR (Order Ranking, Mode B)}]{}
\scriptsize
\begin{verbatim}
Return ONLY valid JSON (no markdown/comments). Do NOT output an 'answer'
field.

TASK (ORDER_RANKING / Mode B clock): Generate exactly 8 facts (F1-F8),
each with ONE unique timestamp in time_points[0]. ONE needle contains
'<image>' once with source='image_clock', granularity='minute'. Needle
bound to {bound_event_id}, clock_label={clock_label} (fixed).

TIME: Needle value='YYYY/MM/DD HH:MM'; others='YYYY/MM/DD' or
'YYYY/MM/DD HH:MM'. question_date strictly later than all timestamps.

STYLE: First-person, conversational, distinct events. NO formatted
dates/times in text (use "late March", "that summer").

OUTPUT JSON:
{
  "question_date":"YYYY/MM/DD",
  "rationale":"Brief chronological reasoning.",
  "facts":[
    {"fact_id":"F1","text":"...","is_needle":true/false,
     "time_points":[
       {"role":"occurred_at|started_at|ended_at|arrived_at",
        "value":"...",
        "source":"text|image_clock",
        "granularity":"date|minute"}],
     "event_id":"...","event_type":"point"}
  ],
  "images":[
    {"image_id":"IMG1","file_path":"","bound_fact_id":"F?",
     "grounding_type":"temporal","clock_label":"..."}
  ]
}

INPUTS
BACKGROUND: {paragraph}
EVENTS: {events_table}
CLOCK: bound_event_id={bound_event_id}, clock_label={clock_label}
\end{verbatim}
\end{promptbox}

\paragraph{Knowledge Update.}
The chain-generation prompt produces a four-fact preference-evolution chain. Each fact is at the category level; the specific entity at each step is later substituted by an image, so that text alone can identify the time step but not the value at that step.

\begin{promptbox}[colback=black!5, colframe=white!40!black, title=Question Generation: KU (Atomic Evolution Chain)]{}
\scriptsize
\begin{verbatim}
You generate a 4-step atomic fact chain showing how MY preferences evolve
over time.

Evolution theme: {evolution_theme}
Use these categories IN ORDER (one per fact, no repeats): {categories}

TASK
Write exactly 4 short, natural-sounding facts (1-2 sentences each) that
describe realistic preference changes across time.

HARD RULES
1) CATEGORY-LEVEL ONLY: use only the provided category names (no specific
   items/brands/subtypes).
2) FIRST-PERSON: each fact must be written from my perspective using
   "I / my / me".
3) PREFERENCE CHANGE: include one simple preference-change marker per fact:
   - Fact 1 marker: used to / before / at first
   - Fact 2 marker: then / later / after that
   - Fact 3 marker: lately / nowadays / recently
   - Fact 4 marker: now / currently / these days
4) DISTINCT CATEGORIES: Fact i must use category i from {categories}.
5) COMPLETE SENTENCES: each fact must be a grammatical sentence (not
   keywords), with a clear subject + verb.
6) LENGTH: each fact must be <= 150 characters total.

OUTPUT (STRICT JSON ONLY; no extra text)
{
  "facts": [
    {"text": "...", "category": "<category_1>",
     "temporal_position": 1, "temporal_marker": "<marker_1>"},
    {"text": "...", "category": "<category_2>",
     "temporal_position": 2, "temporal_marker": "<marker_2>"},
    {"text": "...", "category": "<category_3>",
     "temporal_position": 3, "temporal_marker": "<marker_3>"},
    {"text": "...", "category": "<category_4>",
     "temporal_position": 4, "temporal_marker": "<marker_4>"}
  ]
}

Generate the chain for theme "{evolution_theme}" using categories
(in order): {categories}.
\end{verbatim}
\end{promptbox}

\paragraph{Multi-Session Reasoning.}
The visual identity-match prompt is the canonical MSR generator (sub-pattern D1). It produces a scenario where one named entity and one vague reference are scattered across temporally distinct contexts, and an image attached to the vague reference is the only signal that resolves identity. Five anti-leakage rules (temporal isolation, no ownership disambiguation, no descriptive leakage, no text-based identity resolution, no cultural narrowing) are enforced inside the prompt and re-checked by the text-only judge of subsection~\ref{app:prompts-filter}.

\begin{promptbox}[colback=black!5, colframe=white!40!black, title=Question Generation: MSR (Entity Resolution / Visual Identity Match)]{}
\scriptsize
\begin{verbatim}
TASK: Generate a multi-session entity resolution question where the model
must determine whether an ambiguously referenced entity is the SAME AS a
specifically named entity.

Text facts reference an entity using VAGUE terms (e.g., "the dog I
adopted", "that plant from the market"). One fact has an image that
reveals the entity's visual identity. The model must match the image to a
named entity from another fact to answer.

THEME: {theme} ({theme_description})
EXAMPLE ITEMS: {example_entities}
NUM_FACTS: {num_facts}

RULES:
- At least 1 TEXT NEEDLE names a specific entity (e.g., "I got a golden
  retriever puppy named Max")
- At least 1 TEXT NEEDLE references an entity VAGUELY (e.g., "I adopted a
  dog from the shelter last week")
- EXACTLY 1 IMAGE NEEDLE with <image> token: shows the vaguely referenced
  entity visually. Text uses only vague terms (e.g., "Here's the dog I
  brought home <image>"). The image reveals whether it matches the named
  entity.
- Question: "Is the [vague reference] the same as [named entity]?"
  Answer: "Yes" or "No"
- {num_facts} facts as first-person chat messages ("I", "my"),
  2-4 sentences each
- CRITICAL: Every fact MUST be a first-person declarative statement.
  NEVER phrase facts as questions.
- EVERY fact must be vital to resolving the entity identity. Only use
  fact_type "text_needle" or "image_needle".
- Cross-modality: text alone cannot determine if the vague reference
  matches the named entity (the name is never stated for the vague
  reference). The image is the only way to confirm.
- Each fact should reference the entity with "entity_referenced" field
  tracking which entity it refers to.

ANTI-LEAKAGE RULES (violations cause rejection):
1. TEMPORAL ISOLATION: The named entity and the vague reference MUST
   appear in unrelated temporal contexts. NEVER place both in the same
   event, occasion, trip, or timeframe.
2. NO OWNERSHIP DISAMBIGUATION: Don't assign entities to different named
   owners if ownership alone reveals identity. Multiple entities should
   plausibly belong to the same person.
3. NO DESCRIPTIVE LEAKAGE: The vague reference text must NOT include
   adjectives that confirm or contradict the named entity's known
   characteristics.
4. NO TEXT-BASED IDENTITY RESOLUTION: No text fact may state what the
   vague reference IS. Phrases like "matches", "is the same as", "turns
   out to be", "which is actually" must NEVER appear in any fact.
5. NO CULTURAL NARROWING: The context must allow multiple plausible
   entity types.

NEGATIVE EXAMPLES (DO NOT generate scenarios like these):
- BAD (temporal co-occurrence): F1: "I named my new puppy Max, a golden
  retriever." F2: "I also got a dog from the shelter last week."
  Both acquired recently -> reader infers same event.
- BAD (ownership leakage): F1: "My roommate's cat is named Whiskers."
  F2: "The cat in my bedroom..." Different owners disambiguate.
- BAD (descriptive contradiction): F1: "I got a huge Great Dane named
  Duke." F2: "Here's the tiny puppy I found <image>" - "tiny" rules out
  Great Dane.
- BAD (explicit resolution): F1: "The plant I bought turned out to be a
  monstera." Text resolves identity directly.

OUTPUT (JSON only, no markdown):
{
  "theme": "{theme}",
  "named_entity": "<the specifically named entity>",
  "vague_reference": "<the vague reference used>",
  "same_entity": true/false,
  "question": "Is [vague_reference] the same [category] as [named_entity]?",
  "answer": "Yes" or "No",
  "explanation": "Step-by-step: F1 names [entity]. F2 refers vaguely to
                  [reference]. F3's image shows [what], which is/isn't
                  [named_entity]. Therefore Yes/No.",
  "facts": [
    {"fact_id":"F1","text":"...","fact_type":"text_needle",
     "has_image":false,"entity_referenced":"..."},
    ...,
    {"fact_id":"F3","text":"... <image> ...",
     "fact_type":"image_needle","has_image":true,
     "entity_referenced":"...","image_description":"...",
     "image_search_query":"...","image_search_object":"...",
     "image_provides":"..."}
  ]
}

Generate a creative, realistic scenario now:
\end{verbatim}
\end{promptbox}

\paragraph{Answer Refusal.}
The few-shot prompt below produces a question that is on-topic for the given context but whose answer is not present in the context. The fixed gold answer is the literal string \texttt{NOT\_MENTIONED}, which the judge rubric in subsection~\ref{app:prompts-eval} treats as a successful abstention.

\begin{promptbox}[colback=black!5, colframe=white!40!black, title=Question Generation: AR (Abstention Few-Shot)]{}
\scriptsize
\begin{verbatim}
You are generating Abstention QA for a multimodal benchmark.

Goal: craft a plausible, on-topic question that appears answerable from
the given info (paragraph + image caption if provided) but is actually
unanswerable. The gold answer must be the literal string "NOT_MENTIONED".

Rules
1) Use ONLY the provided context:
   * CONTEXT: the paragraph
   * VISUAL: the image or its caption (if provided)
2) The question must be topically related to entities/events/attributes
   present in the context.
   - Good: asks for a missing detail about a mentioned item.
   - Bad : asks about something totally unrelated.
3) The correct answer must be truly unknowable from the given info.
4) Output one compact JSON object with EXACTLY these keys:
   {
     "question": "one interrogative sentence",
     "answer": "NOT_MENTIONED",
     "evidence": "short extractive snippet proving the info is absent",
     "explanation": "one-sentence rationale why we must abstain"
   }
5) Do NOT add extra keys or commentary outside the JSON.

---------- Example ----------
INPUT
CONTEXT:
User: I upgraded my old 10-gallon tank, which has my betta fish, Bubbles.
User: I added decorations to the 20-gallon tank for more hiding places.

OUTPUT
{
  "question": "How many fish are there in my 30-gallon tank?",
  "answer": "NOT_MENTIONED",
  "evidence": "The user never mentions owning a 30-gallon tank.",
  "explanation": "No sentence discusses fish in a 30-gallon tank, so the
                  answer cannot be inferred."
}
-----------------------------
\end{verbatim}
\end{promptbox}

\subsection{Evidence Session Construction}
\label{app:prompts-evidence}

The evidence pipeline wraps each generated needle fact into a multi-turn session that is structurally indistinguishable from a haystack session. The box below contains the prompt that drives the assistant side of these sessions. Length is targeted at 250 to 350 words per turn, with a knowledge-oriented follow-up rather than a personal-social one, so that the evidence session looks informative without advertising the role as the resolution channel for any specific needle question. The user side is generated by a paired template that injects the needle fact into the message body using a directive that is itself constrained to first-person conversational style, after which a six-stage validator chain (rule-based length, photo-directive, n-gram leakage, semantic leakage, ambiguity preservation, and end-marker checks) decides whether the turn is accepted or regenerated.

\begin{promptbox}[colback=black!5, colframe=white!40!black, title=Evidence Session: Assistant Turn Template]{}
\scriptsize
\begin{verbatim}
You are {assistant_name}, a helpful AI assistant having a casual
conversation about {topic}.

Current conversation:
{chat_history}

{user_name}'s last message: "{last_msg}"

{image_context}

Generate a helpful, conversational response (TARGET: 250-350 words - be
concise, not lecture-like):

STRUCTURE YOUR RESPONSE:
1. Acknowledge what they shared (1-2 sentences)
   - If they shared a photo, briefly acknowledge it (1 sentence max),
     then move on
2. Share your perspective or relevant information (main body, 150-250
   words)
   - Provide helpful context, explanations, or suggestions
   - Keep it focused - cover 1-2 key points, not exhaustive lists
   - If applicable, offer practical tips or recommendations
3. End with a follow-up thought or question (1 sentence). Ask
   knowledge-oriented questions ("Would you like tips on X?" or "What
   part of this are you most stuck on?"), NOT personal/social questions
   ("Do you prefer X or Y?" or "Have you ever tried X?").

IMPORTANT: Write like a helpful chat assistant, NOT a textbook. Keep
responses focused and avoid padding with extra context the user didn't
ask for.

STYLE GUIDELINES:
- Be warm, conversational, and genuinely helpful
- Use natural paragraph breaks for readability
- Keep photo acknowledgment BRIEF (don't describe or analyze the image
  in detail)
- Provide substantive value - don't just agree, add to the conversation
- Do NOT repeat back personal details verbatim
- Do NOT use emojis

Output only your response, nothing else.
\end{verbatim}
\end{promptbox}

\subsection{Anti-Shortcut Quality Filter}
\label{app:prompts-filter}

After question generation, every candidate is passed through a text-only judge that simulates a model with no access to images. The box below shows the prompt used for this judge. A candidate is rejected when the judge returns \texttt{answerable\_without\_image = true}, which guarantees that the released benchmark cannot be solved by a strong text-only baseline through chain-of-thought guessing alone. A second visual judge (omitted from this appendix, but structurally identical apart from the role swap) checks the converse, that an image together with the question is sufficient to determine the answer; the two filters together enforce the cross-modal grounding contract of the construction pipeline.

\begin{promptbox}[colback=black!5, colframe=white!40!black, title=Filter: Text-Only Leakage Judge]{}
\scriptsize
\begin{verbatim}
[SYSTEM]
You are a judge evaluating whether a question can be answered from text
alone, WITHOUT seeing any images.

Your task is to determine if the answer to a question can be inferred
from the conversation text.

Rules:
1. Carefully read the conversation text
2. Try to answer the question using ONLY the text (no images)
3. If you can confidently answer the question from text alone ->
   answerable_without_image = true
4. If the answer requires visual inspection of an image ->
   answerable_without_image = false

Be strict: even if you can make an educated guess, if the text doesn't
explicitly contain the answer, mark it as NOT answerable from text.

[USER]
<conversation_text>{conversation_text}</conversation_text>
<question>{question}</question>
<reference_answer>{answer}</reference_answer>

<task>
Can this question be answered from the conversation text alone, without
seeing any images?

Return JSON:
{
  "answerable_without_image": true|false,
  "text_based_answer": "your answer attempt from text only, or
                        'CANNOT_DETERMINE'",
  "confidence": "high|medium|low",
  "explanation": "why the answer is/isn't determinable from text",
  "leakage_evidence": ["list of text snippets that reveal the answer,
                        if any"]
}
</task>
\end{verbatim}
\end{promptbox}

\section{Supplementary Experiments and Analysis}
\label{app:supplementary_experiments}

\subsection{Extended Results Tables}
\label{app:extended}

\begin{table}[h]
\centering
\caption{Comprehensive per-type accuracy (\%) for all 27 LVLMs at 32K / 64K / 128K contexts (LLM-as-Judge, $n=789$). IE: Information Extraction, MSR: Multi-Session Reasoning, TR: Temporal Reasoning, KU: Knowledge Update, AR: Answer Refusal. Cells marked --- indicate the model's context window does not support the input length. This table is the full roster corresponding to the representative subset in Figure~\ref{fig:per_type_heatmap}; agents are reported separately in Table~\ref{tab:per_type_full_agent}.}
\label{tab:per_type_full_vlm}
\scriptsize
\setlength{\tabcolsep}{2.5pt}
\resizebox{\linewidth}{!}{%
\begin{tabular}{@{}l|cccccc|cccccc|cccccc@{}}
\toprule
 & \multicolumn{6}{c|}{\textbf{32K}} & \multicolumn{6}{c|}{\textbf{64K}} & \multicolumn{6}{c}{\textbf{128K}} \\
\cmidrule(lr){2-7} \cmidrule(lr){8-13} \cmidrule(lr){14-19}
\textbf{Model} & IE & MSR & TR & KU & AR & Ov. & IE & MSR & TR & KU & AR & Ov. & IE & MSR & TR & KU & AR & Ov. \\
\midrule
Claude Sonnet 4.5 & 28.05 & 13.29 & 40.21 & 29.31 & 97.78 & 36.50 & 26.42 & 9.09 & 35.05 & 21.55 & 94.44 & 32.45 & 19.51 & 2.80 & 32.99 & 16.38 & 93.33 & 27.76 \\
Gemini-3.1-Pro & 57.32 & 32.17 & 40.93 & 49.14 & 97.75 & 54.10 & 58.94 & 31.47 & 42.27 & 48.62 & 95.56 & 52.99 & 55.79 & 29.37 & 41.24 & 46.03 & 96.67 & 51.99 \\
GPT-5.4 & 69.51 & 28.18 & 39.18 & 47.41 & 97.78 & 52.72 & 63.01 & 27.27 & 42.27 & 43.10 & 96.67 & 52.34 & 60.16 & 21.68 & 39.69 & 43.10 & 94.44 & 49.56 \\
Kimi-K2.5 & 54.78 & 44.06 & 53.09 & 50.86 & 97.78 & 54.88 & 52.44 & 35.66 & 52.06 & 49.14 & 97.78 & 53.99 & 51.63 & 28.67 & 48.25 & 48.59 & 93.33 & 51.99 \\
\midrule
Qwen3.5-122B-A10B & 74.39 & 30.07 & 51.55 & 49.14 & 88.89 & 58.68 & 67.07 & 27.27 & 51.55 & 48.59 & 84.44 & 55.89 & 43.09 & 22.38 & 41.75 & 46.03 & 83.33 & 45.50 \\
Qwen3.5-27B & 70.33 & 29.37 & 39.18 & 42.24 & 86.67 & 52.98 & 62.20 & 20.98 & 35.57 & 43.97 & 67.78 & 46.13 & 54.07 & 13.99 & 34.02 & 38.79 & 63.33 & 40.68 \\
Qwen3.5-9B & 67.48 & 16.78 & 38.66 & 43.10 & 91.11 & 50.32 & 60.57 & 13.99 & 33.51 & 37.93 & 80.00 & 44.36 & 45.12 & 5.59 & 26.80 & 25.86 & 62.22 & 32.57 \\
Qwen3.5-4B & 56.91 & 11.89 & 38.66 & 34.48 & 86.67 & 44.36 & 49.19 & 9.79 & 30.41 & 28.45 & 73.33 & 37.14 & 29.67 & 3.50 & 24.74 & 19.83 & 55.56 & 25.22 \\
Qwen3.5-2B & 64.23 & 17.48 & 32.47 & 33.62 & 72.22 & 44.36 & 2.03 & 0.00 & 2.06 & 0.86 & 15.56 & 3.04 & 1.22 & 0.00 & 1.55 & 0.00 & 1.11 & 0.89 \\
\midrule
Nemotron-Nano-12B & 63.82 & 16.78 & 44.85 & 39.66 & 45.56 & 44.99 & 56.10 & 20.98 & 42.78 & 34.48 & 35.56 & 40.94 & --- & --- & --- & --- & --- & --- \\
Cosmos-Reason2-8B & 55.69 & 23.78 & 48.45 & 31.90 & 68.89 & 46.13 & 46.34 & 20.28 & 45.88 & 24.14 & 54.44 & 39.16 & 36.99 & 20.28 & 43.81 & 12.93 & 35.56 & 31.94 \\
Phi4-Multimodal & 31.71 & 20.28 & 43.81 & 19.83 & 43.33 & 32.19 & 28.05 & 20.98 & 39.18 & 24.14 & 21.11 & 28.14 & 22.76 & 22.38 & 30.93 & 14.66 & 15.56 & 22.69 \\
\midrule
GLM-4.6V & 59.35 & 20.28 & 53.09 & 43.97 & 93.33 & 52.34 & 61.79 & 23.78 & 53.09 & 34.48 & 64.44 & 49.05 & 52.85 & 23.08 & 43.81 & 28.45 & 30.00 & 39.04 \\
GLM-4.5V & 61.79 & 15.38 & 39.69 & 24.14 & 51.11 & 41.19 & 45.93 & 15.38 & 33.51 & 14.66 & 32.22 & 31.18 & --- & --- & --- & --- & --- & --- \\
\midrule
Gemma3-27B & 34.96 & 23.78 & 41.24 & 30.17 & 68.89 & 37.64 & 26.42 & 22.38 & 44.33 & 27.59 & 61.11 & 34.22 & 19.11 & 18.18 & 37.11 & 13.79 & 48.89 & 25.98 \\
Gemma3-12B & 36.18 & 22.38 & 43.81 & 29.31 & 53.33 & 36.50 & 30.89 & 22.38 & 44.33 & 20.69 & 38.89 & 32.07 & 23.58 & 18.18 & 28.35 & 16.38 & 28.89 & 23.32 \\
Gemma3-4B & 25.20 & 21.68 & 41.75 & 23.28 & 14.44 & 27.12 & 17.89 & 18.88 & 28.87 & 14.66 & 8.89 & 19.26 & 14.63 & 12.59 & 24.23 & 10.34 & 5.56 & 14.96 \\
\midrule
Qwen3-VL-235B (T) & 51.63 & 25.17 & 49.48 & 36.21 & 88.89 & 48.29 & 38.62 & 18.18 & 42.78 & 35.34 & 75.56 & 39.67 & 29.27 & 15.38 & 43.30 & 30.17 & 67.78 & 34.73 \\
Qwen3-VL-235B (I) & 60.98 & 18.88 & 55.67 & 40.52 & 97.78 & 53.23 & 56.10 & 20.98 & 54.12 & 36.21 & 95.56 & 50.82 & 51.22 & 18.18 & 54.64 & 30.17 & 92.22 & 47.66 \\
Qwen3-VL-30B (T) & 41.46 & 11.89 & 35.57 & 30.17 & 71.11 & 36.38 & 39.02 & 11.89 & 37.11 & 29.31 & 65.56 & 35.23 & 33.74 & 10.49 & 36.60 & 20.69 & 60.00 & 31.31 \\
Qwen3-VL-30B (I) & 65.45 & 18.88 & 60.82 & 37.93 & 93.33 & 55.01 & 56.91 & 20.28 & 55.15 & 39.66 & 78.89 & 49.81 & 56.50 & 16.08 & 52.58 & 24.14 & 63.33 & 44.23 \\
Qwen3-VL-8B (T) & 51.22 & 17.48 & 52.58 & 32.76 & 80.00 & 46.01 & 42.28 & 18.88 & 50.52 & 30.17 & 54.44 & 39.67 & 33.33 & 8.39 & 39.18 & 24.14 & 45.56 & 30.29 \\
Qwen3-VL-8B (I) & 52.85 & 17.48 & 58.25 & 33.62 & 90.00 & 49.18 & 44.72 & 19.58 & 47.42 & 29.31 & 82.22 & 42.84 & 32.52 & 20.28 & 44.85 & 15.52 & 66.67 & 34.73 \\
Qwen3-VL-4B (T) & 46.75 & 20.98 & 43.30 & 33.62 & 82.22 & 43.35 & 28.05 & 11.19 & 26.80 & 18.10 & 70.00 & 28.01 & 17.07 & 9.09 & 13.92 & 10.34 & 52.22 & 17.87 \\
Qwen3-VL-4B (I) & 52.03 & 17.48 & 42.78 & 25.86 & 92.22 & 44.23 & 42.28 & 19.58 & 35.57 & 19.83 & 75.56 & 37.01 & 28.46 & 16.08 & 35.05 & 13.79 & 50.00 & 28.14 \\
Qwen3-VL-2B (T) & 26.83 & 9.09 & 19.07 & 11.21 & 87.78 & 26.36 & 19.11 & 6.99 & 11.86 & 16.38 & 54.44 & 18.76 & 16.67 & 12.59 & 5.67 & 8.62 & 14.44 & 11.79 \\
Qwen3-VL-2B (I) & 32.52 & 16.78 & 50.52 & 20.69 & 72.22 & 36.88 & 32.52 & 21.68 & 40.72 & 16.38 & 65.56 & 33.97 & 25.20 & 18.88 & 32.99 & 8.62 & 47.78 & 26.11 \\
\bottomrule
\end{tabular}%
}
\end{table}

\begin{table}[h]
\centering
\caption{Comprehensive per-type accuracy (\%) for all seven memory-augmented agents at 32K / 64K / 128K / 256K contexts, evaluated on 195-question canonical subsets. IE: Information Extraction, MSR: Multi-Session Reasoning, TR: Temporal Reasoning, KU: Knowledge Update, AR: Answer Refusal. Only memory agents are evaluated at 256K; direct LVLMs are not (see Table~\ref{tab:per_type_full_vlm}).}
\label{tab:per_type_full_agent}
\scriptsize
\setlength{\tabcolsep}{2.0pt}
\resizebox{\linewidth}{!}{%
\begin{tabular}{@{}l|cccccc|cccccc|cccccc|cccccc@{}}
\toprule
 & \multicolumn{6}{c|}{\textbf{32K}} & \multicolumn{6}{c|}{\textbf{64K}} & \multicolumn{6}{c|}{\textbf{128K}} & \multicolumn{6}{c}{\textbf{256K}} \\
\cmidrule(lr){2-7} \cmidrule(lr){8-13} \cmidrule(lr){14-19} \cmidrule(lr){20-25}
\textbf{Model} & IE & MSR & TR & KU & AR & Ov. & IE & MSR & TR & KU & AR & Ov. & IE & MSR & TR & KU & AR & Ov. & IE & MSR & TR & KU & AR & Ov. \\
\midrule
Mem0 & 13.11 & 25.71 & 50.00 & 17.24 & 77.27 & 31.79 & 11.48 & 22.86 & 45.83 & 24.14 & 77.27 & 31.28 & 8.20 & 20.00 & 50.00 & 24.14 & 77.27 & 30.26 & 11.48 & 22.86 & 47.92 & 20.69 & 72.73 & 30.77 \\
MemOS & 18.03 & 22.86 & 39.58 & 24.14 & 68.18 & 30.26 & 18.03 & 22.86 & 39.58 & 24.14 & 68.18 & 30.77 & 16.39 & 22.86 & 43.75 & 17.24 & 81.82 & 31.28 & 16.39 & 22.86 & 37.50 & 17.24 & 72.73 & 29.74 \\
MemAgent-7B & 18.03 & 25.71 & 62.50 & 41.38 & 13.64 & 32.82 & 13.11 & 28.57 & 45.83 & 24.14 & 4.55 & 25.64 & 14.75 & 28.57 & 60.42 & 20.69 & 9.09 & 28.21 & 11.48 & 22.86 & 50.00 & 20.69 & 4.55 & 24.62 \\
Memory-T1 & 18.03 & 25.71 & 62.50 & 13.79 & 9.09 & 28.72 & 19.67 & 22.86 & 56.25 & 17.24 & 18.18 & 28.72 & 21.31 & 22.86 & 56.25 & 27.59 & 13.64 & 30.26 & 18.03 & 25.71 & 58.33 & 24.14 & 9.09 & 29.23 \\
M3-Agent & 18.03 & 22.86 & 29.17 & 6.90 & 13.64 & 19.49 & 18.03 & 17.14 & 37.50 & 3.45 & 4.55 & 18.97 & 22.95 & 20.00 & 22.92 & 6.90 & 18.18 & 19.49 & 22.95 & 28.57 & 27.08 & 13.79 & 22.73 & 23.59 \\
M3C & 8.20 & 25.71 & 31.25 & 10.34 & 18.18 & 18.46 & 9.84 & 25.71 & 27.08 & 6.90 & 22.73 & 17.95 & 8.20 & 25.71 & 27.08 & 0.00 & 9.09 & 14.87 & 11.48 & 25.71 & 29.17 & 0.00 & 13.64 & 16.92 \\
M2A & 14.75 & 8.57 & 2.08 & 0.00 & 22.73 & 15.38 & 18.03 & 14.29 & 25.00 & 0.00 & 13.64 & 15.90 & 21.31 & 11.43 & 27.08 & 0.00 & 13.64 & 16.92 & 21.31 & 11.43 & 25.00 & 0.00 & 13.64 & 16.41 \\
\bottomrule
\end{tabular}%
}
\end{table}

\subsection{Canonical 195-question Subset for Agent Evaluation}
\label{app:canonical195}

The seven memory-augmented agents in \S\ref{subsec:analysis} (full analysis in Appendix~\ref{app:agent_underperformance}) are evaluated on a 195-question canonical subset rather than the full 789-question benchmark, because agent pipelines are substantially slower than direct-LVLM inference (M2A takes roughly $60{\times}$ longer per question). This appendix documents how the 195-question set is derived and confirms that its per-type composition preserves the full-benchmark distribution.

\paragraph{Derivation.}
The 195-question (a quarter of the original dataset) canonical subset is constructed by stratified sampling (seed${=}42$) from the full 789-question benchmark, preserving per-type proportions. All seven memory-augmented agents are evaluated on this same subset.

\paragraph{Per-type composition matches the full benchmark.}
Table~\ref{tab:canonical195_strat} compares the 195-subset's per-type proportions against the full 789-question benchmark. Differences are below 0.2 percentage points for every type, so rankings computed on the subset transfer to the full benchmark without systematic bias.

\begin{table}[h]
\centering
\small
\caption{Per-type composition of the 195-question canonical subset compared against the full 789-question benchmark. Proportions are preserved to within 0.2 percentage points per type.}
\label{tab:canonical195_strat}
\begin{tabular}{@{}lccc@{}}
\toprule
Type & Full 789 (\%) & 195-subset & 195 share (\%) \\
\midrule
Information Extraction (IE)      & 31.18 & 61 & 31.28 \\
Multi-Session Reasoning (MSR)    & 18.12 & 35 & 17.95 \\
Temporal Reasoning (TR)          & 24.59 & 48 & 24.62 \\
Knowledge Update (KU)            & 14.70 & 29 & 14.87 \\
Answer Refusal (AR)              & 11.41 & 22 & 11.28 \\
\midrule
Total                            & 100.00 & 195 & 100.00 \\
\bottomrule
\end{tabular}
\end{table}

\paragraph{Direct-LVLM overlay on the 195-subset.}
To enable apples-to-apples comparison with the seven memory agents, we re-score every direct-LVLM run used elsewhere in the paper on exactly the 195 canonical qids. Table~\ref{tab:agent_vs_vlm_195} reports overall and per-type accuracy for six representative LVLMs at 32K/64K/128K and the four agents at 32K/64K/128K/256K. This is the matched-subset version of Figure~\ref{fig:per_type_heatmap}; rankings transfer between the subset and the full benchmark at 32K with Spearman $\rho = 0.94$ ($p < 0.01$, $n = 6$ direct LVLMs), so conclusions drawn on the 195-subset are not an artifact of subset choice.

\begin{table}[h]
\centering
\scriptsize
\setlength{\tabcolsep}{3pt}
\caption{Agent-vs-direct-LVLM comparison on the canonical 195-question subset. Overall and per-type accuracy (\%) at 32K/64K/128K (direct LVLMs) and 32K/64K/128K/256K (agents). The agents listed are the original four systems; the six direct LVLMs span the best API system (GPT-5.4), the best closed-commercial (Gemini-3.1-Pro), the best open (Qwen3.5-122B), and the three Qwen3-VL sizes that serve as matched-backbone counterparts. The remaining agents (Mem0, MemOS, MemAgent-7B) are reported in Table~\ref{tab:per_type_full_agent}. Per-type $n$ counts are shown once at the top (61 IE / 35 MSR / 48 TR / 29 KU / 22 AR).}
\label{tab:agent_vs_vlm_195}
\begin{tabular}{@{}lcccccc@{}}
\toprule
 & Overall & IE & MSR & TR & KU & AR \\
System / Ctx & (n=195) & (61) & (35) & (48) & (29) & (22) \\
\midrule
\multicolumn{7}{l}{\emph{Direct LVLMs on the 195-subset}} \\
Gemini-3.1-Pro / 32K      & 55.38 & 60.66 & 45.71 & 39.58 & 48.28 & 100.00 \\
Gemini-3.1-Pro / 64K      & 58.46 & 59.02 & 48.57 & 41.67 & 65.52 & 100.00 \\
Gemini-3.1-Pro / 128K     & 58.46 & 68.85 & 42.86 & 43.75 & 48.28 & 100.00 \\
GPT-5.4 / 32K             & 51.28 & 63.93 & 17.14 & 35.42 & 55.17 & 100.00 \\
GPT-5.4 / 64K             & 56.92 & 65.57 & 37.14 & 47.92 & 44.83 & 100.00 \\
GPT-5.4 / 128K            & 51.79 & 67.21 & 28.57 & 37.50 & 41.38 &  90.91 \\
\midrule
Qwen3.5-122B / 32K        & 63.59 & 75.41 & 45.71 & 58.33 & 44.83 &  95.45 \\
Qwen3.5-122B / 64K        & 59.49 & 67.21 & 37.14 & 52.08 & 58.62 &  90.91 \\
Qwen3.5-122B / 128K       & 49.23 & 45.90 & 31.43 & 43.75 & 58.62 &  86.36 \\
\midrule
Qwen3-VL-30B (I) / 32K    & 58.46 & 65.57 & 25.71 & 62.50 & 48.28 &  95.45 \\
Qwen3-VL-30B (I) / 64K    & 52.31 & 57.38 & 25.71 & 56.25 & 44.83 &  81.82 \\
Qwen3-VL-30B (I) / 128K   & 50.77 & 57.38 & 25.71 & 66.67 & 27.59 &  68.18 \\
Qwen3-VL-8B (I) / 32K     & 50.77 & 52.46 & 22.86 & 58.33 & 44.83 &  81.82 \\
Qwen3-VL-8B (I) / 64K     & 47.69 & 52.46 & 25.71 & 50.00 & 41.38 &  72.73 \\
Qwen3-VL-8B (I) / 128K    & 34.36 & 27.87 & 22.86 & 54.17 & 13.79 &  54.55 \\
Qwen3-VL-2B (I) / 32K     & 38.46 & 27.87 & 20.00 & 62.50 & 24.14 &  63.64 \\
Qwen3-VL-2B (I) / 64K     & 35.38 & 29.51 & 37.14 & 47.92 & 10.34 &  54.55 \\
Qwen3-VL-2B (I) / 128K    & 27.69 & 18.03 & 28.57 & 41.67 & 10.34 &  45.45 \\
\midrule
\multicolumn{7}{l}{\emph{Memory-augmented agents on the 195-subset}} \\
M3-Agent / 32K            & 19.49 & 18.03 & 22.86 & 29.17 &  6.90 & 13.64 \\
M3-Agent / 64K            & 18.97 & 18.03 & 17.14 & 37.50 &  3.45 &  4.55 \\
M3-Agent / 128K           & 19.49 & 22.95 & 20.00 & 22.92 &  6.90 & 18.18 \\
M3-Agent / 256K           & 23.59 & 22.95 & 28.57 & 27.08 & 13.79 & 22.73 \\
M2A / 32K                 & 15.38 & 14.75 &  8.57 &  2.08 &  0.00 & 22.73 \\
M2A / 64K                 & 15.90 & 18.03 & 14.29 & 25.00 &  0.00 & 13.64 \\
M2A / 128K                & 16.92 & 21.31 & 11.43 & 27.08 &  0.00 & 13.64 \\
M2A / 256K                & 16.41 & 21.31 & 11.43 & 25.00 &  0.00 & 13.64 \\
M3C / 32K                 & 18.46 &  8.20 & 25.71 & 31.25 & 10.34 & 18.18 \\
M3C / 64K                 & 17.95 &  9.84 & 25.71 & 27.08 &  6.90 & 22.73 \\
M3C / 128K                & 14.87 &  8.20 & 25.71 & 27.08 &  0.00 &  9.09 \\
M3C / 256K                & 16.92 & 11.48 & 25.71 & 29.17 &  0.00 & 13.64 \\
Memory-T1 / 32K           & 28.72 & 18.03 & 25.71 & 62.50 & 13.79 &  9.09 \\
Memory-T1 / 64K           & 28.72 & 19.67 & 22.86 & 56.25 & 17.24 & 18.18 \\
Memory-T1 / 128K          & 30.26 & 21.31 & 22.86 & 56.25 & 27.59 & 13.64 \\
Memory-T1 / 256K          & 29.23 & 18.03 & 25.71 & 58.33 & 24.14 &  9.09 \\
\bottomrule
\end{tabular}
\end{table}

\paragraph{Bootstrap confidence intervals on overall agent accuracy.}
We resample with replacement at the question level (1000 iterations, percentile method) on the canonical 195 questions to bound the subset-induced uncertainty in each agent's overall accuracy. The 95\% confidence intervals at 32K (mean $\pm$ half-width, $n=195$) are: Mem0 $31.79 \pm 6.67$, MemOS $30.26 \pm 6.15$, MemAgent-7B $32.82 \pm 6.42$, Memory-T1 $28.72 \pm 6.15$, M3-Agent $19.49 \pm 5.38$, M3C $18.46 \pm 5.38$, M2A $15.38 \pm 5.38$. At 128K: Mem0 $30.26 \pm 6.67$, MemOS $31.28 \pm 6.67$, MemAgent-7B $28.21 \pm 6.41$, Memory-T1 $30.26 \pm 6.41$, M3-Agent $19.49 \pm 5.38$, M3C $14.87 \pm 4.87$, M2A $16.92 \pm 5.13$. All half-widths fall within $[4.87, 6.67]\%$ (mean $5.93\%$); the four text-only / caption-based pipelines (Mem0, MemOS, MemAgent-7B, Memory-T1) cluster within their overlapping intervals at both lengths and are uniformly above the three multimodal pipelines (M3-Agent, M3C, M2A), whose intervals also overlap with one another but not with the upper cluster. Subset resampling therefore preserves the qualitative cluster structure used in \S\ref{subsec:analysis} and Table~\ref{tab:agent_vs_vlm_195}, and the agent--LVLM gap remains larger than any subset-induced confidence band.

\paragraph{Scope of agent claims.}
The accuracies in Table~\ref{tab:agent_vs_vlm_195} and Table~\ref{tab:per_type_full_agent} reflect each memory pipeline as released, with the input adaptations documented in Appendix~\ref{app:eval_setup}: text-only agents (Mem0, MemOS, MemAgent-7B, Memory-T1) consume BLIP-2 captions in lieu of pixels; M3-Agent consumes per-session composite images rendered to fit its video-LVLM input format; M2A and M3C consume original images via their native LVLM backbones. We do not claim these accuracies are upper bounds on memory-augmented architectures in general --- a pipeline that retained pixel-level evidence at retrieval time, or that swapped BLIP-2 for a stronger captioner (cf.\ Appendix~\ref{app:limitations}), might score higher. The conclusion supported by Table~\ref{tab:agent_vs_vlm_195} and \S\ref{subsec:analysis} is the narrower one: under each agent's released visual interface, current memory pipelines lose faithfulness to original visual evidence relative to direct-LVLM grounding on the same 195 questions, while remaining length-robust across the 32K--256K range.

\subsection{Coverage and Per-Answer Accuracy}
\label{app:coverage_full}

\begin{table}[h]
\centering
\caption{Coverage and Per-Answer Accuracy ($n=789$, 699 answerable). Coverage = \% of answerable questions attempted. PA = accuracy on attempted answers. Models are grouped by family (APIs, then open-source LVLMs). The 64K and 128K coverage decompositions inherit the same trend (frontier APIs over-refuse, open-weight LVLMs over-attempt).}
\label{tab:coverage_full}
\small
\begin{tabular}{@{}llcccc@{}}
\toprule
\textbf{Model} & \textbf{M} & \textbf{Judge} & \textbf{Coverage} & \textbf{PA} & \textbf{Refused} \\
\midrule
Claude Sonnet 4.5 & I & 36.50 & 49.50 & 57.51 & 353 \\
Gemini-3.1-Pro & I & 54.10 & 66.76 & 69.10 & 232 \\
GPT-5.4 & I & 52.72 & 63.95 & 73.15 & 252 \\
Kimi-K2.5 & I & 54.88 & 72.39 & 67.98 & 193 \\
\midrule
Qwen3.5-122B & I & 58.68 & 87.70 & 62.32 & 86 \\
\midrule
Cosmos-Reason2-8B & I & 46.13 & 99.43 & 43.31 & 4 \\
\midrule
GLM-4.6V & I & 52.34 & 95.85 & 48.96 & 29 \\
\midrule
Qwen3-VL-235B & I & 53.23 & 79.26 & 59.75 & 145 \\
Qwen3-VL-30B & I & 55.01 & 96.99 & 51.47 & 21 \\
Qwen3-VL-8B & I & 49.18 & 92.13 & 47.52 & 55 \\
\bottomrule
\end{tabular}
\end{table}

\paragraph{Coverage--accuracy trade-off.}
\label{app:coverage_analysis}

Overall accuracy conflates retrieval ability with calibration. Decomposing performance into coverage (fraction of answerable questions attempted) and per-answer accuracy (PA) reveals two contrasting strategies. GPT-5.4 reaches the highest PA (73.15\%) but refuses 36.05\% of answerable questions; Qwen3.5-122B attempts 87.70\% but achieves only 62.32\% PA. Claude Sonnet 4.5, which ranks 23rd overall, achieves a respectable 57.51\% PA behind a 50.50\% refusal rate. The full decomposition appears in Table~\ref{tab:coverage_full}. At 128K, most models answer \emph{more} questions than at 32K (coverage rises) despite lower accuracy, confirming that false confidence, not increased caution, is the dominant dynamic at longer contexts.

\paragraph{Claude Sonnet 4.5: a calibration-driven outlier.}
Per-type decomposition sharpens the diagnosis. Claude's coverage varies from 41.26\% (MSR) to 64.95\% (TR), uniformly below the frontier-API median: GPT-5.4 and Gemini-3.1-Pro cover 45--83\% and 52--80\% respectively. On IE and KU, where Claude's overall accuracy appears weak (28.05\% and 29.31\%), its per-answer accuracy reaches 63.55\% and 62.96\%---competitive with models ranked considerably higher---but a ${\sim}$55\% refusal rate on these types suppresses the headline number. TR has the highest coverage (64.95\%), consistent with timestamps providing explicit retrieval anchors that raise the model's confidence. MSR is the one type where both coverage \emph{and} per-answer accuracy are low (41.26\% / 32.20\%), suggesting that cross-session aggregation genuinely challenges the model rather than merely triggering its refusal heuristic. We note that Sonnet is a mid-tier offering in Anthropic's model family; the refusal-driven deficit reflects model-specific calibration rather than a benchmark fairness concern.

\subsection{Wrong-Answer Error Analysis}
\label{app:wrong_answer_taxonomy}

\begin{figure}[h]
\centering
\includegraphics[width=0.70\linewidth]{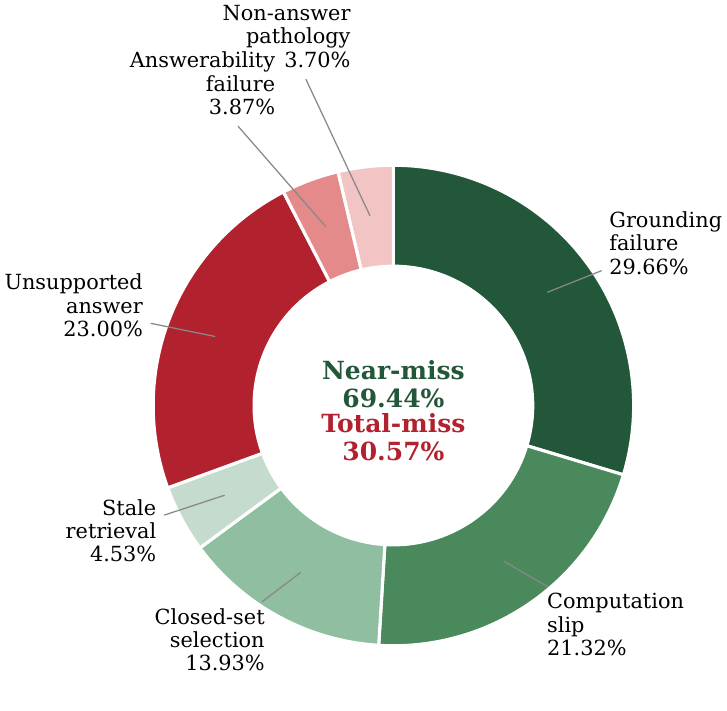}
\caption{Distribution of wrong-answer types at 32K context ($n=5{,}592$ attempted-but-incorrect LVLM answers, out of $27 \times 789 = 21{,}303$ evaluations). Near-miss errors (evidence located before erring) account for 69.44\% of wrong answers; total-miss errors (no correct evidence anchor) for 30.57\%.}
\label{fig:wrong_answer_pie}
\end{figure}

\begin{figure}[h]
\centering
\includegraphics[width=0.96\linewidth]{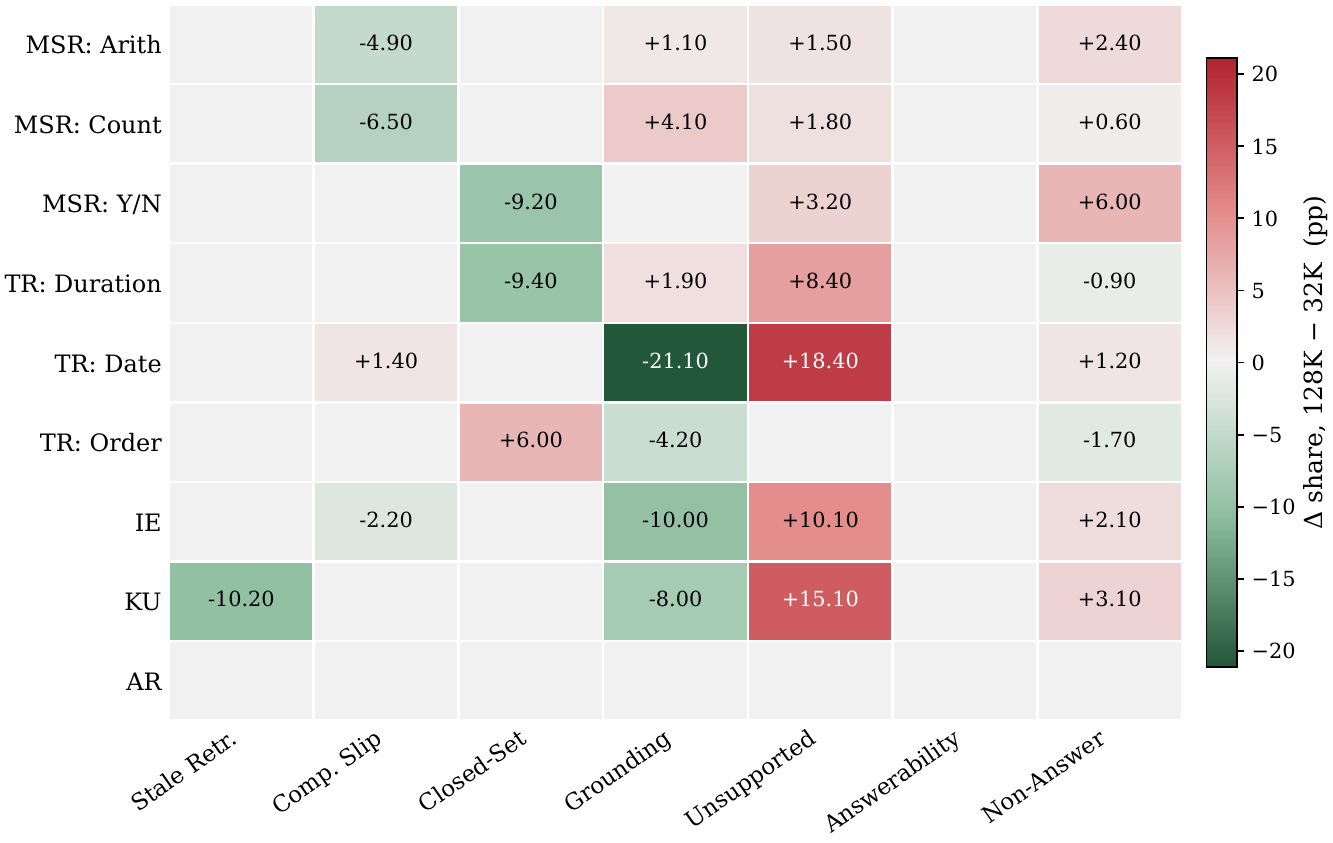}
\caption{Wrong-answer error-type shift from 32K to 128K by question type ($n=789$). Unsupported answer (+10.23\%) replaces grounding failure ($-$5.38\%) and computation slip ($-$4.82\%).}
\label{fig:context_delta_heatmap}
\end{figure}

This appendix supplements the wrong-answer error analysis in \S\ref{subsec:analysis} with
(a)~the motivation for the seven-label taxonomy, (b)~the per-label detection procedure used
to produce Figure~\ref{fig:wrong_answer_pie} and Figure~\ref{fig:context_delta_heatmap}
(Appendix~\ref{app:wrong_answer_figures}), and (c)~the modality re-grouping
(Table~\ref{tab:modality_mapping}) that turns the seven labels into the five categories shown in
Figure~\ref{fig:visual_error} of the main text.

\paragraph{Folding into two meta-categories.}
The four near-miss labels (grounding failure, computation slip,
closed-set selection, stale retrieval) share a property: each requires
the model to have located the relevant evidence before erring.
The three total-miss labels (unsupported answer, answerability
failure, non-answer pathology) share the opposite property: the model
produced content, or a non-answer, without a correct evidence anchor.
This dichotomy turns retrieval success into a binary signal that can
be read off the wrong-answer distribution, which is what permits the
retrieval-degradation account in \S\ref{subsec:analysis}.

\paragraph{Two labels are not enough.}
The $32\text{K}\!\to\!128\text{K}$ aggregate shift is $\pm 15.17\%$
by partition identity, so reporting only the meta-category totals
hides which labels absorb the shift. Unsupported answer alone carries
$+10.23\%$ of the shift, while answerability failure and non-answer
pathology each move by less than $3\%$. The single dominant label
inside total-miss cannot be identified without the seven-way split.

\paragraph{Grouping by question type is not enough either.}
Partitioning wrong answers by question type (five categories) obscures
the per-type failure signature: MSR Arithmetic errors are computation
slips, TR order-ranking errors are closed-set swaps, and KU errors are stale
retrievals. These signatures surface only when the error label is
separated from the question type; collapsing either dimension merges
failure modes that have different responses to context length.

\paragraph{Why exactly seven.}
Each of the seven labels either binds to a structural constraint of a
question-type subset (numeric answer $\Rightarrow$ computation slip;
closed answer set $\Rightarrow$ closed-set selection; KU chain
$\Rightarrow$ stale retrieval; AR item $\Rightarrow$ answerability
failure; non-terminating generation $\Rightarrow$ non-answer
pathology), or is a catch-all for free-form answers (grounding failure
when the prediction is evidence-anchored but wrong; unsupported answer
when it is not). Every attempted-but-incorrect answer receives exactly
one label, and no label admits a strictly finer partition without
introducing sub-labels that collapse to zero support on subsets of the
benchmark. Table~\ref{tab:wrong_answer_labels} summarises the seven labels with their detection method and category assignment.

\paragraph{Detection, Phase~1 (rule-based).}
Five labels are assigned by deterministic rules that depend only on
the question type and the prediction string.
\emph{Stale retrieval} (KU only): the prediction fuzzy-matches the
item's \texttt{old\_answer} field with \texttt{SequenceMatcher}
ratio $\geq 0.7$ on normalized (lower-cased, alphanumeric) strings.
\emph{Computation slip} (MSR Counting and Arithmetic): the prediction
and the reference both parse as numbers, and the two differ.
\emph{Closed-set selection} (TR Duration A/B, MSR Yes/No, TR order-ranking):
the prediction lies in the closed answer set for the item but is not
the reference element.
\emph{Answerability failure} (AR items): the prediction is a
substantive answer rather than a refusal, detected by the absence of
any refusal keyword from a fixed list of 14 phrases such as
``insufficient information'' and ``cannot be determined''.
\emph{Non-answer pathology} (any item): the prediction is empty, or
is verbose reasoning with no ``answer:'', ``therefore'', or ``final
answer'' anchor. Phase~1 labels take priority over Phase~2 when both
would apply; Phase~1 covers roughly 40\% of attempted-but-incorrect
items at each context.

\paragraph{Detection, Phase~2 (LLM-judged).}
The remaining items have free-form answers with no Phase~1 rule
matching. A GPT-4o-mini call reads the question, the reference
answer, and the prediction, and assigns one of the two free-form
labels. \emph{Grounding failure}: the prediction is semantically
related to the reference but incomplete, over-specific, mislocalized,
or at the wrong granularity. \emph{Unsupported answer}: the
prediction is a specific entity or value with no meaningful semantic
overlap with the reference. LLM classifications are cached in
\texttt{phase2\_cache\_\{ctx\}.jsonl} and reused across runs, so the
results are reproducible without new API calls.

\paragraph{Coverage and caveats.}
Phase~1 and the Phase~2 cache jointly cover roughly 90\% of wrong
answers per context; the residual are invalid answer which failed to give a confirmed final answer. This inflates
the unsupported-answer share by at most $\sim\!10\%$ of its own
value; rerunning with the full LLM pass is left as an extension
because the direction and magnitude of the near-miss $\to$
total-miss shift are stable under that default assignment.

\label{app:wrong_answer_figures}

\begin{table}[h]
\centering
\small
\caption{Seven-label wrong-answer taxonomy. Labels partition every attempted-but-incorrect answer into near-miss (evidence located) vs.\ total-miss (no evidence anchor).}
\label{tab:wrong_answer_labels}
\begin{tabular}{@{}llll@{}}
\toprule
\textbf{Label} & \textbf{Category} & \textbf{Detection} & \textbf{Definition} \\
\midrule
Grounding failure      & Near-miss  & LLM  & Right region, wrong detail. \\
Computation slip       & Near-miss  & Rule & Right operands, wrong arithmetic. \\
Closed-set selection   & Near-miss  & Rule & Right set, wrong element. \\
Stale retrieval        & Near-miss  & Rule & Right fact, pre-update version. \\
Unsupported answer     & Total-miss & LLM  & No anchor; fabricated content. \\
Answerability failure  & Total-miss & Rule & Answered an unanswerable item. \\
Non-answer pathology   & Total-miss & Rule & Never produced a final answer. \\
\bottomrule
\end{tabular}
\end{table}

\paragraph{Five-category modality view of wrong answers (Figure~\ref{fig:visual_error}).}
The wrong-answer pie (Figure~\ref{fig:wrong_answer_pie}) reports the seven-label distribution at the
attempted-but-incorrect level; the per-type bars in Figure~\ref{fig:visual_error} (main text) re-group every
wrong answer along a complementary modality axis by joining its seven-label tag with the per-question
image-dependency annotation (image-essential / image-supportive / text-sufficient, defined in
\S\ref{subsec:cross_modality}). The result is five disjoint modality categories that exhaust every
attempted-but-incorrect answer; Table~\ref{tab:modality_mapping} gives the exact mapping.
\begin{itemize}[leftmargin=1.2em,topsep=2pt,itemsep=2pt]
\item \textbf{Visual.} Grounding failure or unsupported answer on an image-essential question---the visual evidence was the source of the missing information.
\item \textbf{Textual.} Either (i) grounding failure or unsupported answer on a text-sufficient question, or (ii) any stale-retrieval error (intrinsically a textual-update miss on KU)---text was the source of the missing information.
\item \textbf{Mixed.} Grounding failure or unsupported answer on an image-supportive question, where image and text both contribute to the answer and either could be at fault.
\item \textbf{Reasoning.} Computation slip, closed-set selection, or answerability failure---a label-specific reasoning-shaped error (off-by-one arithmetic, wrong A/B choice, or substantive answer to an unanswerable item).
\item \textbf{Output.} Non-answer pathology (empty or non-extractable response).
\end{itemize}

\begin{table}[h]
\centering
\small
\caption{Mapping from the seven-label wrong-answer taxonomy ($\times$ per-question image dependency) to the five modality categories of Figure~\ref{fig:visual_error}. A single column entry indicates that the dependency does not affect the assignment.}
\label{tab:modality_mapping}
\begin{tabular}{@{}lccc@{}}
\toprule
\textbf{Seven-label} & \textbf{image-essential} & \textbf{image-supportive} & \textbf{text-sufficient} \\
\midrule
Grounding failure       & Visual    & Mixed    & Textual \\
Unsupported answer      & Visual    & Mixed    & Textual \\
Stale retrieval         & \multicolumn{3}{c}{Textual} \\
Computation slip        & \multicolumn{3}{c}{Reasoning} \\
Closed-set selection    & \multicolumn{3}{c}{Reasoning} \\
Answerability failure   & \multicolumn{3}{c}{Reasoning} \\
Non-answer pathology    & \multicolumn{3}{c}{Output} \\
\bottomrule
\end{tabular}
\end{table}

The seven-label and five-modality views are two lenses on the same wrong-answer set. The seven-label view records what \emph{kind} of error the model made (e.g., right answer set but wrong element); the five-modality view records which \emph{evidence channel} the error relied on. Reading the two together explains the per-type asymmetries in Figure~\ref{fig:visual_error}: a question type whose closed-set or arithmetic structure makes computation-slips and selection-swaps the dominant Phase-1 labels (e.g., MSR Counting/Arithmetic, MSR Yes/No, TR Duration A/B, TR order-ranking) will accordingly show a large Reasoning share regardless of how reliably the model retrieves evidence---which is why the Reasoning share in MSR/TR should be cross-checked against the oracle-retrieval diagnostic in \S\ref{app:msr_ceiling} before being interpreted as a reasoning bottleneck.

\paragraph{Case studies.}
To illustrate each error category concretely, we present one representative failure case per label below. Each card shows the question, the reference answer, and the model's incorrect prediction at the 32K context length.

\vspace{0.5em}

\begin{tcolorbox}[
  title={Case Study: Grounding Failure},
  enhanced, sharp corners,
  colback=gray!3, colframe=black!40,
  boxrule=0.4pt, fonttitle=\bfseries,
  left=4pt, right=4pt, top=3pt, bottom=3pt
]
\footnotesize

\textit{The model located the relevant evidence region but extracted or misread a specific detail---answering at wrong granularity or confusing a neighboring value.}
\vspace{0.4em}

\textbf{Model:} claude-sonnet-4.5 \hfill \textbf{Task:} Information Extraction \hfill \textbf{Context:} 32K
\vspace{0.3em}

\textbf{Question.} Where on the sign is the photo of the mountains located?
\vspace{0.3em}

\textbf{Reference Answer.} In the bottom right corner

\textbf{Model Prediction.} \textcolor{red}{In the upper right corner of the kiosk sign.}

\end{tcolorbox}

\begin{tcolorbox}[
  title={Case Study: Computation Slip},
  enhanced, sharp corners,
  colback=gray!3, colframe=black!40,
  boxrule=0.4pt, fonttitle=\bfseries,
  left=4pt, right=4pt, top=3pt, bottom=3pt
]
\footnotesize

\textit{The model identified the correct operands and operation type but produced an incorrect arithmetic result (off-by-one, summation error, or unit mismatch).}
\vspace{0.4em}

\textbf{Model:} claude-sonnet-4.5 \hfill \textbf{Task:} MSR --- Arithmetic \hfill \textbf{Context:} 32K
\vspace{0.3em}

\textbf{Question.} How much total have I spent on coffee makers?
\vspace{0.3em}

\textbf{Reference Answer.} \$260.00

\textbf{Model Prediction.} \textcolor{red}{\$150.00}

\end{tcolorbox}

\begin{tcolorbox}[
  title={Case Study: Closed-Set Selection Error},
  enhanced, sharp corners,
  colback=gray!3, colframe=black!40,
  boxrule=0.4pt, fonttitle=\bfseries,
  left=4pt, right=4pt, top=3pt, bottom=3pt
]
\footnotesize

\textit{The model recognised the closed answer set (A/B, Yes/No, or a fixed ordering) but selected the wrong element---typically swapping two options.}
\vspace{0.4em}

\textbf{Model:} glm4.6v \hfill \textbf{Task:} TR --- Duration Comparison \hfill \textbf{Context:} 32K
\vspace{0.3em}

\textbf{Question.} Which of the following two durations is longer? Duration 1: the time spent working at JTB Corporation; Duration 2: the time spent working at Rakuten Group.
\vspace{0.3em}

\textbf{Reference Answer.} B

\textbf{Model Prediction.} \textcolor{red}{a}

\end{tcolorbox}

\begin{tcolorbox}[
  title={Case Study: Stale Retrieval},
  enhanced, sharp corners,
  colback=gray!3, colframe=black!40,
  boxrule=0.4pt, fonttitle=\bfseries,
  left=4pt, right=4pt, top=3pt, bottom=3pt
]
\footnotesize

\textit{The model retrieved a factually correct but \emph{outdated} answer---returning a previous state instead of the most recent update in the knowledge-update chain.}
\vspace{0.4em}

\textbf{Model:} cosmos-reason2-8b \hfill \textbf{Task:} Knowledge Update \hfill \textbf{Context:} 32K
\vspace{0.3em}

\textbf{Question.} What's my favorite dip now?
\vspace{0.3em}

\textbf{Reference Answer.} spinach artichoke dip

\textbf{Model Prediction.} \textcolor{red}{Hummus}

\textbf{Previous (Outdated) Answer.} \textcolor{gray}{hummus}

\end{tcolorbox}

\begin{tcolorbox}[
  title={Case Study: Unsupported Answer},
  enhanced, sharp corners,
  colback=gray!3, colframe=black!40,
  boxrule=0.4pt, fonttitle=\bfseries,
  left=4pt, right=4pt, top=3pt, bottom=3pt
]
\footnotesize

\textit{The model's prediction has no meaningful semantic overlap with the reference; the response appears fabricated with no correct evidence anchor.}
\vspace{0.4em}

\textbf{Model:} cosmos-reason2-8b \hfill \textbf{Task:} MSR --- Arithmetic \hfill \textbf{Context:} 32K
\vspace{0.3em}

\textbf{Question.} How much total have I spent on blenders?
\vspace{0.3em}

\textbf{Reference Answer.} \$433.99

\textbf{Model Prediction.} \textcolor{red}{You have spent a total of \$40 on blenders.}

\end{tcolorbox}

\begin{tcolorbox}[
  title={Case Study: Answerability Failure},
  enhanced, sharp corners,
  colback=gray!3, colframe=black!40,
  boxrule=0.4pt, fonttitle=\bfseries,
  left=4pt, right=4pt, top=3pt, bottom=3pt
]
\footnotesize

\textit{The question is deliberately unanswerable (evidence image removed), yet the model produced a confident substantive answer instead of refusing.}
\vspace{0.4em}

\textbf{Model:} glm4.5v \hfill \textbf{Task:} Answer Refusal \hfill \textbf{Context:} 32K
\vspace{0.3em}

\textbf{Question.} What color are the words "rice is life" written in?
\vspace{0.3em}

\textbf{Reference Answer.} \texttt{NOT\_MENTIONED}

\textbf{Model Prediction.} \textcolor{red}{The words "rice is life" are written in black.}

\end{tcolorbox}

\begin{tcolorbox}[
  title={Case Study: Non-Answer Pathology},
  enhanced, sharp corners,
  colback=gray!3, colframe=black!40,
  boxrule=0.4pt, fonttitle=\bfseries,
  left=4pt, right=4pt, top=3pt, bottom=3pt
]
\footnotesize

\textit{The model either produced an empty response or generated verbose reasoning without an extractable final answer.}
\vspace{0.4em}

\textbf{Model:} glm4.5v \hfill \textbf{Task:} MSR --- Entity Resolution \hfill \textbf{Context:} 32K
\vspace{0.3em}

\textbf{Question.} Based on the tweets and my observations, is my vehicle, Bluebird, subject to the recall mentioned by @Woolworths\_SA?
\vspace{0.3em}

\textbf{Reference Answer.} No

\textbf{Model Prediction.} \textcolor{red}{To determine if your vehicle, Bluebird, is subject to the recall mentioned by @Woolworths\_SA, follow these steps: 1. Identify the Recall Details: Check the Woolworths tweet for the specific vehicle mo\ldots}

\end{tcolorbox}

\subsection{Context-Length Analysis}
\label{app:extended_analysis}

This appendix collects the context-length-centric analyses that complement the memory-ability and multimodal perspectives in the main text.

\begin{figure}[h]
\centering
\includegraphics[width=0.85\linewidth]{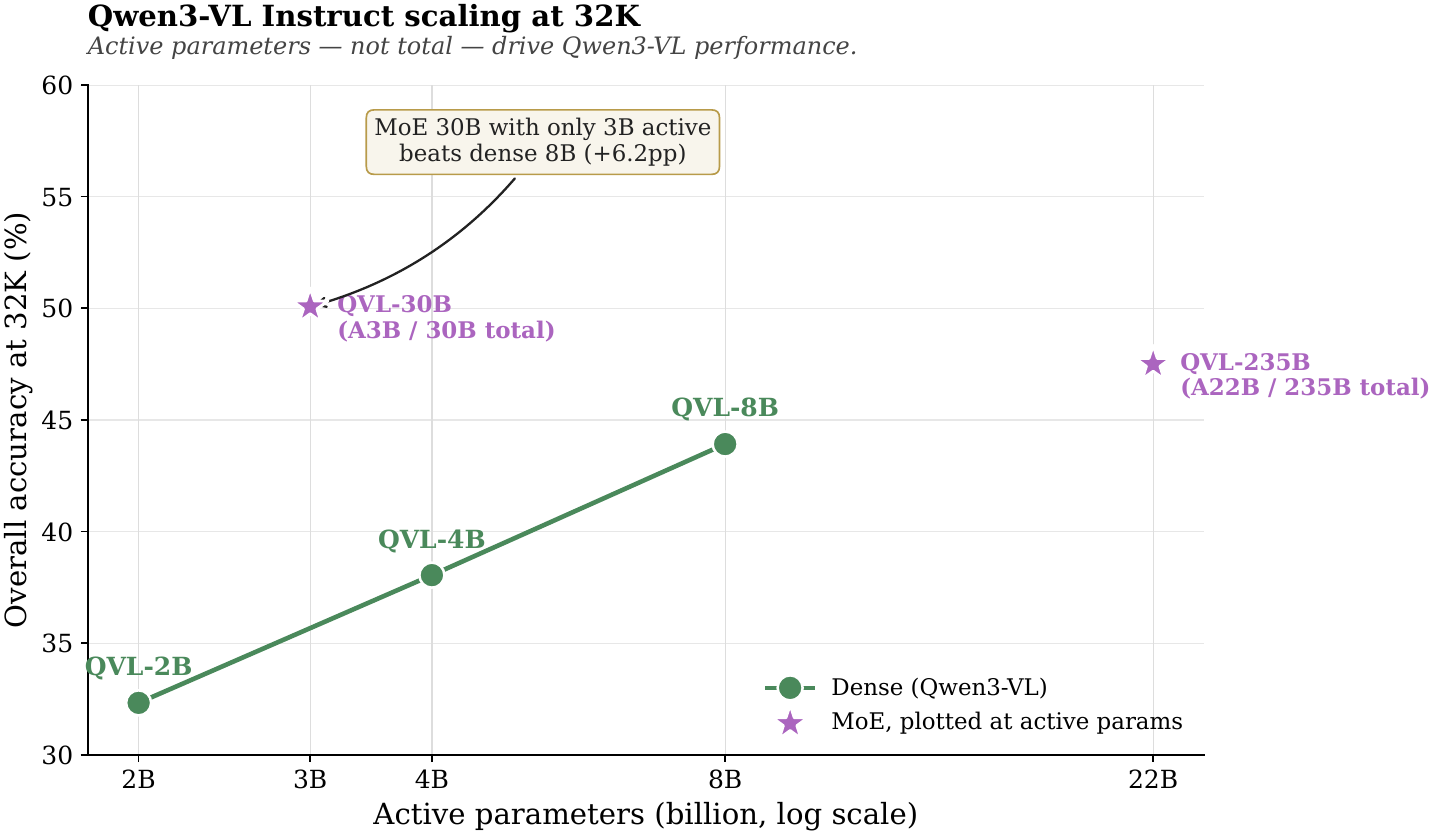}
\caption{Model-size scaling within the Qwen3-VL Instruct family ($n=789$). Dense scaling from 2B to 8B is monotone; the 30B (3B-active) MoE outperforms the 8B dense variant, indicating that active-parameter efficiency drives short-context performance more than total size.}
\label{fig:scaling_curves}
\end{figure}

\paragraph{Size helps long-context retention within a family.}
Among the 22 open-source LVLMs with complete 32K--128K data, the Spearman correlation between model size and the 32K-to-128K retention ratio reaches $\rho = 0.62$ ($p = 0.002$, Figure~\ref{fig:scaling_curves}): larger models generally hold up better at 128K. Within the Qwen3-VL Instruct family, retention climbs from $0.71$ at 8B to $0.90$ at 235B; within the Qwen3.5 dense series, 27B retains $0.77$ versus $0.65$ at 9B and $0.57$ at 4B.

\paragraph{Architectural family shifts both the ceiling and the degradation profile.}
At roughly matched scale, Qwen3.5-122B-A10B loses more than twice as much accuracy at 128K as Qwen3-VL-235B-A22B ($-$13.18\% vs.\ $-$5.57\%), despite similar active-parameter budgets. The type-wise profile also diverges: the Qwen3-VL Instruct branch keeps TR high at 128K (52--55\%) but lets KU collapse (30--24\%), whereas the Qwen3.5 branch preserves KU (46.03\%) at the cost of TR (41.75\%). Family choice therefore shapes both overall retention and which question types give way first.

\paragraph{No model family wins on every type.}
Every top-tier open-weight model at 128K is an MoE variant, so the informative architectural axis is training recipe, not sparse versus dense. Kimi-K2.5 leads MSR at 32K (44.06\%) but drops to 28.67\% at 128K; Gemini-3.1-Pro is the only model simultaneously top-2 on IE, best on KU, and above 29\% on MSR at 128K. Specialization follows training recipe rather than dense-vs-sparse design.

\paragraph{Hard questions stay hard across context lengths.}
The 280 questions that are hard at 32K (solved by $<$20\% of models) average 9.07\% accuracy at 32K and 8.52\% at 128K, so expanding context does not unlock them. The difficulty floor reflects the skill requirements rather than an artifact of context length.

\paragraph{Statistical validation of degradation monotonicity.}
In a well-behaved needle-in-haystack benchmark, accuracy should decrease (or hold flat) as context grows, since the evidence is identical and only the surrounding noise increases.
We systematically scanned all 27~LVLMs and 7~agents across five types and two context transitions (32K$\to$64K, 64K$\to$128K), identifying 72 transitions where accuracy nominally increases---38 among LVLMs and 34 among agents.
None survive Bonferroni correction ($\alpha/72 = 0.0007$).
For the 38 LVLM transitions we applied McNemar's exact test on paired per-question binary outcomes (same 789 questions at both lengths), classifying each question as stable-correct, stable-wrong, flip-to-correct, or flip-to-wrong.
Only one LVLM transition reaches $p < 0.05$ uncorrected: Qwen3-VL-2B-Thinking MSR at 64K$\to$128K ($+$5.59\%, $p=0.022$, 9 flip-to-correct vs.\ 1 flip-to-wrong).
This improvement is a degenerate-output artifact: 8 of 9 flip-to-correct questions hit the 2{,}048-token generation cap at 64K but produced normal outputs at 128K---stochastic recovery in a highly unstable model (60.8\% MSR degenerate rate).
All 34 agent anomalies are attributable to small per-type sample sizes ($n = 22$--$61$): a single question flip produces a 2--5\% swing.

\paragraph{Bidirectional churn underlies apparent reversals.}
Per-question tracking across 21~LVLMs with judge data at all three context lengths reveals that context transitions induce \emph{bidirectional} churn: at each step, models simultaneously lose 60--150 questions they previously answered correctly and gain 30--80 new correct answers.
The observed accuracy decay is the net of these two opposing flows, not a uniform loss of signal.
Of questions that flip at 32K$\to$64K, 38.7\% flip back at 64K$\to$128K, confirming that a substantial fraction of single-step churn is stochastic.
Context-robust models are not exempt from this churn---they simply balance most of it: at 64K$\to$128K, Gemini-3.1-Pro flips ${\sim}$71 questions in each direction (correct$\to$wrong and wrong$\to$correct) at an ${\approx}$18\% churn rate, and Kimi-K2.5 flips ${\sim}$83 in each direction; the slight asymmetry yields the small net Overall drops shown in Table~\ref{tab:per_type_full_vlm} (1pp for Gemini, 2pp for Kimi) rather than a true zero-sum reversal.
This balanced replacement, rather than static resistance, explains their flat accuracy trajectories.

We also tested whether MSR questions are systematically easier at 64K, since 10 individual LVLMs show MSR improvement at that length.
Cross-model question-level analysis reveals no systematic positioning effect: more MSR questions become harder at 64K than easier across models, ruling out a dataset-level artifact.

In summary, the expected monotonic degradation holds across the benchmark: no context-length artifact is detected, and the rare per-model reversals are explained by bidirectional churn, model-specific answer variability, or small-sample noise in agent subsets.

\paragraph{AR accuracy degradation.}
\label{app:ar_degrade}

AR accuracy degrades monotonically with context length across all Qwen3-VL sizes, but the rate depends on both scale and decoding mode (Table~\ref{tab:ar_degrade}). At the 235B Instruct tier the drop is modest ($-$5.6\% from 32K to 128K), whereas 2B-Thinking collapses from 87.8\% to 14.4\% ($-$73.3\%): truncated reasoning traces produce substantive answers instead of refusals.

\begin{table}[h]
\centering
\caption{Answer Refusal (AR) accuracy across input lengths for Qwen3-VL Instruct vs.\ Thinking modes. All models lose AR accuracy at longer inputs; thinking mode degrades faster.}
\label{tab:ar_degrade}
\small
\begin{tabular}{@{}llccc|c@{}}
\toprule
\textbf{Size} & \textbf{Mode} & \textbf{32K AR} & \textbf{64K AR} & \textbf{128K AR} & \textbf{$\Delta_{32\to128}$} \\
\midrule
235B & Instruct & 97.8 & 95.6 & 92.2 & $-$5.6 \\
235B & Thinking & 88.9 & 75.6 & 67.8 & $-$21.1 \\
30B & Instruct & 93.3 & 78.9 & 63.3 & $-$30.0 \\
30B & Thinking & 71.1 & 65.6 & 60.0 & $-$11.1 \\
8B & Instruct & 90.0 & 82.2 & 66.7 & $-$23.3 \\
8B & Thinking & 80.0 & 54.4 & 45.6 & $-$34.4 \\
4B & Instruct & 92.2 & 75.6 & 50.0 & $-$42.2 \\
4B & Thinking & 82.2 & 70.0 & 52.2 & $-$30.0 \\
2B & Instruct & 72.2 & 65.6 & 47.8 & $-$24.4 \\
2B & Thinking & 87.8 & 54.4 & 14.4 & $-$73.3 \\
\bottomrule
\end{tabular}
\end{table}

\subsection{Agent Underperformance: Where in the Pipeline Is Information Lost?}
\label{app:agent_underperformance}

The main-text analysis (\S\ref{subsec:analysis}) establishes that agents trail LVLMs by 18--34\% overall, exhibit inverted type profiles, and suffer a 22\% modality gap on image-essential questions. This appendix asks \emph{where in the pipeline} the information is lost. We evaluate all seven agents on the canonical 195-question subset of Appendix~\ref{app:canonical195}; the 32K-to-256K per-type roster is reproduced in Table~\ref{tab:per_type_full_agent}.

\begin{figure}[h]
\centering
\includegraphics[width=0.85\linewidth]{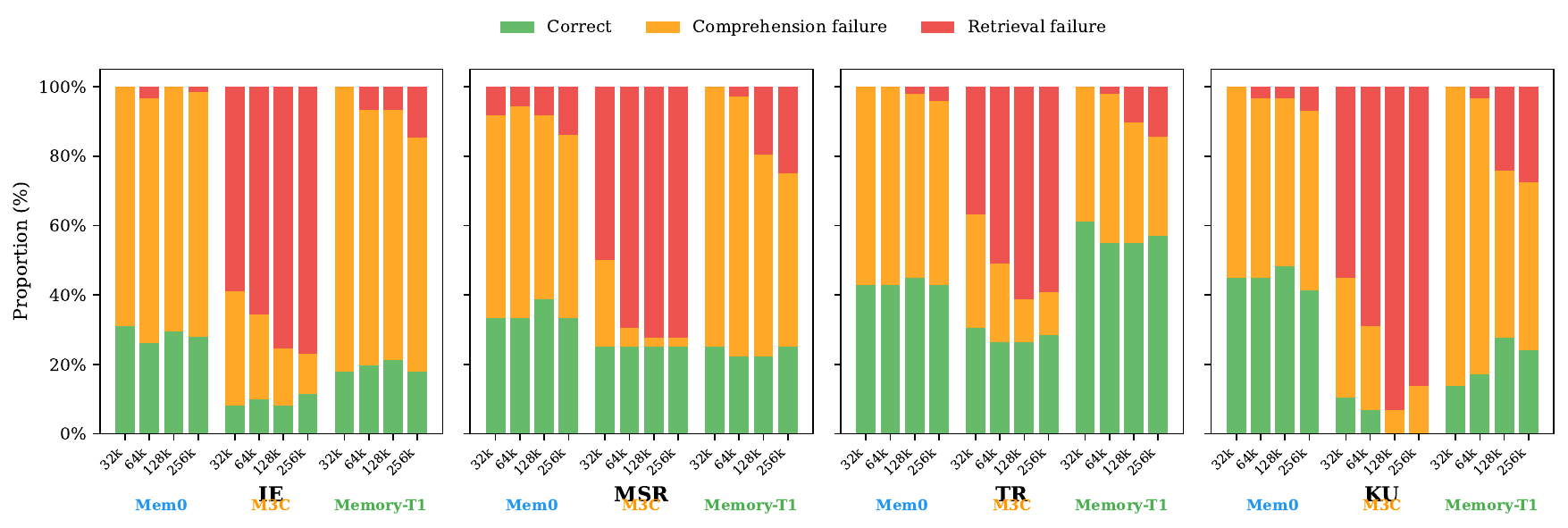}
\caption{Retrieval attribution for three agents with retrieval logs at 32K, decomposed by question type. Each bar partitions outcomes into correct (green), comprehension failure (yellow, evidence retrieved but answer wrong), and retrieval failure (red, evidence not retrieved). M3C is retrieval-dominated across all types; Mem0 and Memory-T1 are comprehension-dominated despite high evidence recall.}
\label{fig:retrieval_attribution}
\end{figure}

\paragraph{Agent failures split into retrieval-dominated and comprehension-dominated modes.}
For the three agents with retrieval logs (Mem0, Memory-T1, and M3C), we decompose each wrong answer by whether the evidence was retrieved before the error occurred. Evidence recall is defined as the fraction of ground-truth evidence sessions surfaced by the retriever, and an error is classified as a retrieval failure when recall falls below 0.5 (fewer than half of the required evidence sessions are retrieved); the threshold is conservative, since most questions in \bench{} require all evidence sessions to answer correctly. M3C is retrieval-bottlenecked, with 78.1\% of its errors occurring because the LoRA session retriever never surfaces the relevant evidence (mean recall 0.26). Mem0 and Memory-T1 sit at the opposite end: they retrieve the evidence at high recall (0.82 to 0.89), yet 87 to 95\% of their errors occur after successful retrieval, indicating that the backbone model cannot reason over the surfaced content (Figure~\ref{fig:retrieval_attribution}). The two failure modes call for different interventions---better retrieval for M3C and stronger reading comprehension for Mem0 and Memory-T1.

\paragraph{Pipeline architecture dominates backbone quality.}
M2A builds on Qwen3-VL-8B-Instruct, a current-generation 8B stock backbone, yet it scores only 14.21\%. Memory-T1, whose backbone is a 2.5$\times$ smaller text-only Qwen2.5-3B with RL fine-tuning, scores 29.50\%, a 15.29\% advantage on a weaker substrate. The comparison conflates architecture with task-specific training, since the RL objective of Memory-T1 targets temporal retrieval and partly explains its TR dominance (63.46\%), but the direction of the effect is consistent across agents. The cleanest quantitative measure of pipeline cost is the backbone-matched contrast: M2A at 14.21\% against direct evaluation of Qwen3-VL-8B-Instruct at 49.18\% yields a 34.97\% deficit on the same backbone. Mem0 (32.50\%, Qwen3-8B backbone) and MemOS (34.00\%, Qwen3-8B backbone) share the text-only Qwen3-8B substrate but cannot be matched against a direct LVLM at the same scale, since text-only Qwen3-8B is not in our LVLM roster; their backbone ablations (Table~\ref{tab:backbone_ablation}) span 14.65\% for Mem0 and 2.50\% for MemOS across the alternative backbones, large in absolute terms yet still well below the 34.97\% deficit measured on M2A. Taken together, the overall agent system---combining architecture and training recipe---matters more than backbone scale alone, a conclusion now supported by both the cross-agent comparison above and the within-architecture ablation below.

\paragraph{Controlled backbone ablation.}
The cross-agent comparison above is confounded by architecture, training recipe, and backbone quality varying simultaneously. To isolate the backbone factor, we re-evaluate Mem0 and MemOS with alternative backbones while keeping each pipeline unchanged (Table~\ref{tab:backbone_ablation}). Within the Mem0 FAISS architecture, swapping the default Qwen3-8B for the larger gpt-4.1-mini lifts overall accuracy by 10.65\% to 43.15\%, while substituting Qwen2.5-7B drops it by 4.00\% to 28.50\%, for a total spread of 14.65\% across the three backbones. Within MemOS, replacing the default Qwen3-8B with Qwen2.5-7B counter-intuitively raises overall accuracy by 2.50\%, indicating that newer-generation backbones do not always translate to better in-pipeline behavior at this scale. However, both spreads remain well below the 34.97\% deficit measured on M2A, reinforcing the conclusion that architecture is the dominant factor.

Per-type profiles shift markedly across backbones. In Mem0, the default Qwen3-8B reaches 77.27\% AR while the Qwen2.5-7B variant achieves perfect refusal (100\%), indicating that the backbone's intrinsic calibration against hallucination propagates through the memory pipeline. Conversely, Qwen3-8B leads on TR (50.00\% vs.\ 32.69\%) within the same FAISS architecture, suggesting complementary strengths that the pipeline cannot arbitrate. Crucially, all backbone variants preserve the context-length invariance observed for the default configurations: the 32K-to-256K range stays below 5\% for every variant, confirming that flatness is an architectural property independent of backbone quality.

\begin{table}[h]
\centering
\small
\caption{Backbone ablation within fixed agent architectures at 32K.
Each row uses the same pipeline with a different backbone model,
evaluated on the 195-question canonical subset.
$\Delta$: spread between best and worst backbone within each framework.}
\label{tab:backbone_ablation}
\begin{tabular}{@{}llcccccc@{}}
\toprule
Framework & Backbone & IE & MSR & TR & KU & AR & Ov. \\
\midrule
Mem0 & Qwen3-8B (default) & 13.11 & 25.00 & 50.00 & 17.24 & 77.27 & 32.50 \\
Mem0 & gpt-4.1-mini & 31.15 & 33.33 & 42.86 & 44.83 & 90.91 & 43.15 \\
Mem0 & Qwen2.5-7B & 3.28 & 19.44 & 32.69 & 31.03 & 100.00 & 28.50 \\
\cmidrule{2-8}
& $\Delta$ & & & & & & 14.65 \\
\midrule
MemOS & Qwen3-8B (default) & 18.03 & 22.22 & 40.38 & 24.14 & 68.18 & 34.00 \\
MemOS & Qwen2.5-7B & 29.51 & 22.22 & 40.38 & 20.69 & 90.91 & 36.50 \\
\cmidrule{2-8}
& $\Delta$ & & & & & & 2.50 \\
\bottomrule
\end{tabular}
\end{table}

\paragraph{Context invariance is real but insufficient.}
The context stability noted in \S\ref{subsec:analysis}, with six of seven agents staying within $\pm 7\%$ across 32K to 256K, is genuine robustness rather than a floor effect. On correctly answered questions, the Jaccard overlap between adjacent context lengths exceeds the random baseline of independent draws at the observed accuracy by a factor of 3.3 to 6.9 across agents. At 128K this advantage narrows the gap to the Qwen3-VL-8B tier from 35\% to 17\%, yet the absolute deficit remains large because the information lost during memorization and retrieval continues to outweigh the context-robustness advantage.

\paragraph{A common question subset defeats multiple agents regardless of architecture.}
On the shared answerable subset at 32K, 52\% of the questions are answered incorrectly by all four agents with complete per-question logs (M3-Agent, M3C, M2A, and Memory-T1, with overall accuracy in the 14 to 30\% range), despite architecturally disjoint retrieval pipelines that include ColPali (M3-Agent), LoRA session retrieval (M3C), dual-layer SQLite (M2A), and BM25 (Memory-T1). Mem0, MemOS, and MemAgent-7B are excluded from this overlap analysis because their evaluation outputs lack per-question identifiers needed for cross-agent alignment, and the four included agents already span the four retrieval paradigms represented in our pool. The all-wrong set is dominated by KU (69\%) and MSR (69\%), with TR the least affected type (21\%). Whether this ceiling persists for stronger agents or instead reflects a shared bottleneck of current sub-10B retrieval-based memory systems is an open question that would require matched-scale implementations we leave to future work.

\subsection{MSR Ceiling Diagnostic: Retrieval-Bounded Difficulty}
\label{app:msr_ceiling}

MSR is the hardest ability in \bench{} (\S\ref{subsec:main_results}), with cross-session aggregation over three to eight sessions defeating every evaluated system. A natural concern is whether this ceiling reflects intrinsic question difficulty---i.e., the aggregation tasks are too hard for current models---or a retrieval bottleneck imposed by the long-context evaluation format.

The cross-modality ablation (Table~\ref{tab:mm_purity}) provides a direct diagnostic. When frontier models receive ground-truth evidence sessions with their associated images---bypassing the haystack retrieval challenge entirely---MSR accuracy reaches 100.00\% for GPT-5.4 and 90.21\% for Gemini-3.1-Pro. Both models achieve near-perfect cross-session aggregation when the required three to eight evidence sessions are delivered directly, confirming that the counting and arithmetic operations underlying MSR are well within frontier reasoning capacity.

The gap between oracle-retrieval MSR (90--100\%) and full-benchmark MSR therefore confirms that the ceiling is \emph{retrieval-bounded} rather than \emph{reasoning-bounded}: models can solve the aggregation task once evidence is located, but fail to collect the required evidence sessions from a long multi-session conversation. This interpretation refines the per-type error analysis in \S\ref{subsec:analysis}: surface aggregation failures in the error breakdown are downstream consequences of incomplete evidence collection rather than deficits in reasoning capacity. When a model locates only four of six required sessions, its count is necessarily wrong, registering as an aggregation error in the taxonomy despite originating from a retrieval miss. The remaining ${\sim}$10\% gap for Gemini-3.1-Pro (90.21\% vs.\ 100\%) suggests that a small fraction of MSR items---likely the more complex arithmetic patterns (MSR-Arithmetic)---do challenge reasoning capacity even with perfect evidence delivery, but the dominant bottleneck is evidence location, not evidence comprehension.

\subsection{Session Indistinguishability Validation}
\label{app:indistinguishability}

Evidence sessions and haystack sessions are produced by the same generation pipeline. Both use GPT-5.1 as the user and Gemini-3-Pro as the assistant, follow the same prompt templates with identical style constraints (250--350 words per turn, knowledge-oriented follow-ups rather than personal or social ones), and share the same image integration pathway (\S\ref{app:prompts-evidence}). The sole structural difference is that an evidence session embeds one or more needle facts into the user turns; a six-stage validator chain (rule-based length and format checks, n-gram overlap detection, semantic leakage filtering, chain-ambiguity preservation, and end-marker verification) then screens each generated turn, rejecting and regenerating any output that departs from the haystack register or reveals the embedded fact through surface cues. Session indistinguishability is therefore a construction-level property: because both session types emerge from the same generator under the same constraints, systematic stylistic divergence would require the validator chain to fail consistently across thousands of sessions.

The conversation history assembly stage imposes type-specific ordering constraints beyond the random placement and non-contamination invariants described in the main text. For MSR and KU questions, whose answers depend on the temporal sequence of facts across sessions (a four-step knowledge update chain or a cross-session counting task, for instance), the evidence sessions are inserted into the haystack sequence in their original fact order, preserving the chronological progression that the question probes. For TR questions, each evidence session is assigned a timestamp that strictly precedes the question's reference date, so that temporal reasoning over session dates remains internally consistent. These ordering constraints, together with the non-contamination invariant and the visual-density normalization by text-only filler sessions, ensure that evidence positions carry no structural signal beyond what random interleaving would produce.

To validate the indistinguishability claim quantitatively, we trained two binary classifiers to distinguish evidence from haystack sessions. From the evidence and haystack sessions in the benchmark, we drew a balanced 698/698 sample and applied five-fold cross-validation with question-grouped splits, ensuring that no question appears in both the training and test sets of any fold. Speaker labels were normalized to a uniform User:/Assistant: format and the \texttt{<image>} placeholder token was removed, eliminating trivial classification shortcuts. A TF-IDF logistic regression model (unigrams and bigrams, 10{,}000 features, sublinear term frequency) achieved 56.71\% $\pm$ 1.20\% accuracy with an ROC-AUC of 0.59, and a fine-tuned DeBERTa-v3-base classifier (512-token truncation, three epochs) reached 57.92\% $\pm$ 0.85\% with an ROC-AUC of 0.55. Both results sit only marginally above the 50\% chance floor for balanced binary classification (Cohen's $h < 0.16$, a small effect). A pooled binomial test against $H_0{:}~p = 0.5$ on the TF-IDF predictions (805 of 1{,}396 correct) yields a two-sided $p < 10^{-7}$; the rejection reflects the large sample size rather than a practically meaningful signal, as the absolute accuracy gap of ${\sim}7$ percentage points above chance provides negligible leverage for locating evidence within a 30{+}-session history.

A follow-up vocabulary ablation progressively removed the $K$ most discriminative n-grams from the TF-IDF feature set ($K = 20, 50, 100, 200, 500, 1{,}000$; balanced removal from each class direction, ranked on training-fold coefficients only to prevent test-fold leakage). Across the entire sweep, accuracy remained in a narrow band between 54.77\% and 56.71\%, never exceeding 7 percentage points above chance. Removing up to 1{,}000 top-ranked n-grams leaves accuracy essentially unchanged, indicating that the weak signal is diffuse rather than concentrated in identifiable stylistic markers. Together, the construction-level design and the post-hoc classifier results establish that evidence sessions carry no practically exploitable stylistic fingerprint relative to the surrounding haystack; the retrieval difficulty reported in \S\ref{subsec:main_results} is not inflated by surface-level cues that might allow a model to shortcut evidence location.

\section{Detailed Limitations and Future Work}
\label{app:limitations}

\paragraph{Synthetic conversations and human-in-the-loop naturalness.}
The evidence and haystack sessions in \bench{} are LLM-generated (GPT-5.1 user, Gemini-3-Pro assistant; \S\ref{subsec:construction}), with ShareGPT and UltraChat~\cite{ding2023enhancing} dialogues used as filler in the surrounding context. Naturalness is enforced through an intensive human-in-the-loop review pipeline rather than automatic metrics alone (\S\ref{app:annotation}): Round~1 audits each candidate question; Round~2 audits each session against a conversational-naturalness criterion covering colloquial phrasing, turn-taking coherence, and indirect embedding of factual content, returning stilted or overly formal sessions to the generation pipeline until annotators accept them as plausible user--assistant exchanges; Round~3 audits the assembled haystack. This pipeline reduces the candidate pool from ${\sim}20$k to 789 final questions, following the human-review practice established by LongMemEval~\cite{wu2025longmemevalbenchmarkingchatassistants}. As a complementary surface-shortcut control, the session-indistinguishability validation (\S\ref{app:indistinguishability}) trains binary classifiers to separate evidence from haystack on text features and reports only marginally above-chance accuracy (DeBERTa F1 57.92\%, TF-IDF 56.71\%), confirming that evidence carries negligible stylistic signal for localization. \bench{} therefore stands as a controlled diagnostic stress test for cross-modal evidence retrieval, multi-session aggregation, and length-controlled comparability under uniform construction; characterizing the gap to the distribution of real long-term human--assistant interactions remains an open research question.

\paragraph{Question generator and test-taker overlap.}
The final question-generation model in our pipeline is Gemini-3-Pro (\S\ref{subsec:construction}, \S\ref{app:abstraction}), and the top-evaluated model in our leaderboard is Gemini-3.1-Pro (\S\ref{subsec:main_results}, Table~\ref{tab:per_type_full_vlm}); these are different versions in the same Gemini-3 family, so the residual concern is intra-family generator familiarity rather than strict model-identity circularity. The construction pipeline constrains question content rather than rewarding any particular output style: each question is conditioned on a fixed (entity, image, abstracted paragraph) triple under a deterministic prompt template, and is then filtered by automated text-leakage and quality checks followed by human review (\S\ref{app:abstraction}, \S\ref{subsec:quality_control}). Two oracle-retrieval diagnostics already in the paper provide indirect evidence that the 128K leaderboard position of Gemini-3.1-Pro is not explained by intra-family generator familiarity. On the answerability test with full evidence supplied (Table~\ref{tab:mm_purity}), GPT-5.4 reaches 93.13\% overall versus 89.42\% for Gemini-3.1-Pro; on the MSR cross-modality oracle (\S\ref{app:msr_ceiling}), GPT-5.4 reaches 100.00\% versus 90.21\% for Gemini-3.1-Pro. Under conditions where retrieval is bypassed---the conditions in which generator-style familiarity would most plausibly manifest---Gemini-3.1-Pro does not lead, so the 128K full-benchmark inversion is more consistent with retrieval-robustness than with intra-family familiarity bias. We did not, however, run a direct test in which an independent generator (e.g., Claude-Sonnet-4.5, GPT-5.4) regenerates a stratified subset of questions on the same input triples and the top LVLMs are re-ranked on those items; this targeted ablation would isolate any residual generator-familiarity confound and is left to future work.

\paragraph{Judge limitations.}
LLM-as-judge validation in this paper covers 800 of 73{,}784 judge calls (${\approx}$1.08\%) for cross-family agreement against GPT-5.4-mini and 484 items for human verification by three annotators with consensus adjudication.
Three limitations follow.
First, $\kappa = 0.86$ measures judge-vs-consensus-label agreement; we do not separately report inter-annotator $\kappa$ among the three raters, since disagreements were resolved to a single label before release, so inter-annotator reliability on this benchmark remains an open question.
Second, the 29 false positives versus 2 false negatives indicate the judge is systematically lenient on borderline partial-match and verbose-correct outputs; the format-dependent bias correction (Appendix~\ref{app:judge_validation}) addresses the short-output subset but not the full leniency channel.
Third, per-subtype judge agreement (Appendix~\ref{app:judge_validation}) inherits the reporting taxonomy of \S\ref{app:subtype_detail} (9 subtypes, all $n \geq 46$); finer per-generator splits inside MSR Entity and TR Temporal Grounding remain statistically noisy.
A separate inter-annotator reliability study on the 120 double-annotation-tagged items and a per-rare-type audit are left to future work.

\paragraph{Static-length vs.\ streaming evaluation.}
\bench{} queries each frozen multi-session history offline at four context lengths (32K--256K tokens); a complementary online streaming protocol that preserves temporal causality between memory writes and queries~\cite{zheng2026lifedialbench} would further probe irreversible-update and forgetting dynamics in the multimodal setting, and is left to future work.

\paragraph{Broader impacts.}
\label{app:broader_impacts}
\bench{} provides a length-controlled diagnostic for multimodal conversational memory, intended to make memory-faithfulness regressions in long, multi-session multimodal assistants visible to model developers before deployment. Two negative impacts are foreseeable. First, leaderboard targeting can encourage over-fitting to the construction pipeline (\S\ref{subsec:construction}); the released anti-shortcut filter (\S\ref{app:prompts-filter}) and the canonical 195-question subset (\S\ref{app:canonical195}) are designed so that such over-fitting becomes detectable rather than rewarded. Second, the synthetic dialogue construction does not capture the full distribution of real user--assistant interactions, so accuracy on \bench{} is a necessary but not sufficient indicator of deployed memory quality. As mitigation, \bench{} is released with versioned frozen tags so that any specific evaluation run remains reproducible; images are sourced from the public web and each retains its original source-site license, as documented in \S\ref{app:image_release}.

\paragraph{Ethics statement.}
\bench{} is released for the evaluation of multimodal long-term conversational memory, and we do not endorse using its dialogues for training or fine-tuning, since exposure of evaluation items would compromise diagnostic value; the versioned frozen tags (\S\ref{app:image_release}) and the per-image perceptual hashes support detection of leakage in derived models. Two dual-use considerations follow. First, memory-augmented assistants whose deployed memory faithfulness regresses below the levels reported here may amplify mis-recall in user-facing settings; the per-ability decomposition (\S\ref{subsec:analysis}) is designed so that such regressions are visible before deployment. Second, the synthetic-dialogue construction (\S\ref{subsec:construction}) does not capture the distribution of real long-term human--assistant interactions, so \bench{} scores are a necessary but not sufficient indicator of deployed memory quality. Image sourcing, licensing, privacy, and takedown protocol are documented in Appendix~\ref{app:image_release}.

\section{Design Rationale for the Memory Ability Taxonomy}
\label{app:taxonomy_rationale}

This appendix motivates the five memory abilities evaluated in \bench{} and argues that they jointly cover the space of capabilities required for long-term multimodal conversational agents.

Recent memory-agent benchmarks converge on three functional dimensions of conversational memory~\cite{memgallery,wu2025longmemevalbenchmarkingchatassistants}: retrieval (can the agent recall stored information?), reasoning (can it synthesize inferences across distributed evidence?), and knowledge update (can it correctly update its internal state as the conversation evolves?). Our taxonomy is designed so that the five abilities collectively span all three dimensions. IE addresses retrieval at single-session granularity; MSR and TR exercise reasoning from complementary angles; and KU and AR probe opposing facets of knowledge update. We justify each ability below.

\paragraph{Information Extraction~(IE): single-session retrieval.}
IE operationalizes the most fundamental function of memory: accurate recall of previously encountered information. Each IE question requires the agent to locate a specific evidence session within a long interaction history and extract the relevant fact. The multimodal extension adds a layer of image understanding: Entity-subtype questions present an abstracted entity visible only in an image, requiring visual recognition before textual retrieval, while PrevInfo-subtype questions ask the agent to recall a visual detail from an image shared in an earlier session. This two-hop structure ensures that IE measures genuine cross-modal memory rather than unimodal text matching.

\paragraph{Multi-Session Reasoning~(MSR): cross-session aggregation.}
MSR extends single-session retrieval to aggregative reasoning over information distributed across three to eight sessions. The tested operations (counting, arithmetic, and entity resolution) all require the agent to identify multiple relevant sessions, extract the pertinent facts from each, and combine them into a coherent answer. We retain a subset of text-only MSR questions alongside multimodal ones. These text-only items serve as a controlled modality ablation: they isolate cross-session reasoning from the added difficulty of visual retrieval, yielding a cleaner signal of the reasoning capability itself. Including them also preserves ecological validity, since not every piece of cross-session information in a realistic conversation history involves images. Mem-Gallery~\cite{memgallery} adopts a compatible design in its Multi-entity Reasoning subtask, where the entities involved can be textual or visual.

\paragraph{Temporal Reasoning~(TR): temporal awareness.}
Temporal reasoning has emerged as a distinct research focus in LLM evaluation~\cite{wang2023cola,wang2022subeventwriter,wei2025time,wu2025longmemevalbenchmarkingchatassistants}. In long-running interaction histories, temporal information takes heterogeneous forms: natural-language date expressions in utterances, session-level timestamps in metadata, and implicit ordering from the session sequence itself. Our TR questions test whether the agent can jointly process these signals together with visual content (e.g., clock faces and calendar images that replace textual temporal expressions) to answer questions about durati fon comparison, event ordering, and temporal grounding. This ability is especially relevant for personal assistants, where users frequently ask questions that require temporal contextualization of past interactions.

\paragraph{Knowledge Update~(KU): current-state tracking.}
A defining requirement for personalized conversational agents is the ability to track how user attributes evolve over time. Users routinely update preferences, correct earlier statements, and revise plans across sessions~\cite{xu2024knowledge}. Each KU question presents a chain of four successive updates to a single attribute, and the agent must identify the current state rather than earlier, superseded values. This tests selective forgetting as much as recall: the agent must not only retrieve the relevant update chain but also recognize which value is most recent. Mem-Gallery~\cite{memgallery} identifies the same concern under Knowledge Resolution, framing it as maintaining consistency when ``new, contradictory information appears in the dialogue.''

\paragraph{Answer Refusal~(AR): epistemic calibration.}
AR tests whether the agent can recognize the limits of its available evidence. Each AR question is constructed by removing all evidence sessions from an otherwise answerable instance, so that no supporting information remains in the conversation history. A correct agent must decline to answer rather than hallucinate a plausible response. This ability addresses a well-documented failure mode of large language models~\cite{zhang2024rtuning} and is essential for trustworthy deployment: a personal assistant that fabricates answers from absent evidence is more harmful than one that acknowledges uncertainty.

\paragraph{Coverage and empirical distinctiveness.}
The five abilities span the retrieval--reasoning--update space with minimal redundancy. IE and MSR cover retrieval at single-session and cross-session granularity, respectively. TR and MSR address reasoning from complementary angles: temporal inference over timestamps and session metadata versus aggregative computation over distributed facts. KU and AR target opposite facets of knowledge update: KU penalizes failure to revise stored values, while AR penalizes failure to abstain. The cross-type Spearman correlation analysis in \S\ref{subsec:analysis} validates this design empirically: no pair of abilities exhibits consistently strong positive correlation across context lengths, confirming that each probes a distinct aspect of long-term multimodal memory.


\bigskip
\noindent Code: \url{https://github.com/xrenaf/MEMLENS} (the repository contains pointers to the dataset and per-image metadata).

\end{document}

%% file: extra_pkgs.tex
\usepackage[export]{adjustbox}
\usepackage[ruled]{algorithm2e}
\usepackage[inline, shortlabels]{enumitem}
\usepackage[T1]{fontenc}
\usepackage[hidelinks]{hyperref}
\usepackage{microtype}
\usepackage{pifont}
\usepackage{xcolor}
\usepackage{xurl}
\usepackage{graphicx}
\usepackage{booktabs}
\usepackage{tabularx}
\usepackage{tabularray}
\usepackage{listings}
\usepackage{amsmath, amsfonts}
\usepackage{nicefrac}
\usepackage[most]{tcolorbox}

\UseTblrLibrary{booktabs}

\newtcolorbox{promptbox}[2][]{
  enhanced, sharp corners, breakable,
  fonttitle=\bfseries,
  left=4pt, right=4pt, top=3pt, bottom=3pt,
  #1,
}

\makeatletter
\newcommand{\ssymbol}[1]{\@fnsymbol{#1}}
\newcommand{\romanNumeral}[1]{\expandafter\@slowromancap\romannumeral #1@}
\makeatother